\documentclass{article} %
\usepackage{iclr2027_conference,times}

\usepackage[utf8]{inputenc}
\usepackage[T1]{fontenc}
\usepackage{hyperref}
\usepackage{url}
\usepackage{booktabs}
\usepackage{array}
\usepackage{multirow}
\usepackage{amsfonts}
\usepackage{amsmath}
\usepackage{amssymb}
\usepackage{amsthm}
\usepackage{nicefrac}
\usepackage{microtype}
\usepackage{xcolor}
\usepackage{graphicx}
\usepackage{enumitem}
\usepackage{algorithm}
\usepackage{algpseudocode}
\usepackage{subcaption}
\usepackage{tikz}
\usetikzlibrary{arrows.meta}

\usepackage{pgfplots}
\pgfplotsset{compat=1.16}
\usetikzlibrary{backgrounds,calc,fit,positioning,shadows}
\usepgfplotslibrary{fillbetween,groupplots}

\newsavebox{\scaledfigurebox}
\newcommand{\figurefontsizes}{%
  \renewcommand{\tiny}{\fontsize{5}{6}\selectfont}%
  \renewcommand{\scriptsize}{\fontsize{7}{8}\selectfont}%
  \renewcommand{\footnotesize}{\fontsize{8}{9.5}\selectfont}%
  \renewcommand{\small}{\fontsize{9}{11}\selectfont}%
  \renewcommand{\normalsize}{\fontsize{10}{12}\selectfont}%
  \renewcommand{\large}{\fontsize{12}{14}\selectfont}%
  \setlength{\parindent}{15pt}\setlength{\parskip}{0pt plus 1pt}%
  \normalsize\ignorespaces}
\newenvironment{scaledfigure}[1][1pt]{\def\scaledfigureborder{#1}\begin{lrbox}{\scaledfigurebox}\figurefontsizes}{%
  \end{lrbox}%
  \resizebox{\linewidth}{!}{\raisebox{0pt}[\dimexpr\ht\scaledfigurebox+\scaledfigureborder\relax][\dimexpr\dp\scaledfigurebox+\scaledfigureborder\relax]{%
    \kern\scaledfigureborder\usebox{\scaledfigurebox}\kern\scaledfigureborder}}}

\definecolor{cppo}{HTML}{56B4E9}   %
\definecolor{cgrpo}{HTML}{0072B2}  %
\definecolor{cjppo}{HTML}{CC79A7}  %
\definecolor{cdj}{HTML}{009E73}    %
\definecolor{cace}{HTML}{D55E00}   %
\definecolor{cerl}{HTML}{999933}   %
\definecolor{cbasem}{HTML}{8C8C8C} %

\newcommand{\resultsfigstyle}{%
\pgfplotsset{
  panel/.style={
    axis lines=left, tick align=outside, ymajorgrids, grid style={black!8},
    tick label style={font=\tiny}, label style={font=\scriptsize},
    title style={font=\scriptsize, yshift=-5pt},
    xlabel shift=-3pt, ylabel shift=-3pt,
    yticklabel style={/pgf/number format/fixed,/pgf/number format/precision=1},
    scaled x ticks=false, unbounded coords=discard,
    every axis plot/.append style={line width=0.6pt, line join=round, mark=none},
  },
  m ppo/.style={cppo}, m grpo/.style={cgrpo}, m jppo/.style={cjppo},
  m dj/.style={cdj}, m erl/.style={cerl}, m ace/.style={cace, line width=0.9pt},
  m pass/.style={cgrpo, densely dashed},
  m base/.style={cbasem},
  pts/.style={mark=*, mark size=0.7pt, mark options={solid}},
  band/.style={fill=##1, fill opacity=0.15, draw=none, forget plot},
  legend style={font=\scriptsize, draw=none, fill=none, legend columns=-1,
    /tikz/every even column/.append style={column sep=6pt}},
  legend image post style={xscale=0.7},
}}
\newcommand{\testcurve}[4]{%
  \addplot[forget plot, #1, pts] table[x=step, y=#4, col sep=comma] {#3};
  \addplot[name path=up, draw=none, forget plot] table[x=step, y expr=\thisrow{#4}+\thisrow{#4_ci}, col sep=comma] {#3};
  \addplot[name path=lo, draw=none, forget plot] table[x=step, y expr=\thisrow{#4}-\thisrow{#4_ci}, col sep=comma] {#3};
  \addplot[band=#2] fill between[of=up and lo];
}
\newcommand{\scalecurve}[4]{%
  \addplot[forget plot, #1, mark=*, mark size=0.6pt, mark options={solid},
    error bars/.cd, y dir=both, y explicit, error bar style={line width=0.4pt, #2}, error mark options={rotate=90, mark size=1pt, #2}]
    table[x=k, y=#4, y error=#4_ci, col sep=comma] {#3};
}

\usepackage{listings}
\lstdefinestyle{prompt}{basicstyle=\ttfamily\scriptsize,breaklines=true,columns=fullflexible,keepspaces=true,showstringspaces=false,frame=single,rulecolor=\color{black!30}}
\usepackage{inconsolata}

\newcommand{\ace}{\textsc{ACE}}
\newcommand{\acerl}{\textsc{ACE-RL}}
\newcommand{\ehrl}{\textsc{EHRL}}
\newcommand{\grpo}{\textsc{GRPO}}

\newcolumntype{L}[1]{>{\raggedright\arraybackslash}p{#1}}

\title{Experience-Handoff Reinforcement Learning for \\ Recursive Self-Improvement}

\author{Doyoung Kim$^{1}$, Byeongguk Jeon$^{2}$, Kailin Zhang$^{1}$, Sherry Yang$^{1}$ \\
$^{1}$New York University \quad $^{2}$KAIST AI \\
\texttt{\{doyoungkim,sherryyang\}@nyu.edu}}

\iclrfinalcopy 

\begin{document}

\maketitle

\begin{abstract}
It is common to train Large Language Models (LLMs) for agentic tasks with Reinforcement Learning (RL) from scalar rewards. However, this throws away the fine-grained information about the environment that the agent experienced during the trial (e.g., error logs). To utilize this rich information, recent works wrap the agent in a harness that refines its own inputs through reflection or compaction and hands past experience to future rollouts, but the construction of the handed-off artifacts (e.g., a playbook) is itself not learned. We want the best of both worlds: gradient updates from a scalar reward, while also utilizing environment information to hand off experience. As a concrete testbed, we fix the Agentic Context Engineering (\ace{}) harness and train one model to act, reflect, and curate task-local playbooks. Our method, \acerl{}, evaluates candidate experience-handoff artifacts by their downstream effect on future actor rollouts, and co-optimizes the role-specific behaviors (action generator, insight reflector, and playbook curator) under a fixed scaffold. Across agentic benchmarks, \acerl{} achieves higher held-out performance than baseline methods. The results suggest that agent training should optimize not only the actions in the current rollout, but also the model-written experience that future rollouts inherit.
\end{abstract}
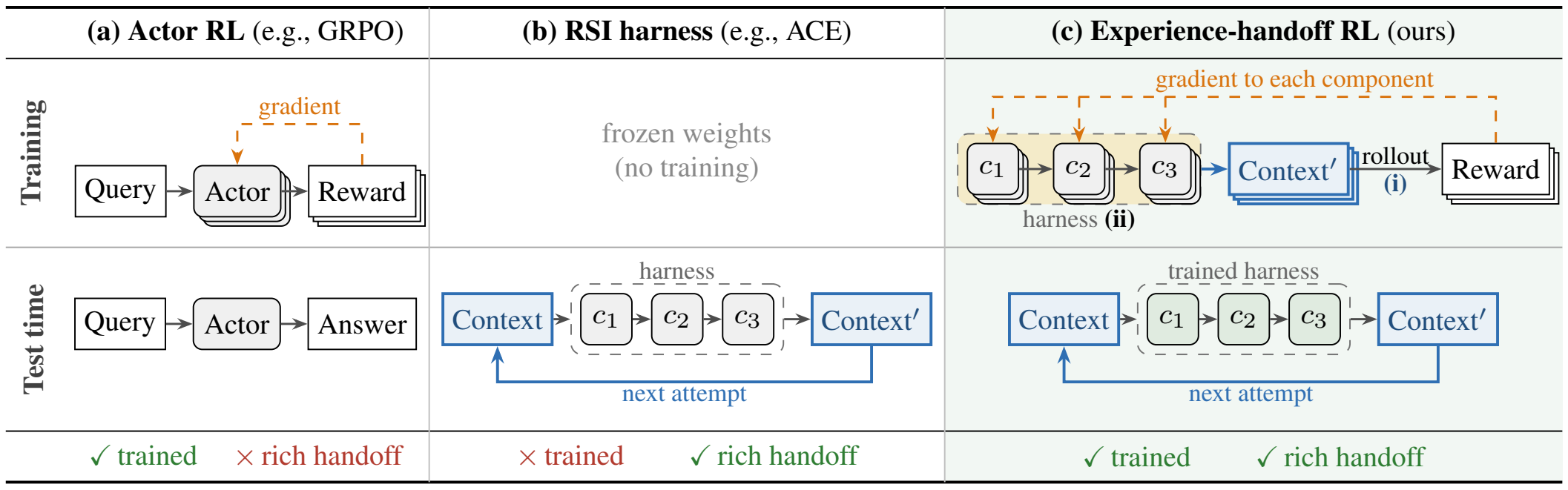
\begin{figure*}[!h]
  \centering
\definecolor{scalar}{HTML}{D9730D}   %
\definecolor{artif}{HTML}{2F6FB3}    %
\definecolor{grp}{HTML}{F4EBC9}      %
\definecolor{mut}{HTML}{8A8A8A}
\definecolor{bad}{HTML}{B23A2E}
\definecolor{good}{HTML}{2E7D32}
\tikzset{
  font=\footnotesize,
  role/.style={draw,rounded corners=2.5pt,fill=gray!12,minimum height=5mm,minimum width=8.5mm,inner ysep=1.5pt,inner xsep=2.6pt},
  comp/.style={role,minimum width=5mm,inner xsep=2pt},
  art/.style={draw,minimum height=5mm,minimum width=8.5mm,inner ysep=1.5pt,inner xsep=2.6pt},
  ctx/.style={art,draw=artif,fill=artif!10,text=artif!70!black,thick},
  trained/.style={fill=good!15},
  harness/.style={draw=black!55,dashed,rounded corners=3pt,inner sep=2.5pt},
  hlab/.style={font=\scriptsize,text=black!60,inner sep=1pt},
  grpstack/.style={double copy shadow={shadow xshift=1.3pt,shadow yshift=-1.3pt}},
  flow/.style={-{Stealth[length=1.5mm]},semithick,draw=black!70},
  grad/.style={-{Stealth[length=1.6mm]},semithick,dashed,draw=scalar},
  hand/.style={-{Stealth[length=1.7mm]},thick,draw=artif},
  colhdr/.style={font=\footnotesize\bfseries,anchor=center},
  rowhdr/.style={font=\footnotesize\bfseries,rotate=90,anchor=center,text=black!75},
  tag/.style={font=\scriptsize\bfseries,inner sep=1pt},
}
\newcommand{\yes}[1]{\textcolor{good}{\checkmark\,#1}}
\newcommand{\no}[1]{\textcolor{bad}{$\times$\,#1}}
\begin{scaledfigure}[1pt]
\newsavebox\aTr \newsavebox\aTe \newsavebox\bTr \newsavebox\bTe \newsavebox\cTr \newsavebox\cTe
\newsavebox\aHd \newsavebox\bHd \newsavebox\cHd%
\newcommand{\gap}{2.6mm}%

\sbox\aTr{\begin{tikzpicture}
\node[art] (q) {Query};
\node[role,grpstack,right=\gap of q] (m) {Actor};
\node[art,grpstack,fill=white,right=\gap of m] (r) {Reward};
\draw[flow] (q)--(m); \draw[flow] (m)--(r);
\draw[grad] (r.north) -- ++(0,0.4) -| node[pos=0.25,above,font=\scriptsize,text=scalar,inner sep=1pt]{gradient} (m.north);
\end{tikzpicture}}%

\sbox\aTe{\begin{tikzpicture}
\node[art] (q) {Query};
\node[role,right=\gap of q] (m) {Actor};
\node[art,right=\gap of m] (r) {Answer};
\draw[flow] (q)--(m); \draw[flow] (m)--(r);
\end{tikzpicture}}%

\sbox\bTr{\begin{tikzpicture}
\node[text=mut,align=center] {frozen weights\\(no training)};
\end{tikzpicture}}%

\sbox\bTe{\begin{tikzpicture}
\node[ctx] (x) {Context};
\node[comp,right=\gap of x] (h1) {$c_1$};
\node[comp,right=1.8mm of h1] (h2) {$c_2$};
\node[comp,right=1.8mm of h2] (h3) {$c_3$};
\node[harness,fit=(h1)(h3)] (H) {};
\node[hlab,above=0pt of H] {harness};
\node[ctx,right=\gap of H] (y) {Context$'$};
\draw[flow] (x)--(H); \draw[flow] (h1)--(h2); \draw[flow] (h2)--(h3); \draw[flow] (H)--(y);
\draw[hand] (y.south) -- ++(0,-0.35) -| node[pos=0.25,below,font=\scriptsize,text=artif,inner sep=1pt]{next attempt} (x.south);
\end{tikzpicture}}%

\sbox\cTr{\begin{tikzpicture}
\node[comp,grpstack] (h1) {$c_1$};
\node[comp,grpstack,right=3.3mm of h1] (h2) {$c_2$};
\node[comp,grpstack,right=3.3mm of h2] (h3) {$c_3$};
\draw[flow] (h1)--(h2); \draw[flow] (h2)--(h3);
\begin{scope}[on background layer]
  \node[fill=grp,harness,fit=(h1)(h3)] (H) {};
\end{scope}
\node[hlab,below=0pt of H] {harness \textbf{\textcolor{black}{(ii)}}};
\node[ctx,grpstack,right=\gap of H] (y) {Context$'$};
\node[art,grpstack,fill=white,right=9mm of y] (r) {Reward};
\draw[hand] (H)--(y);
\draw[flow] (y) -- node[above,font=\scriptsize,inner sep=1pt]{rollout} node[below,tag,text=artif!70!black]{(i)} (r);
\draw[grad] (r.north) -- ++(0,0.45) coordinate (bus) -| node[pos=0.2,above,font=\scriptsize,text=scalar,inner sep=1pt]{gradient to each component} (h1.north);
\draw[grad] (h2.north|-bus) -- (h2.north);
\draw[grad] (h3.north|-bus) -- (h3.north);
\end{tikzpicture}}%

\sbox\cTe{\begin{tikzpicture}
\node[ctx] (x) {Context};
\node[comp,trained,right=\gap of x] (h1) {$c_1$};
\node[comp,trained,right=1.8mm of h1] (h2) {$c_2$};
\node[comp,trained,right=1.8mm of h2] (h3) {$c_3$};
\node[harness,fit=(h1)(h3)] (H) {};
\node[hlab,above=0pt of H] {trained harness};
\node[ctx,right=\gap of H] (y) {Context$'$};
\draw[flow] (x)--(H); \draw[flow] (h1)--(h2); \draw[flow] (h2)--(h3); \draw[flow] (H)--(y);
\draw[hand] (y.south) -- ++(0,-0.35) -| node[pos=0.25,below,font=\scriptsize,text=artif,inner sep=1pt]{next attempt} (x.south);
\end{tikzpicture}}%

\sbox\aHd{\footnotesize\bfseries (a) Actor RL \textmd{(e.g., GRPO)}}%
\sbox\bHd{\footnotesize\bfseries (b) RSI harness \textmd{(e.g., ACE)}}%
\sbox\cHd{\footnotesize\bfseries (c) Experience-handoff RL \textmd{(ours)}}%

\begin{tikzpicture}
\pgfmathsetmacro\pad{0.12}  %
\pgfmathsetmacro\wa{max(\wd\aTr,\wd\aTe,\wd\aHd)/28.4527+2*\pad}
\pgfmathsetmacro\wb{max(\wd\bTr,\wd\bTe,\wd\bHd)/28.4527+2*\pad}
\pgfmathsetmacro\wc{max(\wd\cTr,\wd\cTe,\wd\cHd)/28.4527+2*\pad}
\pgfmathsetmacro\xa{0.55}
\pgfmathsetmacro\xb{\xa+\wa}
\pgfmathsetmacro\xc{\xb+\wb}
\pgfmathsetmacro\xe{\xc+\wc}
\typeout{FIGWIDTH=\xe cm}
\begin{scope}[on background layer]
  \fill[good!6] (\xc,0) rectangle (\xe,-4.65);   %
\end{scope}
\draw[thick] (0,0) -- (\xe,0);
\draw (0,-0.5) -- (\xe,-0.5);
\draw[black!35] (0,-2.35) -- (\xe,-2.35);
\draw (0,-4.15) -- (\xe,-4.15);
\draw[thick] (0,-4.65) -- (\xe,-4.65);
\foreach \x in {\xb,\xc} \draw[black!25] (\x,0) -- (\x,-4.65);

\node[anchor=center] at ({\xa+\wa/2},-0.25) {\usebox\aHd};
\node[anchor=center] at ({\xb+\wb/2},-0.25) {\usebox\bHd};
\node[anchor=center] at ({\xc+\wc/2},-0.25) {\usebox\cHd};
\node[rowhdr] at (0.27,-1.425) {Training};
\node[rowhdr] at (0.27,-3.25) {Test time};

\node[inner sep=0] at ({\xa+\wa/2},-1.45) {\usebox\aTr};
\node[inner sep=0] at ({\xa+\wa/2},-3.1) {\usebox\aTe};
\node[inner sep=0] at ({\xb+\wb/2},-1.425) {\usebox\bTr};
\node[inner sep=0] at ({\xb+\wb/2},-3.2) {\usebox\bTe};
\node[inner sep=0] at ({\xc+\wc/2},-1.4) {\usebox\cTr};
\node[inner sep=0] at ({\xc+\wc/2},-3.2) {\usebox\cTe};

\node at ({\xa+\wa/2},-4.4) {\yes{trained} \quad \no{rich handoff}};
\node at ({\xb+\wb/2},-4.4) {\no{trained} \qquad \yes{rich handoff}};
\node at ({\xc+\wc/2},-4.4) {\yes{trained} \qquad \yes{rich handoff}};
\end{tikzpicture}
\end{scaledfigure}
  \caption{\textbf{Experience-Handoff RL.} (a) Outcome-only RL compresses each trial into a scalar reward. (b) A harness runs several calls to the same model ($c_i$s; in \ace{}, actor, reflector, and curator) and preserves experience as a model-written context, such as a playbook, handed to the next attempt (blue), but the model never learns to write better playbooks. (c) Experience-handoff RL trains the model for the harness it runs in, from task outcomes alone. For example, \acerl{} rewards each handoff artifact for what it does to a future rollout: (i) each new context is scored by a fresh rollout, and (ii) each role is compared within its own group (stacked boxes), which lowers the variance of its credit.}
  \label{fig:overview}
\end{figure*}

\section{Introduction}
\label{sec:intro}

An agent that learns from its own trials needs \emph{experience handoff}: what one attempt revealed about the environment has to reach the attempts that follow. There are two ways to hand off experience. The first is through scalar rewards that compress the trial into a single number, which the learner absorbs into its weights. The second is through rich experience artifacts, such as reflections, notes, or playbooks, that preserve the context of the trial and are passed to future rollouts in the model's input.

Reinforcement learning (RL) is the standard instance of the first way. During training, the agent hands off experience to its future self only through the policy update: each rollout is scored, the score becomes an advantage, and the gradient step shifts the policy toward rewarded behaviour \citep{williams1992reinforce,sutton2018reinforcement,ppo2017}. For language models, reinforcement learning with verifiable rewards (RLVR) reliably improves the \emph{actor} across model families and scales \citep{deepseekmath2024,guo2025deepseek,dapo2025}. However, outcome-only RL compresses a whole trial into one number, discarding the error logs, tool outputs, and discovered constraints that made the trial informative. The agent sees this environment feedback while it acts, but it cannot hand any of it off to future rollouts except through an abstract scalar, and RL offers no recipe for the reflection and note-writing behaviours that would carry it forward.

The second way is through \emph{harnesses}: scaffolds that decide when the model acts, what it sees, and what it carries from one attempt to the next. Harnesses keep experience as text, such as verbal reflections \citep{shinn2024reflexion}, skill libraries and reusable workflows \citep{wang2023voyager,awm2024}, cheatsheets and playbooks \citep{suzgun2026dynamic,ace2025}, evolved prompts \citep{gepa2025}, or archives of evaluated programs in test-time discovery loops \citep{alphaevolve2025}. With a capable frontier model inside, such scaffolds exhibit inference-time \emph{recursive self-improvement} (RSI): a rollout fails, the model reflects on the trace, rewrites its own notes or context, and the next rollout, conditioned on those notes, does better. Throughout this paper, RSI refers only to this loop, in which a model improves its own future rollouts by rewriting the context it will inherit, under fixed weights and a fixed harness.\footnote{We focus on this form because it is a clean setting for the second kind of experience handoff: the only thing passed from one attempt to the next is an artifact the model wrote. The use of RSI in the field more broadly also covers agents that rewrite their own code or scaffold and models that update their own weights; see Appendix~\ref{sec:related}.} However, the harness only preserves experience; the quality of what it hands off is bounded by the frozen model that writes it. The calls that reflect and edit are never trained for the effect of their text on future rollouts, and methods that optimize the harness itself \citep{gepa2025,metaharness2026} change the prompts and code around the model rather than the model's ability to write useful experience (Figure~\ref{fig:motivation}).

Each way therefore has what the other lacks: RL updates weights from a verifiable signal but hands off only a scalar, and harnesses hand off rich experience but never learn to write it. This paper asks:
\begin{center}
\emph{Can a model learn, from task outcomes alone, both to act and to write\\ the experience that its future self will inherit?}
\end{center}

We refer to this kind of training as \emph{experience-handoff RL} (\ehrl{}). As a testbed we fix the \ace{} harness \citep{ace2025} and train one model for its three roles: the \emph{actor} that solves the task, the \emph{reflector} that extracts insights from a trajectory, and the \emph{curator} that edits a task-local \emph{playbook} for the next attempt. One property makes this training work: \textbf{the final playbook is verifiable}. Handing a candidate playbook to a fresh actor rollout and reading the task verifier scores it without a learned reward model, following the success behind RLVR. Our method, \acerl{}, pairs this reward with \textbf{a comparison group for every role}: each reflection is compared with other reflections of the same trajectory and each edit with other edits of the same insight, instead of every output in a chain sharing the chain's final reward.

At equal training compute on Mastermind, MBPP, AppWorld, and MiniSWEBench, \acerl{}'s held-out reward in the harness is above that of baseline methods, including actor-only RL without the harness and end-to-end training through the harness with a learned critic. We also show \acerl{} performs better than the actor-only \grpo{} checkpoint in the same harness, which simulates the common practice of running a model inside a hand-built harness that it was never trained for. We also show that the lead of RSI@$K$ over \grpo{}'s pass@$K$ maintains up to $K=8$, proposing that the helpfulness of the playbook adaptation extrapolates beyond the training cutoff ($K=3$). Finallly, we prove that training objective of \acerl{} actually makes helpful playbook rather than just adapting the actor component.
\section{Experience-Handoff Reinforcement Learning}
\label{sec:ehrl}

This section defines the harness and states the learning problem: the model is trained for the outcome of the harness it runs in, not only for the outcome of its actor.

\subsection{A Fixed Task-Local Harness}
\label{sec:harness}

The harness in this paper is a task-local variant of Agentic Context Engineering (\ace{}) \citep{ace2025}. One model $\pi_\theta$ serves three roles under fixed role instructions $\iota_{\mathrm{act}},\iota_{\mathrm{ref}},\iota_{\mathrm{cur}}$. Let $q$ be a task instance and $P$ its current \emph{playbook}, a short list of natural-language tips that is prepended to the actor's input. One \emph{handoff iteration} consists of
\begin{align}
  \tau &\sim \pi_\theta(\cdot\mid\iota_{\mathrm{act}}(q,P)), &&\text{(act)}\label{eq:act}\\
  z &\sim \pi_\theta(\cdot\mid\iota_{\mathrm{ref}}(\tau)), &&\text{(reflect)}\label{eq:reflect}\\
  u &\sim \pi_\theta(\cdot\mid\iota_{\mathrm{cur}}(P,z)),\qquad P'=\mathrm{Apply}(P,u). &&\text{(curate)}\label{eq:curate}
\end{align}
The actor trajectory $\tau$ interleaves generated actions with environment observations, and $R_q(\tau)\in[0,1]$ is its reward from the benchmark's task verifier. The reflector reads the trajectory, including its reward and whatever the environment revealed (test failures, tool errors, termination reason) and writes an \emph{insight} $z$. The curator reads the incumbent playbook and the insight and emits structured edit operations $u$ (add, update, or remove a tip); $\mathrm{Apply}$ executes valid operations to produce a candidate playbook $P'$. The playbook is local to $q$: a fresh environment episode is started for every actor call, and the only state carried across attempts on the same task is $P$. Prompts for all roles are listed in Appendix~\ref{app:prompts}.

\paragraph{Inference-time recursive self-improvement.}
At test time the harness runs $K$ handoff iterations per task with frozen weights, and each iteration's curator edit is applied to the playbook that the next attempt reads; no extra rollout scores the edit. We call this protocol \emph{RSI@$K$}: a task counts as solved if any of the $K$ iterations solves it. Section~\ref{sec:eval-protocols} specifies the protocol and its rollout accounting. 

\subsection{The Learning Problem}
\label{sec:problem}

\begin{figure*}[t]
  \centering
\definecolor{artif}{HTML}{2F6FB3}   %
\definecolor{mut}{HTML}{8A8A8A}
\definecolor{bad}{HTML}{B23A2E}
\definecolor{good}{HTML}{2E7D32}
\definecolor{cbase}{HTML}{9A9A9A}
\definecolor{cgrpo}{HTML}{4C72B0}
\definecolor{cace}{HTML}{D9730D}
\newcommand{\code}[1]{{\ttfamily #1}}
\newcommand{\att}[3]{\textbf{Attempt #1}\enspace#2\hfill{\color{black!55}#3 turns}\\}
\newcommand{\outcome}[1]{\ifnum#1=1 \textcolor{good}{\textbf{\checkmark\,solved}}\else\textcolor{bad}{\textbf{$\times$\,failed}}\fi}
\tikzset{
  font=\tiny,
  act/.style={draw=black!45,rounded corners=2pt,fill=white,align=left,
    text width=35.5mm,inner sep=1.6pt,minimum height=8.2mm},
  pb/.style={draw=artif,fill=artif!8,align=left,text width=31.5mm,inner sep=2pt,
    minimum height=5.6mm,text=artif!55!black},
  pbnone/.style={pb,draw=mut,fill=gray!6,text=black!55,dashed},
  hand/.style={-{Stealth[length=1.4mm]},thick,draw=artif},
  colhdr/.style={font=\scriptsize\bfseries,anchor=south},
}
\begin{scaledfigure}[1pt]
\begin{tikzpicture}

\begin{axis}[
  name=plot, at={(0,-0.5)}, anchor=north west,
  width=5.4cm, height=4.6cm,
  xmin=0.6, xmax=8.4, ymin=0.1, ymax=0.9,
  xtick={1,...,8}, ytick={0.2,0.4,0.6,0.8},
  yticklabel style={/pgf/number format/fixed,/pgf/number format/fixed zerofill,/pgf/number format/precision=1},
  tick label style={font=\scriptsize}, label style={font=\scriptsize},
  xlabel={attempt $k$ (20 turns each)}, ylabel={solved by attempt $k$ (RSI@$k$)},
  ylabel shift=1pt, xlabel shift=0pt,
  axis lines=left, tick align=outside, ymajorgrids, grid style={black!8},
  every axis plot/.append style={line width=1.0pt, line join=round, mark size=1.2pt},
  legend style={font=\scriptsize, draw=none, fill=none, at={(0.5,-0.36)}, anchor=north,
    legend columns=1, row sep=-1.5pt, inner sep=0pt},
  legend image post style={xscale=0.6},
  legend cell align=left, clip=false,
]
  \draw[cgrpo!70, densely dashed, line width=0.8pt] (axis cs:0.6,0.513) -- (axis cs:8.4,0.513)
    node[pos=1, anchor=north east, font=\tiny, text=cgrpo, inner sep=1.5pt] {GRPO pass@1};
  \draw[cgrpo!70, densely dashed, line width=0.8pt] (axis cs:0.6,0.684) -- (axis cs:8.4,0.684)
    node[pos=0, anchor=south west, font=\tiny, text=cgrpo, inner sep=1.5pt, xshift=1pt] {GRPO pass@8};
  \addplot[cbase, mark=*] table[x=k,y=base_rsi,col sep=comma] {data/campaign_scaling_appworld.csv};
  \addlegendentry{Base model, RSI}
  \addplot[cgrpo, mark=*] table[x=k,y=grpo_rsi,col sep=comma] {data/campaign_scaling_appworld.csv};
  \addlegendentry{Actor-only RL (GRPO), RSI}
  \addplot[cace, mark=*] table[x=k,y=ace_rl_rsi,col sep=comma] {data/campaign_scaling_appworld.csv};
  \addlegendentry{\textbf{ACE-RL (ours), RSI}}
\end{axis}

\def\xg{6.7} \def\xa{10.72}   %
\node[anchor=north west,align=left,text width=77mm,inner sep=0pt,font=\scriptsize] (task) at (4.8,0.3)
  {\textbf{(b)} Held-out task \emph{``Send a reminder on Venmo for all my payment requests to my roommates which have not been approved or denied for 50 or more days''}};
\node[colhdr,text=cgrpo,anchor=north] (hg) at (\xg,-0.5) {Actor-only RL (GRPO)};
\node[colhdr,text=cace,anchor=north]  (ha) at (\xa,-0.5) {ACE-RL (ours)};

\node[act,anchor=north] (g1) at (\xg,-0.96) {\att{1}{\outcome{0}}{20}Reads requests the user received, finds none to roommates, and submits.};
\node[pbnone,below=1.0mm of g1] (g12) {Edits tip [1] of an empty playbook: rejected; playbook stays empty};
\node[act,below=1.0mm of g12] (g2) {\att{2}{\outcome{0}}{20}Filters on guessed roommate names; finds no pending request.};
\node[pbnone,below=1.0mm of g2] (g23) {Again edits a missing tip: rejected; playbook stays empty};
\node[act,below=1.0mm of g23] (g3) {\att{3}{\outcome{0}}{16}Reminders on received requests are refused; submits.};

\node[act,anchor=north] (a1) at (\xa,-0.96) {\att{1}{\outcome{0}}{15}Never identifies the roommates; reminds all 41 old requests it sent.};
\node[pb,below=1.0mm of a1] (a12) {+[1] ``Verify roommates' identities using supervisor or phone app data \dots''};
\node[act,below=1.0mm of a12] (a2) {\att{2}{\outcome{0}}{17}Cannot log into the phone app; guesses roommates from payment history.};
\node[pb,below=1.0mm of a2] (a23) {+[2] ``Calculate days pending using \code{created\_at} and a verified current date \dots''};
\node[act,below=1.0mm of a23] (a3) {\att{3}{\outcome{1}}{15}Finds the roommates and today's date in the phone app; reminds the 6 old requests.};

\node[anchor=north west,inner sep=0pt,font=\scriptsize] at (-1.0,0.3) {\textbf{(a)} RSI over 8 attempts (AppWorld)};
\foreach \p/\q in {g1/g12,g12/g2,g2/g23,g23/g3,a1/a12,a12/a2,a2/a23,a23/a3}
  \draw[hand] ([xshift=-17.2mm]\p.south) -- ([xshift=-17.2mm]\q.north);

\end{tikzpicture}
\end{scaledfigure}
  \caption{Rollout of Qwen3-4B-Instruct-2507 (or checkpoints trained from it) on AppWorld, with eight consecutive attempts per held-out task in the \ace{} harness. \grpo{} and \acerl{} checkpoints are those of Table~\ref{tab:main-results}, within the same training compute. \textbf{(a)} Fraction of tasks solved by attempt $k$ (RSI@$k$). Dashed lines: the same \grpo{} checkpoint without the harness, with one sample (pass@1) and the best of eight independent samples (pass@8). \textbf{(b)} One held-out task through the first three attempts of each checkpoint, with the playbook change its curator makes after each attempt.}
  \label{fig:motivation}
\end{figure*}
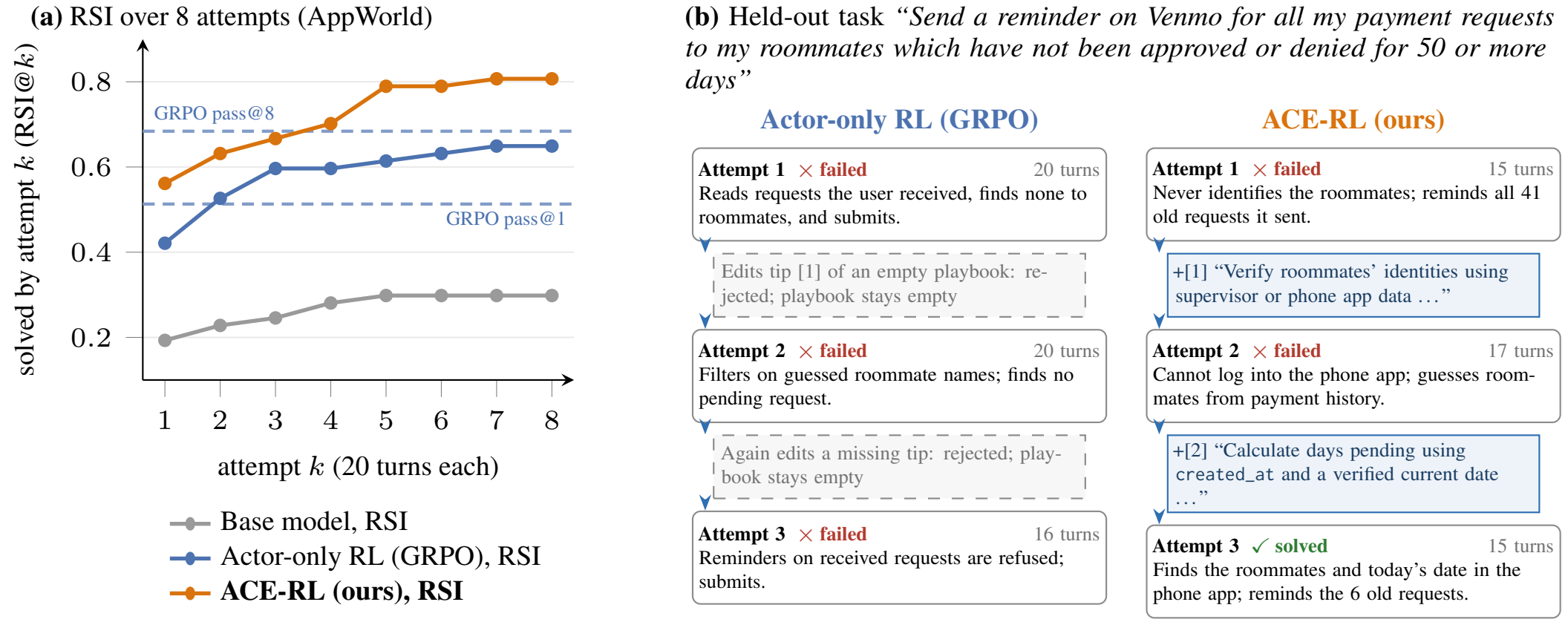

Outcome-only agentic RL optimizes the actor alone, maximizing $J_{\mathrm{act}}(\theta)=\mathbb E_{q}\,\mathbb E_{\tau\sim\pi_\theta(\cdot\mid\iota_{\mathrm{act}}(q,P_0))}[R_q(\tau)]$, where $P_0$ is the empty (or initial) playbook. A model trained this way is then deployed inside a harness, on the expectation that a better actor also makes a better model inside the harness. When the harness does not help, the usual protocol is to change the harness rather than the model: prompts, control flow, and memory formats are engineered by hand or searched over around a fixed model \citep{khattab2023dspy,gepa2025,metaharness2026}, and the result may or may not suit the model it wraps. In neither case is the model trained for what its reflections and edits do to later attempts.

We instead train the model for the harness it runs in. Let $\omega\sim H[\pi_\theta](q)$ denote a complete execution trace of harness $H$ on task $q$. We define the EHRL objective as
\begin{equation}
  \max_\theta\quad J_H(\theta)
  :=\mathbb E_{q}\,\mathbb E_{\omega\sim H[\pi_\theta](q)}
  \bigl[R_q(\omega)\bigr].
  \label{eq:harness-objective}
\end{equation}
The harness fixes the prompts, tools, and control flow; the model is trained for the outcome of their combined execution, so every call it makes, not only the actor, is an object of training. For the \ace{} handoff in Eqs.~\eqref{eq:act}--\eqref{eq:curate}, we can specify this objective to the actor's outcome \emph{after} the playbook update:
\begin{equation}
  \max_\theta\quad J_{\mathrm{ACE}}(\theta)
  :=\mathbb E_{(q,P)\sim\mu}\,
  \mathbb E_{\tau,z,u,\tilde\tau\sim\pi_\theta}
  \bigl[R_q(\tilde\tau)\bigr].
  \label{eq:ehrl-objective}
\end{equation}
Here $\tilde\tau\sim\pi_\theta(\cdot\mid\iota_{\mathrm{act}}(q,P'))$ is a fresh attempt under the proposed playbook $P'$, and $\mu$ is the distribution of pairs $(q,P)$ that training visits, where $P$ is the playbook that task $q$ inherited from its earlier encounters ($P_0$ at the start and after each reset). The reward of Eq.~\eqref{eq:ehrl-objective} is paid only for $\tilde\tau$, yet it depends on all four samples of the chain: the seed $\tau$ supplies the evidence, $z$ what is extracted from it, $u$ how the playbook changes, and $\tilde\tau$ whether the actor uses the change. A single $R_q(\tilde\tau)$ does not say which of the four was at fault; Section~\ref{sec:method} gives each its own reward.

Figure~\ref{fig:motivation} shows why this objective is needed, as a preview of Section~\ref{sec:results} that simulates how a harness is used in practice. First, the base model is placed in the harness as it is: it gains little from its own playbook edits. Second, only the actor is trained, with \grpo{} for $J_{\mathrm{act}}$, and the checkpoint is deployed in the same harness. It is better than the base model, but its training objective is not the harness's: after eight attempts it ends below what the same checkpoint reaches by sampling eight times independently without the harness. Its playbook edits do not adapt it to the task; trying again would do better. Third, the model trained for Eq.~\eqref{eq:ehrl-objective} keeps improving across attempts and ends above that independent-sampling reference; on the task in panel~(b), \grpo{}'s curator never changes the playbook, while \acerl{}'s edits lead it to the data it was missing and the third attempt succeeds. Adapting through the handoff is a behavior of the model in every role, and training the actor alone does not produce it.
\section{ACE-RL}
\label{sec:method}

\acerl{} trains the actor, reflector, and curator together with a \grpo{}-style update for Eq.~\eqref{eq:ehrl-objective}. It has two parts: a verifiable reward for every handoff output, obtained by executing the playbook it leads to (Section~\ref{sec:candidate-learning}), and a sampling tree that gives each role its own comparison group within \grpo{}'s actor-rollout budget (Section~\ref{sec:sparse-tree}). Appendix~\ref{app:credit-derivation} derives it from the full sampling tree.

\subsection{Verifiable Rewards for Handoff Artifacts}
\label{sec:candidate-learning}

Let $P$ be the current playbook of task $q$ and $\pi_{\theta_{\mathrm{old}}}$ the collection policy shared by all roles. A curator output $u$ yields a candidate $P'=\mathrm{Apply}(P,u)$. We define its \emph{handoff value} by executing it:
\begin{equation}
  \tilde\tau\sim\pi_{\theta_{\mathrm{old}}}(\cdot\mid\iota_{\mathrm{act}}(q,P')),\qquad
  \widehat V(P')=R_q(\tilde\tau).
  \label{eq:candidate-value}
\end{equation}
This is a single rollout estimate of the conditional downstream value in Eq.~\eqref{eq:ehrl-objective}. A curator output receives the value of its playbook; a reflection receives the mean value of the admissible playbooks curated from it; every actor rollout, whether sampled under the incumbent playbook or under a candidate, receives its own task reward:
\begin{equation}
  r^{\mathrm{cur}}(u)=\widehat V(\mathrm{Apply}(P,u)),\qquad
  r^{\mathrm{ref}}(z)=\frac{1}{|\mathcal C(z)|}\sum_{u\in\mathcal C(z)}\widehat V(\mathrm{Apply}(P,u)),\qquad
  r^{\mathrm{act}}(\tau)=R_q(\tau),
  \label{eq:component-rewards}
\end{equation}
where $\mathcal C(z)$ is the set of admissible sampled edits derived from $z$, with $r^{\mathrm{ref}}(z)=0$ when this set is empty; inadmissible edits, including edits that leave the playbook unchanged, receive reward $0$ (Appendix~\ref{app:pipeline}). Because actor rollouts are sampled under both the incumbent playbook and the candidates, the actor learns to read both a retained handoff and a fresh one.

\subsection{A Comparison Group for Every Role}
\label{sec:sparse-tree}

\begin{figure}[t]
  \centering
\definecolor{actc}{HTML}{E7E2F3}
\definecolor{actd}{HTML}{5B4B8A}
\definecolor{refc}{HTML}{F6ECC6}
\definecolor{refd}{HTML}{8A6A10}
\definecolor{curc}{HTML}{DFEEDA}
\definecolor{curd}{HTML}{3B6E2E}
\definecolor{artif}{HTML}{2F6FB3}
\definecolor{scalar}{HTML}{D9730D}
\tikzset{
  font=\scriptsize,
  nd/.style={draw=black!65,rounded corners=2pt,fill=white,
    minimum height=3.3mm,minimum width=9mm,inner sep=1.2pt},
  seed/.style={nd,very thick},
  ctx/.style={draw=artif,fill=artif!8,text=artif!70!black,
    minimum height=3.3mm,minimum width=9mm,inner sep=1.2pt},
  val/.style={draw=black!65,fill=white,minimum height=3.3mm,
    minimum width=9mm,inner sep=1.2pt},
  flow/.style={-{Stealth[length=1.2mm]},draw=black!60},
  hand/.style={-{Stealth[length=1.2mm]},semithick,draw=artif},
  grp/.style={rounded corners=3pt,inner sep=1.8pt},
  hdr/.style={font=\scriptsize\bfseries},
  sub/.style={font=\tiny,text=black!60},
  card/.style={rounded corners=2.5pt,minimum height=6.8mm,
    minimum width=30mm,align=center,inner sep=2pt},
  grad/.style={-{Stealth[length=1.4mm]},semithick,draw=scalar},
}
\newcommand{\vdt}[2]{\path ($(#1,{#2})$) coordinate (vdc); \fill[black!55] ($(vdc)+(0,0.07)$) circle (0.35pt) (vdc) circle (0.35pt) ($(vdc)-(0,0.07)$) circle (0.35pt);}
\begin{scaledfigure}[2pt]
\begin{tikzpicture}
\def\xin{0.6}\def\xa{2.45}\def\xr{4.55}\def\xc{6.85}\def\xp{9.25}\def\xv{11.4}
\def\yt{0}\def\ydu{-0.35}\def\yha{-0.72}\def\yh{-1.06}\def\yhc{-1.40}
\def\ydl{-1.77}\def\ybot{-2.12}   %

\node[hdr,anchor=west] at (-0.1,1.16) {(a) Sample one comparison group per role, then execute every candidate};
\foreach \x/\h/\s in {\xa/Act/{$N$ rollouts\\of $(q,P)$},
                      \xr/Reflect/{$M$ reflections of\\one seed $\tau_{j^\star}$},
                      \xc/Curate/{$C$ edits of $z_h$;\\one of each other},
                      \xp/Apply/{candidate\\playbooks},
                      \xv/Evaluate/{one rollout per candidate,\\joins the actor group}}{
  \node[hdr] at (\x,0.82) {\h};
  \node[sub,align=center,inner sep=0pt] at (\x,0.45) {\s};
}

\node[ctx,align=center] (in) at (\xin,\yh) {Task $q$\\Playbook $P$};
\node[nd] (a1) at (\xa,\yt) {$\tau_1$};
\vdt{\xa}{(\yt+\yh)/2}
\node[seed] (as) at (\xa,\yh) {$\tau_{j^\star}$};
\vdt{\xa}{(\yh+\ybot)/2}
\node[nd] (an) at (\xa,\ybot) {$\tau_N$};
\foreach \n in {a1,as,an} \draw[flow] (in.east) -- (\n.west);

\node[nd] (z1) at (\xr,\yt) {$z_1$};
\vdt{\xr}{\ydu}
\node[seed] (zh) at (\xr,\yh) {$z_h$};
\vdt{\xr}{\ydl}
\node[nd] (zM) at (\xr,\ybot) {$z_M$};
\foreach \i in {1,h,M} \draw[flow] (as.east) -- (z\i.west);

\foreach \id/\src/\y in {11/z1/\yt,h1/zh/\yha,hC/zh/\yhc,M1/zM/\ybot}{
  \node[nd]  (u\id) at (\xc,\y) {$u_{\id}$};
  \node[ctx] (p\id) at (\xp,\y) {$P'_{\id}$};
  \node[val] (v\id) at (\xv,\y) {$\tilde\tau_{\id}$};
  \draw[flow] (\src.east) -- (u\id.west);
  \draw[hand] (u\id.east) -- (p\id.west);
  \draw[flow] (p\id.east) -- (v\id.west);
}
\foreach \y in {\ydu,\yh,\ydl}{\foreach \x in {\xc,\xp,\xv} {\vdt{\x}{\y}}}
\draw[flow,black!35] (zh.east) -- ($(\xc,\yh)+(-0.5,0)$);

\begin{scope}[on background layer]
  \node[grp,fill=actc,fit=(a1)(an)] (ga) {};
  \node[grp,fill=refc,fit=(z1)(zM)] (gr) {};
  \node[grp,fill=curc,fit=(uh1)(uhC)] (gc) {};
  \node[grp,fill=actc,fit=(v11)(vM1)] (gv) {};
\end{scope}

\def\yb{-2.72}
\draw[black!25] (-0.1,-2.46) -- (13.2,-2.46);
\node[hdr,anchor=west] at (-0.1,\yb) {(b) Reward each output, normalize within its group, update once};
\def\yk{-3.22}
\node[card,fill=actc] (ca) at (1.5,\yk)
  {\textcolor{actd}{\textbf{Actor group}}\\ Both $R_q(\tau_j)$ and $R_q(\tilde\tau_{ik})$};
\node[card,fill=refc] (cr) at (4.75,\yk)
  {\textcolor{refd}{\textbf{Reflector group}}\\$\operatorname{mean}_k\, R_q(\tilde\tau_{ik})$};
\node[card,fill=curc] (cc) at (8.0,\yk)
  {\textcolor{curd}{\textbf{Curator group}}\\$R_q(\tilde\tau_{hk})$};
\node[card,draw=scalar,fill=scalar!6,minimum width=31mm] (up) at (11.55,\yk)
  {$A=(r-\mu_G)/(\sigma_G+\epsilon)$\\one joint policy update};
\node at ($(ca.east)!0.5!(cr.west)$) {$+$};
\node at ($(cr.east)!0.5!(cc.west)$) {$+$};
\draw[grad] (cc.east) -- (up.west);
\end{tikzpicture}
\end{scaledfigure}
  \caption{\textbf{Sampling tree of \acerl{}} with group sizes $N+M+C-1$, $M$, $C$ for actor, reflector, and curator. (a) Shaded boxes are comparison groups: $M$ reflections of a uniformly chosen seed $\tau_{j^\star}$ and $C$ edits of a uniformly chosen reflection $z_h$ share an input; every other reflection receives one edit. Each edit is applied to a copy of $P$ and executed once, and these rollouts join the $N$ seed rollouts of $(q,P)$ in the actor group (purple). (b) Candidate values reward the edit that produced them and, averaged, its reflection; advantages are normalized within each group and all roles share one update.}
  \label{fig:sampling-tree}
\end{figure}
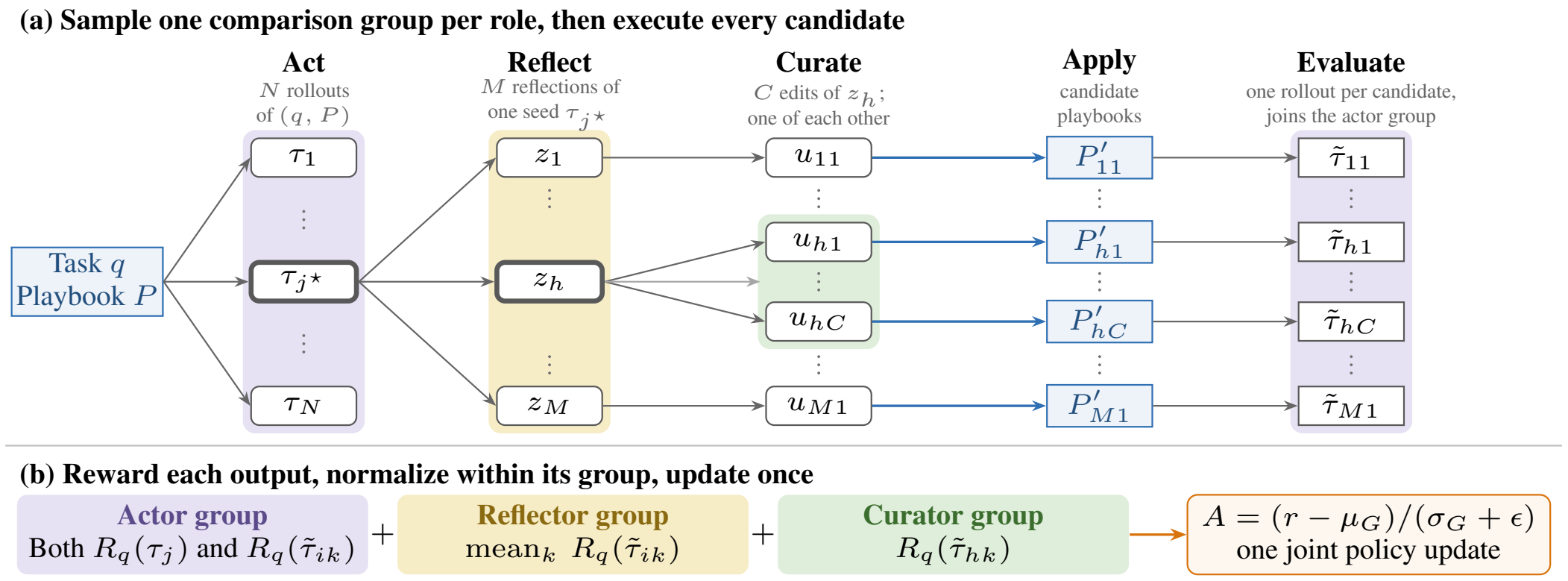

\begin{algorithm}[t]
  \caption{\acerl{}: one training step.}
  \label{alg:acerl-main}
  \begin{algorithmic}[1]
    \Require task batch $\mathcal Q$; policy $\theta$; per-task playbooks $\{P_q\}$ and encounter counts $\{n_q\}$; initial playbook $P_0$; group sizes $(N,M,C)$; reset interval $F$
    \State $\mathcal B\gets\varnothing$;\quad $\theta_{\mathrm{old}}\gets\theta$
    \For{each task $q\in\mathcal Q$}
      \State $n_q\gets n_q+1$;\quad \textbf{if} $n_q\bmod F=0$ \textbf{then} $P_q\gets P_0$ \Comment{periodic reset}
      \State Sample $N$ seed rollouts $\tau_1{:}\tau_N\sim\pi_{\theta_{\mathrm{old}}}(\cdot\mid\iota_{\mathrm{act}}(q,P_q))$; rewards $R_q(\tau_j)$ \label{alg:actor-step}
      \State Pick seed $j^\star\sim\mathrm{Unif}[N]$; sample $M$ reflections $z_1{:}z_M$ of $\tau_{j^\star}$ \label{alg:reflect-step}
      \State If any reflections are usable, choose $z_h$ uniformly; sample $C$ edits of $z_h$ and one of each other usable reflection \label{alg:curate-step}
      \State Apply admissible edits to copies of $P_q$; score each candidate with one fresh rollout (Eq.~\ref{eq:candidate-value}) and add it to the actor group of $q$ \label{alg:evaluate-step}
      \State Assign component rewards (Eq.~\ref{eq:component-rewards}); update $P_q$ (Eq.~\ref{eq:selection-rule}) \label{alg:reward-step}
      \State Add actor, reflector, and curator outputs with rewards and group ids to $\mathcal B$
    \EndFor
    \State Filter groups (Appendix~\ref{app:pipeline}); normalize advantages within each role's group \label{alg:advantage-step}
    \State Update $\theta$ on $\mathcal B$ with the clipped token-level surrogate (Eq.~\ref{eq:shared-objective}) \label{alg:update-step}
  \end{algorithmic}
\end{algorithm}

Group-relative advantages need several outputs sampled from the same input. Algorithm~\ref{alg:acerl-main} and Figure~\ref{fig:sampling-tree} build these groups for task $q$ with playbook $P$: $N$ seed rollouts of $(q,P)$; $M$ reflections of one seed $\tau_{j^\star}$, $j^\star\sim\mathrm{Unif}[N]$; $C$ edits of one usable reflection $z_h$, chosen uniformly, and one edit of each other usable reflection; and one fresh rollout per admissible edit (Eq.~\ref{eq:candidate-value}), which joins the seed rollouts in the actor group.

Each output $i$ receives $A_i=(r_i-\mu_G)/(\sigma_G+\epsilon)$, where $\mu_G$ and $\sigma_G$ are the reward mean and standard deviation of its role's group $G$\footnote{An alternative without per-role groups compares reflections and edits across the chains of different actor rollouts; each reflection's advantage then also carries the quality of the trajectory it read (Appendix~\ref{app:delayed-joint}).}: actions against the other actor rollouts of $q$, reflections against the other reflections of $\tau_{j^\star}$, and edits against the other edits of $(P,z_h)$. A failed seed can therefore still yield a positively rewarded reflection. Groups with constant rewards are dropped, and the outputs of all three roles are optimized in one clipped update of the shared model (Eq.~\ref{eq:shared-objective}).

\paragraph{Playbook selection and reset.}
With $s=R_q(\tau_{j^\star})$ the seed's reward, the playbook that task $q$ carries into its next encounter is
\begin{equation}
  P^{\mathrm{next}}=
  \begin{cases}
    \arg\max_{P'}\ \widehat V(P') & \text{if } \exists P':\ \widehat V(P')\ge s,\\
    P & \text{otherwise,}
  \end{cases}
  \label{eq:selection-rule}
\end{equation}
and every $F$ encounters it is reset to $P_0$.

\section{Experimental Setup}
\label{sec:experiments}

We train Qwen3-0.6B in non-thinking mode \citep{qwen3_2025} on Mastermind and MBPP \citep{mbpp2021}, and Qwen3-4B-Instruct-2507 on AppWorld \citep{appworld2024} and MiniSWEBench \citep{sweagent2024,swebench2023}; Appendix~\ref{app:benchmarks} describes the tasks, splits, and reward design. Every actor episode has 20 turns (6 on MBPP) in training and evaluation. \acerl{}'s 8 actor rollouts per task are $N=4$ seed rollouts and up to four rollouts that score candidate playbooks: $M=2$ reflections of one seed, $C=3$ edits of one reflection and one edit of the other (With $N=4$, $M=2$, and $C=3$, a task yields at most four candidates, so the actor group holds at most eight rollouts, as in \grpo{}.). Each task's playbook is reset every $F=3$ encounters. Delayed-joint PPO samples one reflection and one edit per actor rollout. All methods share the harness code, verifiers, and optimizer settings, and Appendix~\ref{app:pipeline} lists the remaining training details.

\subsection{Compared Methods}
\label{sec:baselines}

\begin{table}[t]
  \caption{\textbf{Compared methods}, all trained with the same task verifier reward $R_q$ from 8 actor rollouts per task. Section~\ref{sec:results-compute} compares all of them at equal compute.}
  \label{tab:training-comparisons}
  \centering
  \small
  \setlength{\tabcolsep}{4pt}
  \begin{tabular}{lllll}
    \toprule
    Method & Trained roles & Advantage & Credit for handoff text & Evaluation \\
    \midrule
    PPO \citep{ppo2017} & actor & critic (GAE) & --- & pass@3 \\
    \grpo{} \citep{deepseekmath2024} & actor & group & --- & pass@3, RSI@3 \\
    ERL \citep{erl2026} & actor, prompt reviser & group & retry reward; internalized & pass@3 \\
    Delayed-joint PPO \citep{compactionrl2026} & actor, reflector, curator & critic (GAE) & next attempt of the same chain & RSI@3 \\
    \textbf{\acerl{} (ours)} & actor, reflector, curator & group per role & executed playbook (Eq.~\ref{eq:component-rewards}) & RSI@3 \\
    \bottomrule
  \end{tabular}
\end{table}

Table~\ref{tab:training-comparisons} lists the compared methods. For task $q$, let $\tau_1,\ldots,\tau_G$ be $G=8$ rollouts with rewards $R_i=R_q(\tau_i)$. Every method maximizes PPO's clipped surrogate \citep{ppo2017} with per-token advantages $A_{i,t}$; they differ in what is sampled and how $A_{i,t}$ is computed. All methods are matched in actor trajectories: each samples $G=8$ actor rollouts per task on 8 tasks per step (we also sample 8 parallel actor rollouts for critic learning baselines). For \acerl{} the 8 include the rollouts that score candidate playbooks, so its actor trains on exactly as many trajectories as \grpo{}'s.

\paragraph{Actor-only RL.}
\grpo{} \citep{deepseekmath2024} gives every token of $\tau_i$ the group-normalized advantage $A_{i,t}=(R_i-\bar R)/(\sigma_R+\epsilon)$, with $\bar R$, $\sigma_R$ the group's mean and standard deviation. PPO \citep{ppo2017,gae2015} trains a critic $V_\phi$ and uses GAE, $A_{i,t}=\sum_{l\ge0}(\gamma\lambda)^l\delta_{i,t+l}$ with $\delta_{i,t}=r_{i,t}+\gamma V_\phi(s_{i,t+1})-V_\phi(s_{i,t})$, where the only reward is $R_i$ at the last token ($\gamma=1$, $\lambda=0.95$); recent long-horizon agentic RL has returned to such critics \citep{sao2026,compactionrl2026}. Neither sees the harness in training, as when agents are post-trained before deployment in a scaffold; we evaluate \grpo{} without and inside the harness.

\paragraph{Reflect, then internalize.}
ERL \citep{erl2026} hands feedback off through the weights. After an unsuccessful attempt $\tau$ under prompt $x$, the model writes a revised prompt $x'$ from $\tau$, its reward, and the feedback, and retries, $\tau'\sim\pi_\theta(\cdot\mid x')$. Attempts, revisions (rewarded by $R_q(\tau')$), and retries get group advantages, and each retry is also trained under $x$ with advantage $\max(R_q(\tau'),0)$, distilling the improvement into the policy that sees only $x$. We port the released code, including its prompt-revision instruction (Appendix~\ref{app:prompts-erl}), and evaluate without a harness, as in its protocol.

\paragraph{Training through the harness.}
\emph{Delayed-joint PPO} adapts CompactionRL \citep{compactionrl2026} to \ace{}. Each actor rollout starts a chain $(\tau_i,z_i,u_i,\tilde\tau_i)$ of Eqs.~\eqref{eq:act}--\eqref{eq:curate} whose four segments all receive the next attempt's reward $R_q(\tilde\tau_i)$, turned into GAE advantages by a critic initialized from the policy, without group normalization or reward-based playbook selection (Appendix~\ref{app:pipeline}). This is the credit structure of methods that train a model whose own summaries become its future context \citep{compactionrl2026,composer22026}; CompactionRL's context replacement and cross-segment correction have no counterpart in \ace{}. \emph{\acerl{}} trains the same three roles but rewards each output by Eq.~\eqref{eq:component-rewards} and normalizes within its role's group (Section~\ref{sec:sparse-tree}).

\subsection{Evaluation Protocols}
\label{sec:eval-protocols}

Checkpoints are evaluated with frozen weights on held-out tasks at actor temperature $0.7$. Vanilla evaluation draws eight independent actor episodes per task without any harness, and pass@$K$ is the expected best reward of $K$ of them. \emph{Harnessed} evaluation runs eight sequential visits: each is one actor episode under the current playbook, followed by one reflection and one curator edit that is applied for the next visit, with no separate candidate-scoring rollout. RSI@$K$ is the best reward among the first $K$ visits; the reported test reward of both uses $K=3$. All comparisons are at equal training compute, the cumulative update FLOPs of actor (plus critic for PPO and delayed joint PPO); Appendix~\ref{app:compute} gives the compute budget and the checkpoint chosen for each run. Each method is trained once, and error bars are one standard error over held-out tasks.
\section{Results}
\label{sec:results}
\label{sec:results-main}

\newcommand{\pmci}[2]{#1{\tiny$\pm$#2}}
\begin{table}[t]
  \caption{\textbf{Test reward at equal training compute}: pass@3 without the harness (top) and RSI@3 in it (bottom), three actor episodes per task either way. Checkpoints as in Appendix~\ref{app:compute}; bold: column maximum.}
  \label{tab:main-results}
  \centering
  \small
  \setlength{\tabcolsep}{3.5pt}
  \begin{tabular}{llcccc}
    \toprule
    & & \multicolumn{2}{c}{Qwen3-0.6B} & \multicolumn{2}{c}{Qwen3-4B-Instruct} \\
    \cmidrule(lr){3-4}\cmidrule(lr){5-6}
    Method & Eval & Mastermind & MBPP & AppWorld & MiniSWEBench \\
    \midrule
    Base model & pass@3 & \pmci{0.431}{0.014} & \pmci{0.396}{0.041} & \pmci{0.252}{0.052} & \pmci{0.146}{0.029} \\
    PPO \citep{ppo2017} & pass@3 & \pmci{0.000}{0.000} & \pmci{0.715}{0.036} & \pmci{0.106}{0.035} & \pmci{0.000}{0.000} \\
    \grpo{} \citep{deepseekmath2024} & pass@3 & \pmci{0.686}{0.010} & \pmci{0.867}{0.027} & \pmci{0.624}{0.060} & \pmci{0.245}{0.040} \\
    ERL \citep{erl2026} & pass@3 & \pmci{0.615}{0.010} & \pmci{0.509}{0.043} & \pmci{0.418}{0.059} & \pmci{0.234}{0.038} \\
    \midrule
    Base model & RSI@3 & \pmci{0.527}{0.015} & \pmci{0.387}{0.045} & \pmci{0.246}{0.058} & \pmci{0.180}{0.039} \\
    \grpo{} \citep{deepseekmath2024} & RSI@3 & \pmci{0.739}{0.012} & \pmci{0.820}{0.033} & \pmci{0.597}{0.066} & \pmci{0.260}{0.044} \\
    Delayed-joint PPO \citep{compactionrl2026} & RSI@3 & \pmci{0.659}{0.011} & \pmci{0.500}{0.047} & \pmci{0.000}{0.000} & \pmci{0.010}{0.010} \\
    \textbf{\acerl{}} (ours) & RSI@3 & \textbf{\pmci{0.775}{0.014}} & \textbf{\pmci{0.987}{0.011}} & \textbf{\pmci{0.667}{0.063}} & \textbf{\pmci{0.340}{0.048}} \\
    \bottomrule
  \end{tabular}
\end{table}

\begin{figure*}[t]
  \centering
\resultsfigstyle
\begin{scaledfigure}[1pt]
\begin{tikzpicture}
\begin{groupplot}[group style={group size=4 by 1, horizontal sep=0.6cm}, panel, width=3.95cm, height=2.9cm]
\nextgroupplot[title={Mastermind}, xmin=0, xmax=1.8335, restrict x to domain=0:1.8335, xlabel={training EFLOPs}, ylabel={test reward}, legend to name=mainleg]
  \addlegendimage{m ppo}\addlegendentry{PPO}
  \addlegendimage{m grpo}\addlegendentry{\textsc{GRPO}}
  \addlegendimage{m erl}\addlegendentry{ERL}
  \addlegendimage{m jppo}\addlegendentry{Delayed-joint PPO}
  \addlegendimage{m ace}\addlegendentry{\textbf{\textsc{ACE-RL}}}
  \testcurve{m ppo}{cppo}{data/campaign_flops_mastermind_ppo.csv}{ppo}
  \testcurve{m grpo}{cgrpo}{data/campaign_flops_mastermind_grpo.csv}{grpo}
  \testcurve{m erl}{cerl}{data/campaign_flops_mastermind_erl.csv}{erl}
  \testcurve{m jppo}{cjppo}{data/campaign_flops_mastermind_joint_ppo.csv}{joint_ppo}
  \testcurve{m ace}{cace}{data/campaign_flops_mastermind_ace_rl.csv}{ace_rl}
\nextgroupplot[title={MBPP}, xmin=0, xmax=0.6838, restrict x to domain=0:0.6838, xlabel={training EFLOPs}]
  \testcurve{m ppo}{cppo}{data/campaign_flops_mbpp_ppo.csv}{ppo}
  \testcurve{m grpo}{cgrpo}{data/campaign_flops_mbpp_grpo.csv}{grpo}
  \testcurve{m erl}{cerl}{data/campaign_flops_mbpp_erl.csv}{erl}
  \testcurve{m jppo}{cjppo}{data/campaign_flops_mbpp_joint_ppo.csv}{joint_ppo}
  \testcurve{m ace}{cace}{data/campaign_flops_mbpp_ace_rl.csv}{ace_rl}
\nextgroupplot[title={AppWorld}, xmin=0, xmax=13.7373, restrict x to domain=0:13.7373, xlabel={training EFLOPs}]
  \testcurve{m ppo}{cppo}{data/campaign_flops_appworld_ppo.csv}{ppo}
  \testcurve{m grpo}{cgrpo}{data/campaign_flops_appworld_grpo.csv}{grpo}
  \testcurve{m erl}{cerl}{data/campaign_flops_appworld_erl.csv}{erl}
  \testcurve{m jppo}{cjppo}{data/campaign_flops_appworld_joint_ppo.csv}{joint_ppo}
  \testcurve{m ace}{cace}{data/campaign_flops_appworld_ace_rl.csv}{ace_rl}
\nextgroupplot[title={MiniSWEBench}, xmin=0, xmax=4.0096, restrict x to domain=0:4.0096, xlabel={training EFLOPs}]
  \testcurve{m ppo}{cppo}{data/campaign_flops_miniswe_ppo.csv}{ppo}
  \testcurve{m grpo}{cgrpo}{data/campaign_flops_miniswe_grpo.csv}{grpo}
  \testcurve{m erl}{cerl}{data/campaign_flops_miniswe_erl.csv}{erl}
  \testcurve{m jppo}{cjppo}{data/campaign_flops_miniswe_joint_ppo.csv}{joint_ppo}
  \testcurve{m ace}{cace}{data/campaign_flops_miniswe_ace_rl.csv}{ace_rl}
\end{groupplot}
\node[anchor=north] at ([yshift=-0.75cm]$(group c2r1.south)!0.5!(group c3r1.south)$) {\pgfplotslegendfromname{mainleg}};
\end{tikzpicture}
\end{scaledfigure}
  \caption{\textbf{Held-out test reward against training compute} (cumulative update FLOPs, including the critic for PPO-based methods): pass@3 for PPO, \grpo{}, and ERL; RSI@3 for delayed-joint PPO and \acerl{}. Curves end at the compute budget (Appendix~\ref{app:compute}). Each curve starts at the base model at 0 EFLOPs, evaluated the same way, except \grpo{} on MiniSWEBench; there \acerl{}'s steps 5, 15, 25, and 35 are from a second seed (Appendix~\ref{app:compute}). Bands: 1 SEM over tasks.}
  \label{fig:learning-curves}
\end{figure*}
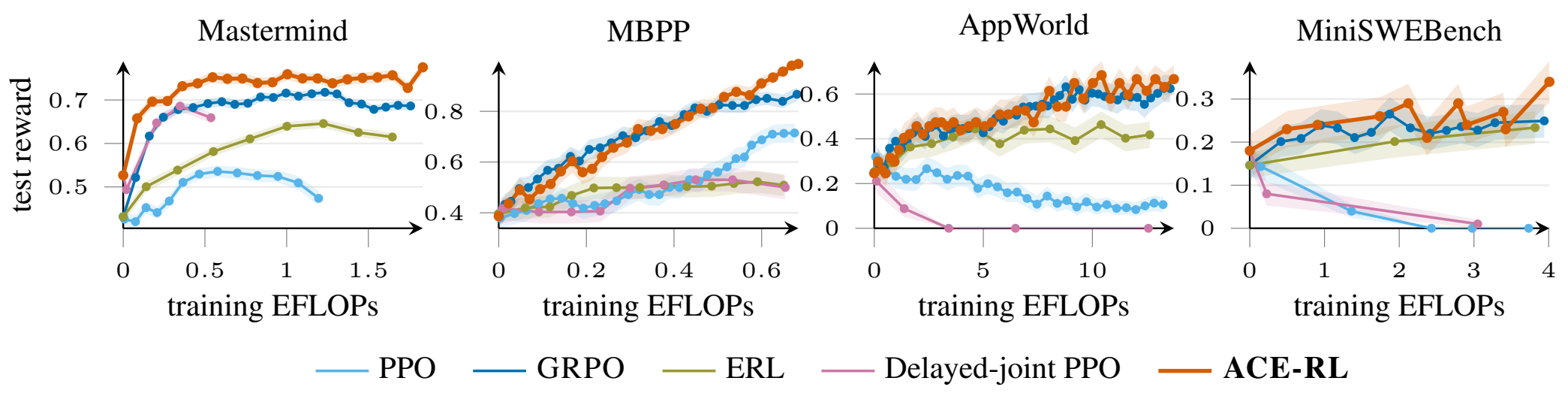

\subsection{Test Reward at Equal Training Compute}
\label{sec:results-compute}

Table~\ref{tab:main-results} and Figure~\ref{fig:learning-curves} compare all methods at \acerl{}'s training compute, with three actor episodes per task: independent ones without the harness for the actor-only methods and ERL (pass@3), and sequential visits in the harness for the methods trained through it (RSI@3). \acerl{} has the highest test reward on all four benchmarks.

\paragraph{Actor-only RL and ERL.}
Paired over tasks, \acerl{}'s RSI@3 is above \grpo{}'s pass@3 by $0.089\pm0.015$ on Mastermind, $0.120\pm0.026$ on MBPP, $0.042\pm0.041$ on AppWorld, and $0.095\pm0.034$ on MiniSWEBench ($\pm1$ SEM). ERL, which hands its revised prompts off only through the weights, is below \grpo{} on all four benchmarks. PPO is also below \grpo{} on all four, and its test reward reaches zero on Mastermind and MiniSWEBench.

\paragraph{Training through the harness.}
Delayed-joint PPO trains the same three roles as \acerl{}, but every output of a chain receives the chain's next-attempt reward, which a learned critic turns into advantages. This is the credit structure of recent systems that train a model whose own summaries become its future context \citep{compactionrl2026,composer22026}. It is below \acerl{} on all four benchmarks, and its test reward peaks early and then declines. On Mastermind its RSI@3 reaches $0.66$ at $0.54$ EFLOPs, against \acerl{}'s $0.74$ at $0.46$ EFLOPs, before its training reward falls below $0.05$ by $0.73$ EFLOPs. On MBPP it peaks at $0.53$, against \acerl{}'s $0.99$ at the budget. On AppWorld and MiniSWEBench it starts at $0.21$ and $0.08$ and falls to near zero (training reward in Appendix~\ref{app:training-reward}). Its critic's explained variance stays above $0.95$ late in each run (Appendix~\ref{app:critic}), so a poorly fit critic does not explain the decline. We cannot conclude that critic-based credit cannot train the handoff, but at least we can see that it's more prone to training collapse.

\subsection{Running the Actor-Only Model in the Harness}
\label{sec:results-actor}

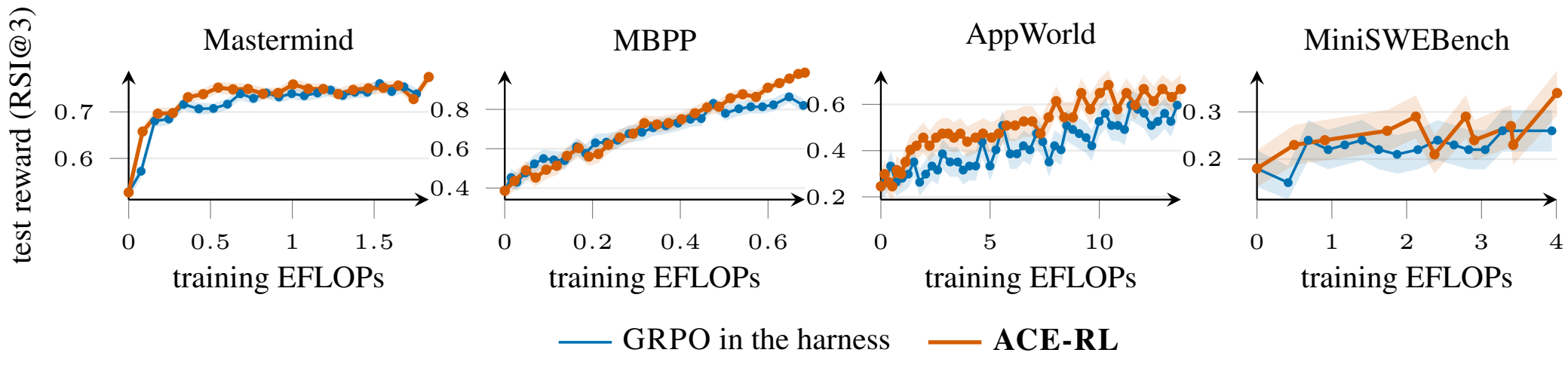
\begin{figure*}[t]
  \centering
\resultsfigstyle
\begin{scaledfigure}[1pt]
\begin{tikzpicture}
\begin{groupplot}[group style={group size=4 by 1, horizontal sep=0.6cm}, panel, width=3.95cm, height=2.6cm]
\nextgroupplot[title={Mastermind}, xmin=0, xmax=1.8335, restrict x to domain=0:1.8335, xlabel={training EFLOPs}, ylabel={test reward (RSI@3)}, legend to name=leg5a]
  \addlegendimage{m grpo}\addlegendentry{\textsc{GRPO} in the harness}
  \addlegendimage{m ace}\addlegendentry{\textbf{\textsc{ACE-RL}}}
  \testcurve{m grpo}{cgrpo}{data/campaign_flops_mastermind_grpo_rsi.csv}{grpo_rsi}
  \testcurve{m ace}{cace}{data/campaign_flops_mastermind_ace_rl.csv}{ace_rl}
\nextgroupplot[title={MBPP}, xmin=0, xmax=0.6838, restrict x to domain=0:0.6838, xlabel={training EFLOPs}]
  \testcurve{m grpo}{cgrpo}{data/campaign_flops_mbpp_grpo_rsi.csv}{grpo_rsi}
  \testcurve{m ace}{cace}{data/campaign_flops_mbpp_ace_rl.csv}{ace_rl}
\nextgroupplot[title={AppWorld}, xmin=0, xmax=13.7373, restrict x to domain=0:13.7373, xlabel={training EFLOPs}]
  \testcurve{m grpo}{cgrpo}{data/campaign_flops_appworld_grpo_rsi.csv}{grpo_rsi}
  \testcurve{m ace}{cace}{data/campaign_flops_appworld_ace_rl.csv}{ace_rl}
\nextgroupplot[title={MiniSWEBench}, xmin=0, xmax=4.0096, restrict x to domain=0:4.0096, xlabel={training EFLOPs}]
  \testcurve{m grpo}{cgrpo}{data/campaign_flops_miniswe_grpo_rsi.csv}{grpo_rsi}
  \testcurve{m ace}{cace}{data/campaign_flops_miniswe_ace_rl.csv}{ace_rl}
\end{groupplot}
\node[anchor=north] at ([yshift=-0.75cm]$(group c2r1.south)!0.5!(group c3r1.south)$) {\pgfplotslegendfromname{leg5a}};
\end{tikzpicture}
\end{scaledfigure}
  \caption{\textbf{Actor-only \grpo{} run in the harness}: test RSI@3 against training compute, with \acerl{}'s role prompts. Bands: 1 SEM over tasks.}
  \label{fig:grpo-rsi}
\end{figure*}

A common practice is to post-train the actor with outcome-only RL and then run it inside a hand-built harness it was not trained for. We show that the practice may be effective on some benchmark, but \acerl{} still leads the performance on matched training compute. Deploying the \grpo{} checkpoint in the \ace{} harness narrows its gap to \acerl{} on Mastermind, slightly on MiniSWEBench, and not on MBPP or AppWorld (Figure~\ref{fig:grpo-rsi}); within the budget it scores at least its own pass@1 without the harness at every Mastermind and MBPP checkpoint and at 39 of 47 on AppWorld. Though \acerl{} still gets the lead; At equal compute it reaches $0.74$ against \acerl{}'s $0.78$ on Mastermind (paired $+0.037\pm0.018$), $0.82$ against $0.99$ on MBPP ($+0.167\pm0.032$), $0.60$ against $0.67$ on AppWorld ($+0.070\pm0.055$), and $0.26$ against $0.34$ on MiniSWEBench ($+0.080\pm0.039$).

\begin{figure*}[t]
  \centering
\resultsfigstyle
\begin{scaledfigure}[1pt]
\begin{tikzpicture}
\begin{groupplot}[group style={group size=4 by 1, horizontal sep=0.6cm}, panel, width=3.95cm, height=2.6cm]
\nextgroupplot[title={Mastermind}, xmin=0.6, xmax=8.4, xtick={1,2,4,8}, legend to name=leg5b, xlabel={attempts $K$}, ylabel={best reward}]
  \scalecurve{m pass}{cgrpo}{data/campaign_scaling_mastermind.csv}{grpo_pass}
  \scalecurve{m grpo}{cgrpo}{data/campaign_scaling_mastermind.csv}{grpo_rsi}
  \scalecurve{m ace}{cace}{data/campaign_scaling_mastermind.csv}{ace_rl_rsi}
  \addlegendimage{m pass}\addlegendentry{\textsc{GRPO} pass@$K$}
  \addlegendimage{m grpo}\addlegendentry{\textsc{GRPO} RSI@$K$}
  \addlegendimage{m ace}\addlegendentry{\textbf{\textsc{ACE-RL}} RSI@$K$}
\nextgroupplot[title={MBPP}, xmin=0.6, xmax=8.4, xtick={1,2,4,8}, xlabel={attempts $K$}]
  \scalecurve{m pass}{cgrpo}{data/campaign_scaling_mbpp.csv}{grpo_pass}
  \scalecurve{m grpo}{cgrpo}{data/campaign_scaling_mbpp.csv}{grpo_rsi}
  \scalecurve{m ace}{cace}{data/campaign_scaling_mbpp.csv}{ace_rl_rsi}
\nextgroupplot[title={AppWorld}, xmin=0.6, xmax=8.4, xtick={1,2,4,8}, xlabel={attempts $K$}]
  \scalecurve{m pass}{cgrpo}{data/campaign_scaling_appworld.csv}{grpo_pass}
  \scalecurve{m grpo}{cgrpo}{data/campaign_scaling_appworld.csv}{grpo_rsi}
  \scalecurve{m ace}{cace}{data/campaign_scaling_appworld.csv}{ace_rl_rsi}
\nextgroupplot[title={MiniSWEBench}, xmin=0.6, xmax=8.4, xtick={1,2,4,8}, xlabel={attempts $K$}]
  \scalecurve{m pass}{cgrpo}{data/campaign_scaling_miniswe.csv}{grpo_pass}
  \scalecurve{m grpo}{cgrpo}{data/campaign_scaling_miniswe.csv}{grpo_rsi}
  \scalecurve{m ace}{cace}{data/campaign_scaling_miniswe.csv}{ace_rl_rsi}
\end{groupplot}
\node[anchor=north] at ([yshift=-0.75cm]$(group c2r1.south)!0.5!(group c3r1.south)$) {\pgfplotslegendfromname{leg5b}};
\end{tikzpicture}
\end{scaledfigure}
  \caption{\textbf{Best reward after $K$ attempts} at the checkpoints of Table~\ref{tab:main-results}. Dashed: independent \grpo{} pass@$K$; solid: sequential RSI@$K$ in the harness. Error bars: 1 SEM over tasks.}
  \label{fig:passk-rsik}
\end{figure*}
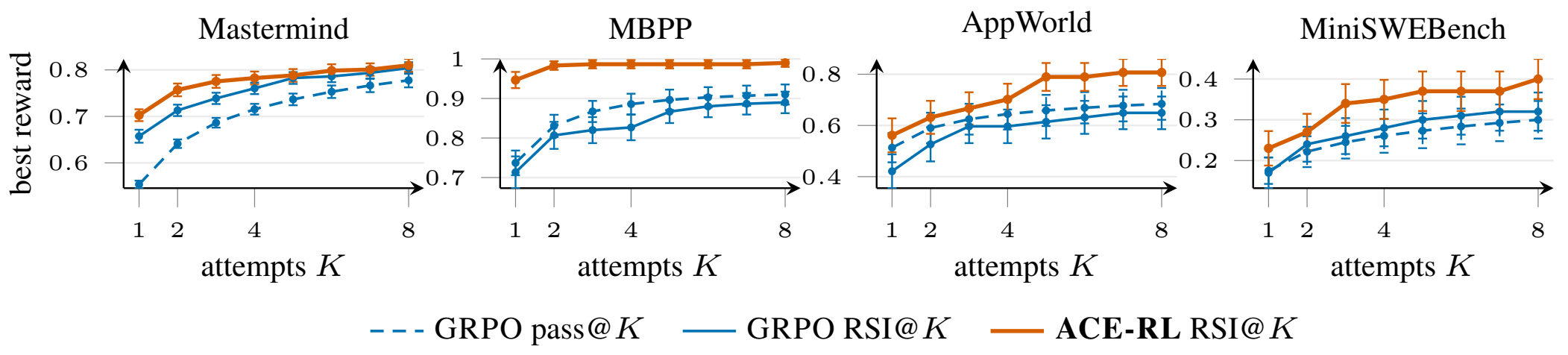

\subsection{Does the Gain Persist beyond the Training Horizon?}
\label{sec:results-scaling}

Training resets each task's playbook every $F=3$ encounters, so from the fourth visit on the actor reads playbooks with more rounds of edits than any seen in training. We compare RSI@$K$, the best of $K$ sequential visits, with \grpo{}'s pass@$K$, the best of $K$ independent rollouts \citep{largemonkeys2024}, which spends the same $K$ actor episodes without a handoff (Figure~\ref{fig:passk-rsik}). \acerl{}'s RSI@$K$ keeps rising past $K=3$ and stays above \grpo{}'s pass@$K$ at every $K$ on all four benchmarks, ending at $0.81$, $0.99$, $0.81$, and $0.40$ against $0.78$, $0.91$, $0.68$, and $0.30$ at $K=8$. From $K=3$ to $K=8$ it gains $0.03$, $0.00$ (MBPP is at its ceiling), $0.14$, and $0.06$, against $0.09$, $0.04$, $0.06$, and $0.06$ for pass@$K$, so the gap narrows on Mastermind, holds on MiniSWEBench, and widens on AppWorld.

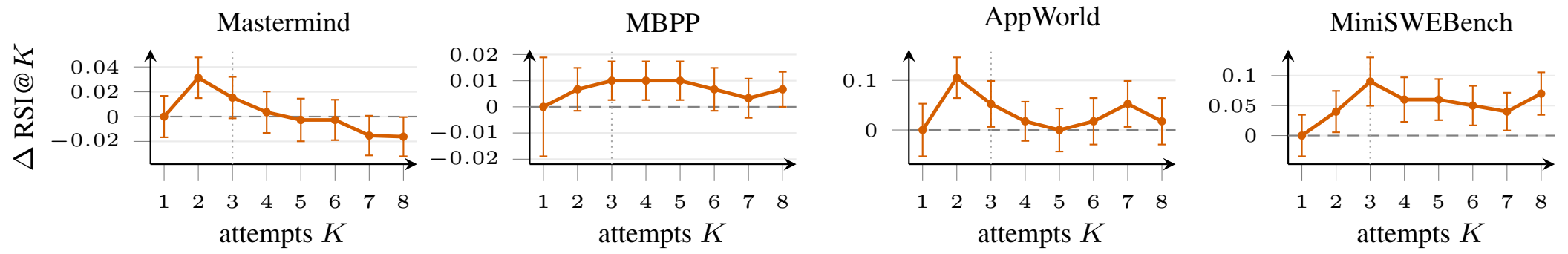
\begin{figure*}[t]
  \centering
\resultsfigstyle
\begin{scaledfigure}[1pt]
\begin{tikzpicture}
\begin{groupplot}[group style={group size=4 by 1, horizontal sep=1.0cm}, panel, width=3.95cm, height=2.6cm, xmin=0.6, xmax=8.4, xtick={1,2,3,4,5,6,7,8}, scaled y ticks=false, enlarge y limits=0.08, xlabel={attempts $K$}, yticklabel style={/pgf/number format/fixed,/pgf/number format/precision=2}]
\nextgroupplot[title={Mastermind}, ylabel={$\Delta$ RSI@$K$}]
  \addplot[forget plot, black!50, densely dashed, line width=0.4pt] coordinates {(0.6,0) (8.4,0)};
  \draw[black!35, densely dotted, line width=0.4pt] ({axis cs:3,0}|-{rel axis cs:0,0}) -- ({axis cs:3,0}|-{rel axis cs:0,1});
  \scalecurve{m ace}{cace}{data/campaign_control.csv}{mastermind}
  \addplot[draw=none, forget plot] table[x=k, y expr=\thisrow{mastermind}+\thisrow{mastermind_ci}, col sep=comma] {data/campaign_control.csv};
  \addplot[draw=none, forget plot] table[x=k, y expr=\thisrow{mastermind}-\thisrow{mastermind_ci}, col sep=comma] {data/campaign_control.csv};
\nextgroupplot[title={MBPP}]
  \addplot[forget plot, black!50, densely dashed, line width=0.4pt] coordinates {(0.6,0) (8.4,0)};
  \draw[black!35, densely dotted, line width=0.4pt] ({axis cs:3,0}|-{rel axis cs:0,0}) -- ({axis cs:3,0}|-{rel axis cs:0,1});
  \scalecurve{m ace}{cace}{data/campaign_control.csv}{mbpp}
  \addplot[draw=none, forget plot] table[x=k, y expr=\thisrow{mbpp}+\thisrow{mbpp_ci}, col sep=comma] {data/campaign_control.csv};
  \addplot[draw=none, forget plot] table[x=k, y expr=\thisrow{mbpp}-\thisrow{mbpp_ci}, col sep=comma] {data/campaign_control.csv};
\nextgroupplot[title={AppWorld}]
  \addplot[forget plot, black!50, densely dashed, line width=0.4pt] coordinates {(0.6,0) (8.4,0)};
  \draw[black!35, densely dotted, line width=0.4pt] ({axis cs:3,0}|-{rel axis cs:0,0}) -- ({axis cs:3,0}|-{rel axis cs:0,1});
  \scalecurve{m ace}{cace}{data/campaign_control.csv}{appworld}
  \addplot[draw=none, forget plot] table[x=k, y expr=\thisrow{appworld}+\thisrow{appworld_ci}, col sep=comma] {data/campaign_control.csv};
  \addplot[draw=none, forget plot] table[x=k, y expr=\thisrow{appworld}-\thisrow{appworld_ci}, col sep=comma] {data/campaign_control.csv};
\nextgroupplot[title={MiniSWEBench}]
  \addplot[forget plot, black!50, densely dashed, line width=0.4pt] coordinates {(0.6,0) (8.4,0)};
  \draw[black!35, densely dotted, line width=0.4pt] ({axis cs:3,0}|-{rel axis cs:0,0}) -- ({axis cs:3,0}|-{rel axis cs:0,1});
  \scalecurve{m ace}{cace}{data/campaign_control.csv}{miniswe}
  \addplot[draw=none, forget plot] table[x=k, y expr=\thisrow{miniswe}+\thisrow{miniswe_ci}, col sep=comma] {data/campaign_control.csv};
  \addplot[draw=none, forget plot] table[x=k, y expr=\thisrow{miniswe}-\thisrow{miniswe_ci}, col sep=comma] {data/campaign_control.csv};
\end{groupplot}
\end{tikzpicture}
\end{scaledfigure}
  \caption{\textbf{Edits-discarded control.} \acerl{}'s RSI@$K$ with edits applied minus discarded (each visit then reads the empty playbook), paired over tasks; above zero, the playbook helps. At $K=1$ both read the empty playbook, so the difference is drawn at zero; dotted: training horizon $F=3$. Checkpoints as in Table~\ref{tab:main-results}. Error bars: 1 SEM over tasks.}
  \label{fig:handoff-control}
\end{figure*}

\subsection{Where Does the Gain Come From?}
\label{sec:results-gain}

We are also curious whether \acerl{}'s harness optimization itself is effective, or it's just actor adaptation that has been improved. To find out, we repeat \acerl{}'s harnessed evaluation with every edit discarded, so each visit reads the empty playbook and the visits become independent attempts of the same actor (Figure~\ref{fig:handoff-control}). The difference between the two runs is the playbook's own contribution, with the actor held fixed. The playbook adaptation helps across benchmarks: within the training horizon, applying edits raises RSI@$K$ at $K=2$ and $K=3$ on all four, most clearly on AppWorld ($0.105\pm0.041$ at $K=2$) and MiniSWEBench ($0.090\pm0.040$ at $K=3$; $\pm1$ SEM). On MBPP the gain is small because both runs are near the ceiling ($0.987$ and $0.977$ at $K=3$). Beyond the horizon, the difference stays at or above zero at every $K$ from 4 to 8 on MiniSWEBench, MBPP, and AppWorld, and on MiniSWEBench applying edits adds $0.04$ to $0.07$ at each of these $K$. On Mastermind the benefit fades after $K=3$, and from $K=5$ on the discarded run is slightly ahead (by $0.016\pm0.016$ at $K=8$).
\section{Conclusion}
\label{sec:conclusion}

We formulated experience-handoff RL, which trains one model both to act and to write the experience that its future attempts inherit, and \acerl{}, which makes this practical within \grpo{}'s actor-rollout budget through a verifiable reward for the executed playbook and a comparison group for every role. At equal training compute on four benchmarks, \acerl{} reaches the highest held-out reward, above actor-only RL, ERL, and end-to-end training through the harness with a learned critic, and above a \grpo{}-trained model run in the same harness. The gain does not fade with more attempts: \acerl{}'s RSI@$K$ stays above \grpo{}'s pass@$K$ up to $K=8$, beyond the three visits seen in training. Appendices~\ref{sec:related} and~\ref{sec:limitations} discuss related work and limitations.

\subsection*{AI use statement}
We used generative AI tools to assist with writing and debugging the experiment and plotting code, and to help shorten and edit the paper text to fit the page limit. All research ideas, experimental design, and claims are the authors' own. We have reviewed all AI-assisted work and take responsibility for the final content of this work, including text, claims, or artifacts produced with the aid of generative AI.

\subsection*{Ethics statement}
This work trains open-weight language models on publicly available benchmarks (Mastermind, MBPP, AppWorld, and MiniSWEBench) in sandboxed environments; it involves no human subjects, personal data, or crowdsourced annotation. Training models to write the experience that their future attempts inherit could make autonomous agents improve with less human oversight. In our setting this is bounded: playbooks are task-local, are scored only by the task verifier, and are reset periodically, and every experiment runs in a sandbox without access to external systems. Deploying such training in open-ended software or interface agents would call for additional scoping and monitoring, which we discuss in Appendix~\ref{sec:limitations}.

\subsection*{Reproducibility statement}
The full training pipeline and hyperparameters are given in Appendix~\ref{app:pipeline}, benchmark settings, splits, and reward design in Appendix~\ref{app:benchmarks}, compute budgets and error bars in Appendix~\ref{app:compute}, and all prompts in Appendix~\ref{app:prompts}. All experiments use publicly available models (Qwen3-0.6B and Qwen3-4B-Instruct-2507) and benchmarks. We will release the training and evaluation code upon publication.

\bibliography{citation}
\bibliographystyle{iclr2027_conference}

\appendix

\section{Related Work}
\label{sec:related}

\paragraph{Harnesses and inference-time self-improvement.}
Reflexion, Voyager, ExpeL, Dynamic Cheatsheet, Agent Workflow Memory, and \ace{} let a frozen model improve later attempts by writing reflections, skills, or playbooks that its future calls consume \citep{shinn2024reflexion,wang2023voyager,expel2024,suzgun2026dynamic,awm2024,ace2025}. This is the form of recursive self-improvement (RSI) we study. RSI more broadly also covers agents that rewrite their own code or scaffold \citep{godelmachine2003,stop2023,godelagent2024,darwin2025}, agents that iterate on programs or research code against an automated evaluator \citep{alphaevolve2025,autoresearch2026}, and models that update their own weights from self-generated data or test-time RL \citep{selfrewarding2024,seal2025,discover2026}; see \citet{selfevolving2025} for a survey. We restrict RSI to rewriting context under fixed weights because that loop is an experience handoff between attempts. These systems assume a model that already writes useful experience; Figure~\ref{fig:motivation} shows that small open models do not. A complementary line optimizes the harness around a fixed model, by compiling prompts, evolving them reflectively, or searching over harness programs \citep{khattab2023dspy,gepa2025,metaharness2026,lin2026agentic}. \acerl{} fixes the harness and trains the model.

\paragraph{Agentic RL with verifiable rewards.}
\grpo{} and its refinements train actors from group-relative advantages on verifiable outcomes \citep{deepseekmath2024,guo2025deepseek,dapo2025,drgrpo2025}, and multi-turn agentic RL extends this to tool use and long horizons \citep{ragen2025,archer2024,rlmle2025,rlanything2026}. Recent open agentic models report large-scale RL pipelines with learned or rubric-based rewards for intermediate agent behaviour \citep{glm45_2025,kimik2_2025}. \acerl{} keeps the verifiable-reward ingredient and extends grouping from the actor to every model call in the harness, using downstream execution rather than a learned reward model to score intermediate text.

\paragraph{Training the model to a fixed harness.}
OpenClaw-RL trains a model on rollouts collected inside a production harness \citep{openclawrl2026}. Composer~2, SUPO, CompactionRL, MEM1, and MemAgent train a policy whose own summaries or compactions become its future context, propagating the final reward through execution and summary segments with sequence-level advantages or a critic \citep{composer22026,supo2025,compactionrl2026,mem1_2025,memagent2025}. OpenClaw-RL has no model-written handoff to credit, and the compaction methods generate summaries sequentially within each rollout and credit them through the rollout's outcome or a learned critic, without branching into alternative summaries from the same intermediate context. Our delayed-joint PPO baseline follows this structure (Section~\ref{sec:baselines}); Appendix~\ref{app:delayed-joint} shows that, with an executed playbook as the reward, the chain-wise structure estimates the same gradient as \acerl{}, so same-input groups change the variance of the credit rather than its target.

\paragraph{Learning reflections and memories from downstream outcomes.}
Reflect--Retry--Reward rewards a reflection by the success of the retry it guides \citep{bensal2025reflect}; ERL adds internalization so that the guided behaviour is reproduced without the reflection \citep{erl2026}; Memory-R1 trains memory operations by downstream answer correctness, freezing the answer agent during memory-manager training and training it separately \citep{memoryr12025}; JERP learns experiential rules alongside an RL-trained actor with a frozen editor \citep{jerp2026}. Training-Free \grpo{} aggregates insights from several rollouts into a shared guidebook without weight updates \citep{tfgrpo2025}. These are the closest precedents for verifiable credit on intermediate text. \acerl{} differs in training one policy for a two-stage handoff (reflect, then curate) with a persistent, editable artifact, and in constructing a comparison group for each stage.

\paragraph{Finer-grained grouping.}
GiGPO groups actions at repeated environment states and AT-GRPO forms agent- and turn-wise groups under shared inputs \citep{gigpo2025,atgrpo2025}. \acerl{} applies the same principle to the dependency tree of a harness, and its sparse tree is a specific way to form such groups in a tree whose leaves must be evaluated by rollouts.
\clearpage
\section{Chain-Wise Sampling: Delayed-Joint \acerl{}}
\label{app:delayed-joint}

The sampling tree of Section~\ref{sec:sparse-tree} is one way to estimate the gradient of Eq.~\eqref{eq:ehrl-objective} from executed playbooks. This appendix analyzes and evaluates the simplest alternative, which we call \emph{delayed-joint \acerl{}}: it keeps the verifiable playbook reward of Section~\ref{sec:candidate-learning} but forms one chain per actor rollout instead of one comparison group per role. The empirical comparison below covers Mastermind, MBPP, and AppWorld; on MiniSWEBench delayed-joint \acerl{} has only older-serving evaluations, shown as a reference. The compared runs differ in more than the grouping: delayed-joint \acerl{} samples eight seed rollouts, trains the actor only on them, and uses the earlier handoff prompts, whereas \acerl{} uses the budget-matched tree of Section~\ref{sec:sparse-tree} and the current prompts.

\subsection{Design}
\label{app:dj-design}

Delayed-joint \acerl{} samples $N=8$ actor rollouts of $(q,P)$. Each rollout $\tau_i$ then receives one reflection $z_i$, one curator edit $u_i$, and, if the edit is admissible, one fresh actor rollout that scores the candidate $P'_i=\mathrm{Apply}(P,u_i)$. The reflection and the edit of chain $i$ both receive that candidate's value $Y_i$, or $0$ if the edit is inadmissible, and the actor keeps its own reward; candidate rollouts are not trained. The $N$ reflections of a task form one group and the $N$ edits another, so each reflection is compared with reflections of \emph{other} trajectories. Playbook selection, resets, and the joint update are those of Section~\ref{sec:method}. Table~\ref{tab:dj-design} contrasts the two designs.

\begin{table}[h]
  \caption{\textbf{Sampling per task and training step}: \acerl{} with $(N,M,C)=(4,2,3)$ and delayed-joint \acerl{} with $N=8$.}
  \label{tab:dj-design}
  \centering
  \small
  \setlength{\tabcolsep}{4pt}
  \begin{tabular}{lL{0.33\linewidth}L{0.33\linewidth}}
    \toprule
    & \acerl{} & Delayed-joint \acerl{} \\
    \midrule
    Actor outputs & 4 rollouts of $(q,P)$ & 8 rollouts of $(q,P)$ \\
    Reflector outputs & 2 reflections of one seed & 1 reflection of each rollout (8) \\
    Curator outputs & 3 edits of one reflection, 1 of the other (4) & 1 edit of each reflection (8) \\
    Candidate evaluations & up to 4 & up to 8 \\
    Reflector group & reflections of the same seed & reflections of the 8 rollouts \\
    Curator group & edits of the same reflection & edits of the 8 reflections \\
    Reflector reward & mean value of its admissible edits & value of its single candidate \\
    Actor credit & own reward, seed and candidate rollouts & own reward, seed rollouts \\
    \bottomrule
  \end{tabular}
\end{table}

\subsection{Relation to the \ehrl{} Gradient}
\label{app:dj-equivalence}

Delayed-joint \acerl{} is the full-tree estimator of Eq.~\eqref{eq:app-full-tree-gradient} with $M=C=L=1$: each initial trajectory has one reflection, one edit, and one evaluation, so $Y_i$ is an unbiased estimate of $Q_{\mathrm r}$ and $Q_{\mathrm c}$ on its chain. Chains of the same task are conditionally independent given $s=(q,P)$. A leave-one-out mean $b_{-i}=\frac1{N-1}\sum_{j\ne i}Y_j$ therefore depends only on outputs that are independent of $z_i$ and $u_i$ given $s$, and is a valid baseline for every score term; the own-inclusive group mean introduces the same $(G-1)/G$ factor as Eq.~\eqref{eq:app-centering-factor}. For the reflector and curator, the delayed-joint update thus has the same expected raw direction as \acerl{}, and it needs none of the uniform-selection reductions of Eq.~\eqref{eq:app-group-sampling}. The two differ in actor credit: \acerl{} also trains the candidate rollouts, the $YS_1$ term of Eq.~\eqref{eq:app-local-credit-gap}, which this variant omits. Empirical normalization and clipping qualify both designs in the same way.

The designs differ in variance. Decomposing the delayed-joint reflector advantage,
\begin{equation}
  Y_i-b_{-i}
  =\underbrace{\bigl(Y_i-Q_{\mathrm r}(z_i)\bigr)}_{\text{evaluation noise}}
  +\underbrace{\bigl(Q_{\mathrm r}(z_i)-Q_0(\tau_i)\bigr)}_{\text{reflection quality}}
  +\underbrace{\bigl(Q_0(\tau_i)-b_{-i}\bigr)}_{\text{seed term}}.
  \label{eq:dj-decomposition}
\end{equation}
Only the middle term depends on the reflection itself. The seed term is constant given $\tau_i$ and has zero mean against $S_{\mathrm r}$, so it adds no bias, but it enters every reflection's advantage and grows with the spread of trajectory quality within a task. Comparing reflections of one seed, as \acerl{} does, replaces $b_{-i}$ by a baseline that shares $\tau$ and removes the seed term; comparing edits of one reflection removes the analogous term for the curator.

\subsection{What It Resembles}
\label{app:dj-resembles}

The chain-wise structure is the credit structure of end-to-end methods that train a model whose own summaries or compactions become its future context. Composer~2 and SUPO allow multiple successive summaries per chain and propagate the chain's outcome to their generated tokens with group-relative advantages \citep{composer22026,supo2025}; delayed-joint \acerl{} uses analogous chain-level credit for a reflection and an edit, with the executed playbook in place of the continued episode. CompactionRL runs the same kind of chain with a learned critic \citep{compactionrl2026}, which is our delayed-joint PPO baseline (Section~\ref{sec:baselines}). Delayed-joint \acerl{} thus sits between the two: it shares its chains with delayed-joint PPO and its reward with \acerl{}.

\subsection{Sampling Budget}
\label{app:dj-tokens}

Per task, delayed-joint \acerl{} generates 32 outputs (8 of each kind, counting evaluations), 16 of them actor rollouts, and \acerl{} at most 14, 8 of them actor rollouts (Table~\ref{tab:dj-design}). Appendix~\ref{app:compute} compares \acerl{} with \grpo{} in training compute.

\subsection{Training and Test Reward}
\label{app:dj-curves}

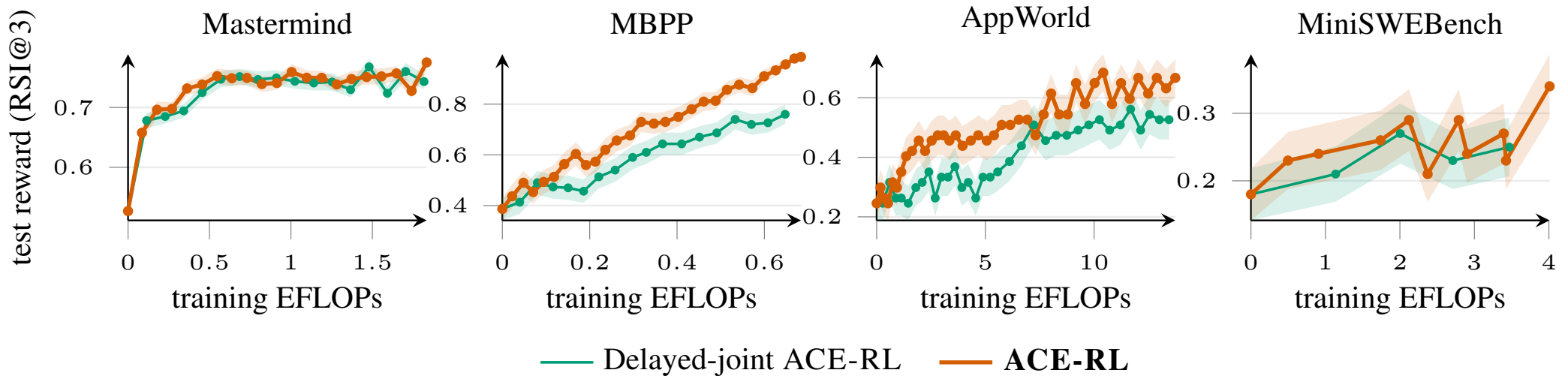
\begin{figure*}[h]
  \centering
\resultsfigstyle
\begin{scaledfigure}[1pt]
\begin{tikzpicture}
\begin{groupplot}[group style={group size=4 by 1, horizontal sep=0.6cm}, panel, width=3.95cm, height=2.9cm]
\nextgroupplot[title={Mastermind}, xmin=0, xmax=1.8335, restrict x to domain=0:1.8335, xlabel={training EFLOPs}, ylabel={test reward (RSI@3)}, legend to name=djleg]
  \addlegendimage{m dj}\addlegendentry{Delayed-joint \textsc{ACE-RL}}
  \addlegendimage{m ace}\addlegendentry{\textbf{\textsc{ACE-RL}}}
  \testcurve{m dj}{cdj}{data/campaign_flops_mastermind_delayed_joint.csv}{delayed_joint}
  \testcurve{m ace}{cace}{data/campaign_flops_mastermind_ace_rl.csv}{ace_rl}
\nextgroupplot[title={MBPP}, xmin=0, xmax=0.6838, restrict x to domain=0:0.6838, xlabel={training EFLOPs}]
  \testcurve{m dj}{cdj}{data/campaign_flops_mbpp_delayed_joint.csv}{delayed_joint}
  \testcurve{m ace}{cace}{data/campaign_flops_mbpp_ace_rl.csv}{ace_rl}
\nextgroupplot[title={AppWorld}, xmin=0, xmax=13.7373, restrict x to domain=0:13.7373, xlabel={training EFLOPs}]
  \testcurve{m dj}{cdj}{data/campaign_flops_appworld_delayed_joint.csv}{delayed_joint}
  \testcurve{m ace}{cace}{data/campaign_flops_appworld_ace_rl.csv}{ace_rl}
\nextgroupplot[title={MiniSWEBench}, xmin=0, xmax=4.0096, restrict x to domain=0:4.0096, xlabel={training EFLOPs}]
  \testcurve{m dj}{cdj}{data/campaign_flops_miniswe_delayed_joint.csv}{delayed_joint}
  \testcurve{m ace}{cace}{data/campaign_flops_miniswe_ace_rl.csv}{ace_rl}
\end{groupplot}
\node[anchor=north] at ([yshift=-0.75cm]$(group c2r1.south)!0.5!(group c3r1.south)$) {\pgfplotslegendfromname{djleg}};
\end{tikzpicture}
\end{scaledfigure}
  \caption{\textbf{Delayed-joint \acerl{} versus \acerl{}}: held-out test reward (RSI@3) against training EFLOPs, up to the compute budget. Both curves start at the base model (0 EFLOPs). Bands: 1 SEM over tasks.}
  \label{fig:dj-curves}
\end{figure*}

\begin{table}[h]
  \caption{\textbf{Test reward (RSI@3) at equal training compute}, chosen as in Table~\ref{tab:main-results}: delayed-joint \acerl{} at steps 160, 170, 340, and 40. Mean {\tiny$\pm$} SEM over tasks; bold: column maximum.}
  \label{tab:dj-results}
  \centering
  \small
  \begin{tabular}{lcccc}
    \toprule
    Method & Mastermind & MBPP & AppWorld & MiniSWEBench \\
    \midrule
    Delayed-joint \acerl{} & 0.743{\tiny$\pm$0.012} & 0.760{\tiny$\pm$0.038} & 0.526{\tiny$\pm$0.067} & 0.250{\tiny$\pm$0.044} \\
    \textbf{\acerl{}} & \textbf{0.775}{\tiny$\pm$0.014} & \textbf{0.987}{\tiny$\pm$0.011} & \textbf{0.667}{\tiny$\pm$0.063} & \textbf{0.340}{\tiny$\pm$0.048} \\
    \bottomrule
  \end{tabular}
\end{table}

Both designs learn to use the harness (Figure~\ref{fig:dj-curves}). Comparing each \acerl{} checkpoint with delayed-joint \acerl{}'s last evaluation at no more compute, \acerl{} is higher at 13 of 19 checkpoints on Mastermind, where the two stay close and end at $0.775$ and $0.743$ (paired $+0.032\pm0.034$). On MBPP it is higher at all 26 checkpoints and ends at $0.987$ against $0.760$ (Table~\ref{tab:dj-results}), on AppWorld at 37 of 40 (two ties), ending at $0.667$ against $0.526$ ($+0.140\pm0.114$), and on MiniSWEBench at 4 of 5, ending at $0.340$ against $0.250$ ($+0.090\pm0.084$).

\subsection{Scaling with Attempts}
\label{app:dj-scaling}

\begin{figure*}[h]
  \centering
\resultsfigstyle
\begin{scaledfigure}[1pt]
\begin{tikzpicture}
\begin{groupplot}[group style={group size=2 by 2, horizontal sep=0.65cm, vertical sep=1.1cm}, panel, width=3.35cm, height=2.35cm]
\nextgroupplot[title={Mastermind}, xmin=0.6, xmax=8.4, xtick={1,2,4,8}, legend to name=legF]
  \scalecurve{m base}{cbasem}{data/campaign_scaling_mastermind.csv}{base_rsi}
  \scalecurve{m dj}{cdj}{data/campaign_scaling_mastermind.csv}{delayed_joint_rsi}
  \scalecurve{m ace}{cace}{data/campaign_scaling_mastermind.csv}{ace_rl_rsi}
  \addlegendimage{m base}\addlegendentry{Base model}
  \addlegendimage{m dj}\addlegendentry{Delayed-joint \textsc{ACE-RL}}
  \addlegendimage{m ace}\addlegendentry{\textbf{\textsc{ACE-RL}}}
\nextgroupplot[title={MBPP}, xmin=0.6, xmax=8.4, xtick={1,2,4,8}]
  \scalecurve{m base}{cbasem}{data/campaign_scaling_mbpp.csv}{base_rsi}
  \scalecurve{m dj}{cdj}{data/campaign_scaling_mbpp.csv}{delayed_joint_rsi}
  \scalecurve{m ace}{cace}{data/campaign_scaling_mbpp.csv}{ace_rl_rsi}
\nextgroupplot[title={AppWorld}, xmin=0.6, xmax=8.4, xtick={1,2,4,8}, xlabel={attempts $K$}]
  \scalecurve{m base}{cbasem}{data/campaign_scaling_appworld.csv}{base_rsi}
  \scalecurve{m dj}{cdj}{data/campaign_scaling_appworld.csv}{delayed_joint_rsi}
  \scalecurve{m ace}{cace}{data/campaign_scaling_appworld.csv}{ace_rl_rsi}
\nextgroupplot[title={MiniSWEBench}, xmin=0.6, xmax=8.4, xtick={1,2,4,8}, xlabel={attempts $K$}]
  \scalecurve{m base}{cbasem}{data/campaign_scaling_miniswe.csv}{base_rsi}
  \scalecurve{m dj}{cdj}{data/campaign_scaling_miniswe.csv}{delayed_joint_rsi}
  \scalecurve{m ace}{cace}{data/campaign_scaling_miniswe.csv}{ace_rl_rsi}
\end{groupplot}
\node[anchor=north] at ([yshift=-0.7cm]$(group c1r2.south)!0.5!(group c2r2.south)$) {\pgfplotslegendfromname{legF}};
\end{tikzpicture}
\end{scaledfigure}
  \caption{\textbf{RSI@$K$ of both designs} at the checkpoints of Table~\ref{tab:dj-results}, and of the untrained base model (gray). Error bars: 1 SEM over tasks.}
  \label{fig:dj-scaling}
\end{figure*}
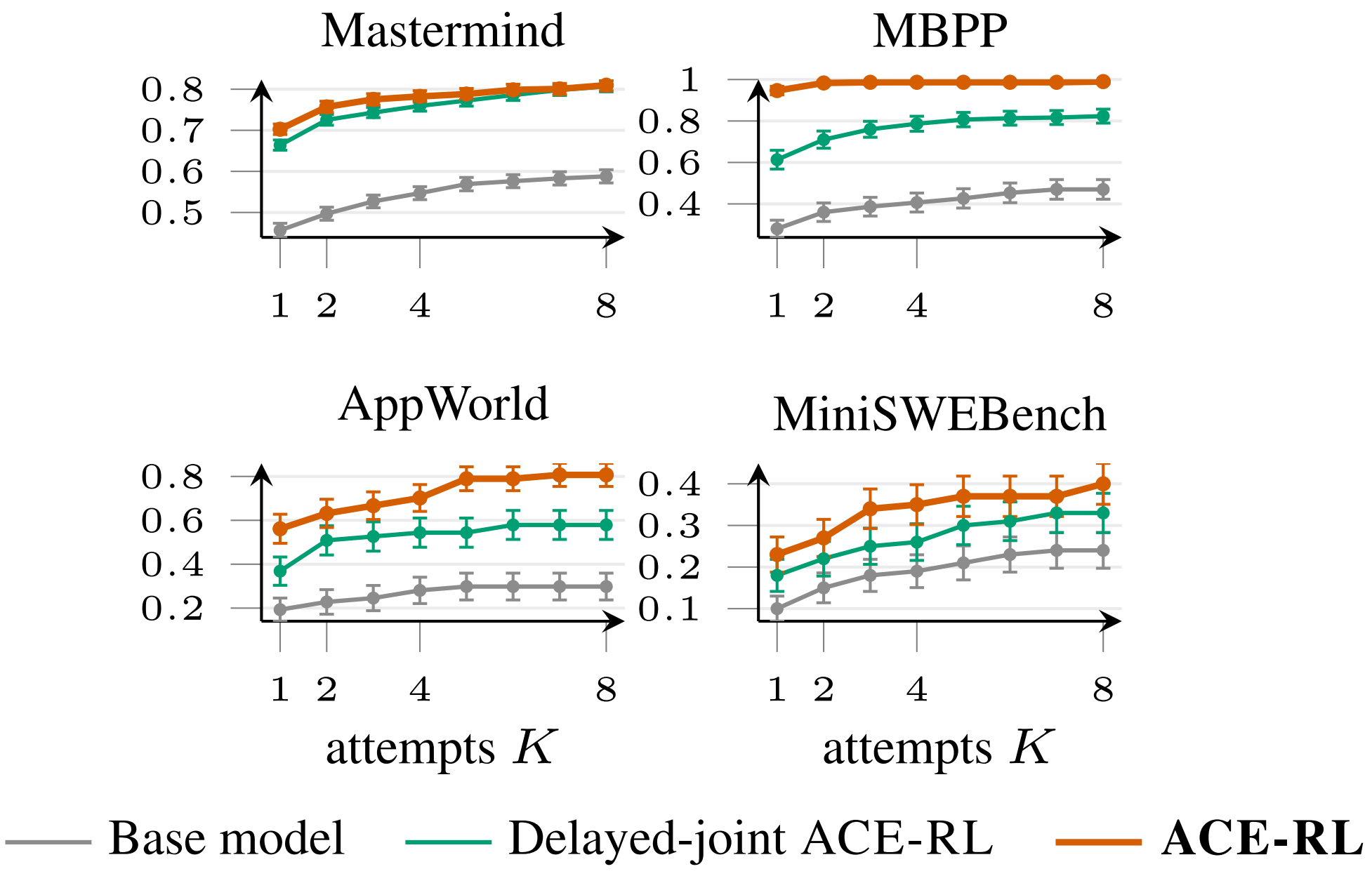

Both designs also scale with the number of visits (Figure~\ref{fig:dj-scaling}). On Mastermind \acerl{} starts higher ($0.703$ against $0.664$ at $K=1$) and the two end level at $K=8$ ($0.810$ and $0.807$). On MBPP \acerl{} leads at every $K$, from $0.95$ against $0.61$ at $K=1$ to $0.99$ against $0.82$ at $K=8$; most of the gap is at the first visit, before any handoff. On AppWorld \acerl{} also leads at every $K$, from $0.56$ against $0.37$ at $K=1$ to $0.81$ against $0.58$ at $K=8$, and on MiniSWEBench from $0.23$ against $0.18$ to $0.40$ against $0.33$.

\subsection{Discussion}
\label{app:dj-discussion}

With the verifiable playbook reward held fixed, the choice between chain-wise comparison and per-role groups changes the variance of the handoff credit, not its target. In our runs at equal compute the two designs are close on Mastermind, and \acerl{} is ahead on MBPP, mostly at the first visit, on AppWorld at almost every checkpoint, and on MiniSWEBench within task-level noise; because the runs also differ in actor sampling, actor credit, and handoff prompts, these results do not isolate the effect of the grouping. Per-role groups remove the seed contribution to advantage variance under the assumptions of Appendix~\ref{app:dj-equivalence}, and \acerl{} reaches these results with half the actor rollouts of the chain-wise design. Measuring that variance directly, and establishing with additional training runs when it pays off, is left to future work.

\clearpage
\clearpage
\section{Role Prompts}
\label{app:prompts}

The listings below are the actor, reflector, and curator prompts used for training on each benchmark. The actor prompt contains the current playbook; the reflector sees the trajectory summary and record, including the environment's final reply; the curator sees the playbook and the reflector's insight and returns JSON edit operations. We use the same prompt for all experiments. For GRPO and PPO experiments we let the playbook blank. The curator and reflector prompts are not used for those experiments. For ERL experiment we use their own prompt.

\subsection{Mastermind}
\label{app:prompts-mastermind}

\paragraph{Actor.}
\begin{lstlisting}[style=prompt]
You are Qwen, created by Alibaba Cloud. You are a helpful assistant playing Mastermind, a code-breaking game. Your goal is to guess the secret code within the allowed number of turns with the help of a playbook.

## Game Rules
1. The secret code is a sequence of digits. Each turn tells you the code length and the valid digits (0 to num_digits-1).
2. The same digit may appear more than once in the secret code.
3. After each guess, you receive feedback:
   - exact_matches: number of digits with the correct value AND the correct position
   - value_matches: number of digits with the correct value but the WRONG position
4. Rewards:
   - Solving the complete code earns reward 1.
   - Otherwise each exact match earns 1/code_length and each correct value in a wrong position earns 0.5/code_length.
   - Your score for the game is your BEST single guess. A new guess can never lower your score, so every turn spent on a new, well-chosen guess is free information.
5. A guess with the wrong number of digits, an invalid digit, or no guess line is rejected with an error. A rejected guess still uses up a turn and earns nothing.
6. Read **Current Playbook** to help you solve the game.

## Current Playbook
You are provided with a playbook containing strategies learned from earlier attempts at this same game.

[PLAYBOOK BEGINS HERE]
{{ playbook }}
[PLAYBOOK ENDS HERE]

## Strategy Reminders
- Start with a guess that has diverse digits to gather information.
- A guess is useful only if it is consistent with ALL feedback in the Guess History: if the secret were your guess, every earlier guess would have received exactly the feedback shown.
- exact_matches + value_matches is how many digits of that guess are in the code. If it equals code_length, the code is a rearrangement of that guess's digits.
- After a high exact_matches count, keep most positions and change only one or two.

## Key Instructions
- First review the playbook before making your guess.
- Playbook contains game-specific tips. Apply a tip if it clearly matches your current situation.
- The feedback in the guess history is always correct. If a tip contradicts the feedback, follow the feedback.
- Make exactly ONE guess per turn. Repeating a guess from the Guess History wastes the turn: it gets the same feedback again.
- Reply in exactly these three lines:
  Previous guess: <your most recent guess from the Guess History, or "none" on the first turn>
  Change: <which positions you change and to which digits, and which feedback line made you choose them>
  ```<your new guess>```
- Your new guess must differ from EVERY guess in the Guess History and must have exactly Code length digits; count them. With Code length 5, a guess line looks like ```0 1 2 3 4```; with Code length 3, like ```0 1 2```.
- Do NOT put a language name after the backticks (no ```python), do NOT write words such as "Guess:" inside the backticks, and do NOT copy the game state into backticks.
\end{lstlisting}

\paragraph{Reflector.}
\begin{lstlisting}[style=prompt]
You are an expert analyst for Mastermind code-breaking trajectories. Your job is to follow a 3-step process below to read a Generator's attempt at one Mastermind game, understand why it succeeded or failed, and propose a concrete and specific key insight about how to deduce this code efficiently (e.g., which elimination step was missed, duplicate-digit accounting, narrowing positions after a high exact count, what to do after a guess whose digits are all in the code).

## Context

The Generator plays one Mastermind game. Each turn it sees the code length, the valid digits, and the history of its own guesses with feedback: exact_matches (correct digit in the correct position) and value_matches (correct digit in the wrong position). It answers with reasoning and one guess. The game ends when the code is solved or the turn limit is reached. Reward per guess is (exact_matches + 0.5 * value_matches) / code_length, and the game's score is its BEST guess, so only a fully solved code scores 1 and a losing game can still score 0.75.

The Generator is already instructed to never repeat a guess, to keep every guess consistent with all earlier feedback, and to end each reply with one guess line such as ```0 1 2 3```. An insight that only repeats these instructions adds nothing. If the Generator broke one of them, say which turn and which situation led to it, and what it should have guessed instead.

You are given the following information about the Generator's attempt:

1. **Trajectory Summary**: The actual result, whether the Generator solved the code, how many turns it used, and the best reward it reached.
2. **Trajectory Record**: A guess table followed by the Generator's replies.
   - The guess table is computed from the environment and is always correct. Each row shows the guess as the environment read it, its exact and value feedback, and its reward. A row marked REJECTED scored nothing. A row marked REPEATS repeated an earlier guess. A row marked "contradicts feedback of turn N" could not be the secret code, given what turn N already showed.
   - The replies show the Generator's own reasoning (which may contain wrong deductions). Long replies are cut.

You must STRICTLY follow these 3 steps in **Your 3-Step Tasks**. Keep each step brief.

## Your 3-Step Tasks

### Step 1 - Extract Actual Facts:

Read the Trajectory Summary. Determine:
- Is the outcome SUCCESS or FAILED?
- How many turns were used, and what is the best reward reached?

Then read the counts line and the guess table. Note:
- How many guesses were accepted, how many were distinct, how many were rejected, how many repeated an earlier guess, and how many contradicted earlier feedback. Copy these numbers; do not recount the rows.
- The two or three most informative rows: the highest exact_matches, and any row where exact_matches + value_matches equals the code length. Copy those rows exactly.
- The first turn that was rejected, repeated, or contradicted earlier feedback, if any.

Do NOT copy the whole table.

### Step 2 - Understand the Generator:

Based on the actual facts from Step 1, analyze the Generator's behavior.

If SUCCESS:
- Which rows narrowed the candidates the most, and what did the Generator conclude from them?
- Were any turns wasted on rejected, repeated, or contradicting guesses? What led to them?

If FAILED:
- What did the most informative rows establish, e.g. how many digits of a guess are in the code, or that most positions of a guess are already right?
- Did the Generator build on those rows, or did it move away from them, change several positions at once, or go back to earlier guesses?
- If guesses were rejected, what did the replies do instead of ending with a guess line (e.g., a language name after the backticks, words inside the backticks, the game state copied into backticks, no guess at all)?
- Given the rows you copied, what kind of next guess was available that would have been new and consistent with all feedback?

### Step 3 - Determine the Case:

Based on your analysis in Step 2, which case does this belong to?

- Case 1: The Generator solved the code efficiently, with no rejected, repeated, or contradicting guesses.
- Case 2: The Generator solved the code but wasted turns on rejected, repeated, or contradicting guesses.
- Case 3: The Generator did not solve the code, and its guesses were accepted but poorly chosen, e.g. it moved away from its most informative rows, changed several positions at once, or mishandled repeated digits.
- Case 4: The Generator did not solve the code, and the failure is behavioral, e.g. many turns were rejected or repeated an earlier guess.

## Episode Info

### Trajectory Summary

{{trajectory_summary}}

### Trajectory Record

[TRAJECTORY RECORD STARTS HERE]
{{conversation_history}}
[TRAJECTORY RECORD ENDS HERE]

## Output Format

Complete Step 1 through Step 3 above before writing your JSON output.

After completing the 3 steps, respond with a JSON object. Do NOT wrap in markdown code blocks. Choose the format matching your case from Step 3.

Your key_insight must be one or two sentences the Generator can act on during its next attempt at this same game. Tie it to the situation in this game by describing the feedback pattern, e.g. "when a guess scores exact + value equal to the code length, keep those digits and only rearrange them" or "when the reply keeps adding a language name after the backticks, the guess is rejected". Do NOT state the secret code, and do NOT state which digits the code contains or which position a digit occupies; the Generator must re-derive them from the feedback it receives.

If the Generator solved the code efficiently (Case 1):

{
  "outcome": "SUCCESS",
  "reasoning": "Which facts from Step 1 and which deductions from Step 2 made this game short? Why Case 1 rather than Case 2?",
  "key_insight": "Which way of using the feedback made this game short, and which part of it may not be reliably reproduced next time?"
}

If the Generator solved the code but wasted turns (Case 2):

{
  "outcome": "SUCCESS",
  "reasoning": "Which turns were wasted (from Step 1) and why (from Step 2)? Why Case 2 rather than Case 1?",
  "key_insight": "What situation led to the wasted turns, and what should the Generator do in that situation instead?"
}

If the Generator did not solve the code and chose its guesses poorly (Case 3):

{
  "outcome": "FAILED",
  "reasoning": "Which informative rows did you copy in Step 1? How did the Generator's later guesses fail to build on them? Why Case 3 rather than Case 4?",
  "key_insight": "Which feedback pattern should the Generator recognize in this game, and how should it choose its next guess when it sees it?"
}

If the Generator did not solve the code because of behavioral problems (Case 4):

{
  "outcome": "FAILED",
  "reasoning": "How many turns were rejected or repeated (from Step 1)? What in the replies caused it (from Step 2)? Why Case 4 rather than Case 3?",
  "key_insight": "What exactly went wrong in the replies (quote the pattern, e.g. a language name after the backticks), and what should the Generator do instead?"
}

## Important Notes
- The guess table is computed from the environment and is always correct. If the Generator's reasoning disagrees with it, trust the table.
- A high partial reward does not mean the code was solved. Use the outcome from the Trajectory Summary.
- Do NOT state the secret code, its digits, or the position of any digit. The insight must describe how to reason, not what the answer is.
- Success does not update the Generator weights between attempts and does not show that the next attempt can recover the reasoning without advice. Never describe a reasoning step as mastered or no longer needed because the Generator applied it in this game.
- Your whole response must fit in about 1500 words. Keep Step 1 to the numbers and the few rows asked for.
\end{lstlisting}

\paragraph{Curator.}
\begin{lstlisting}[style=prompt]
You are an editor for a per-game Mastermind playbook. The playbook is a cheatsheet for playing this specific game, it should be concise, but it should have essential hints about how to deduce the code efficiently (e.g., keeping every guess consistent with all feedback so far, narrowing positions after a high exact_matches count, accounting for repeated digits, avoiding repeated or malformed guesses) that provide marginal, additional, complementary help for the Generator to solve this game on the next attempt. Tips MUST describe how to reason. Tips must NOT name the secret code, the digits it contains, or the position of any digit -- the Generator has to re-derive those from the feedback it receives.

You are given:
- **Current Playbook**: The accumulated playbook from prior epochs for this game (at **Current Playbook** section below).
- **Reflector's Latest Insights**: One or more supplied insights from trajectory analyses of this game, each including the outcome (SUCCESS or FAILED), reasoning, and key insight. Process only the insights actually present; do not invent additional insights.

Follow the **Your 5-Step Task** section below **step by step** to decide how to manage the playbook BEFORE producing your output.

## Your 5-Step Task

### Step 1: Describe the current status of the Playbook as following:

1. Can you identify tips in the **Current Playbook** section? If the playbook is blank, note that it is empty.
2. How many tips are in the Current Playbook? Aim for at most 5 tips. When the book is longer, merge exact duplicates while retaining their useful facts. Do not delete a correct tip merely to meet this target; future attempts still need the same evidence.

### Step 2: Evaluate each insight independently as following:

For each insight (Insight 1, Insight 2, ...):

First, evaluate the quality of this insight:
- Is this insight empty, missing, or not provided? If yes, skip this insight.
- Does this insight name the secret code, the digits the code contains, or the position of a digit? If yes, skip this insight. Such a tip would be useless on the next attempt and must never enter the playbook.
- Is this insight specific and actionable? Does it describe a situation in this game (a feedback pattern, or a way the replies went wrong) and what to do in it, not just restate the outcome or the reward? If not, skip this insight.
- Does this insight only repeat the Generator's standing instructions ("never repeat a guess", "keep guesses consistent with the feedback", "end with one guess line") without naming the situation that broke them and a concrete alternative? If yes, skip this insight.

Then, determine which case this insight belongs to by reading its outcome and content:
- Outcome SUCCESS, and the game was solved without wasted turns? This is **Case 1**.
- Outcome SUCCESS, but turns were wasted on rejected, repeated, or contradicting guesses? This is **Case 2**.
- Outcome FAILED, and the guesses were accepted but poorly chosen (a feedback pattern was not used)? This is **Case 3**.
- Outcome FAILED, and turns were lost to rejected or repeated guesses? This is **Case 4**.

After evaluating all insights, summarize: Which insights passed quality filtering? Which case does each belong to? (e.g., "Insight 1: Case 3, Insight 3: Case 1. Insights 2 and 5 were skipped due to low quality.")

### Step 3: Compare the insights against existing tips as following:

If the playbook is empty, there is nothing to compare against: every insight that passed Step 2 is a candidate for a new tip. Go to Step 4.

If the playbook is non-empty, read each existing tip and ask:
- Does any insight ask to remove this tip? Only plan REMOVE if the insight gives concrete trajectory evidence that this tip is wrong or harmful, or if the same fact remains in an explicitly identified retained tip. A request to remove is not evidence by itself.
- Do any of the remaining insights (those that passed quality filtering in Step 2) describe the same reasoning step as this tip? Which insights?
- If yes, do those insights collectively give the Generator more usable guidance than this tip? Do they add a concrete procedure that this tip lacks?
- If yes, do any of those insights contradict this tip with corrected information?
- If no insights relate to this tip, this tip is unaffected.

Then ask: Are there any remaining insights that do not relate to any existing tip? Which ones? These point to potential new areas for the playbook.

### Step 4: Choose operations based on case and comparison as following:

First rule: if the playbook is empty and at least one insight passed Step 2, you MUST ADD one tip built from the most useful of those insights. An empty playbook gives the Generator no help on its next attempt.

Then process each case group separately. Within each case, use the corresponding guidance below.

**Case 1 insights** (solved efficiently):

Overall tendency: conservative. No operation is the default.
- Is the current playbook adequate given that the Generator solved the game? If yes, no operation.
- Preserve tips that helped the successful attempt. A success does not establish that a fresh Generator attempt will remember the fact without the playbook. REMOVE requires concrete evidence the tip is wrong or harmful, or an exact duplicate whose fact remains in an identified retained tip.
- Do any Case 1 insights describe a deduction step that is not yet covered in the playbook and that the Generator may not reliably repeat? Which insights? If yes, consider ADD. If multiple insights point to the same step, treat them as one and ADD once.

**Case 2 insights** (solved, but wasted turns):

Overall tendency: more active. ADD and UPDATE are the primary operations.
- Do any Case 2 insights describe the situation that wasted turns, and is it not yet covered by any existing tip? Which insights? If yes, ADD. If multiple insights describe the same situation, treat them as one and ADD once.
- Do any Case 2 insights relate to an existing tip that was too vague to prevent the waste? Which insights, which tip? If yes, UPDATE to make it concrete.
- The Generator needed extra turns -- REMOVE is almost never justified.

**Case 3 insights** (unsolved, guesses poorly chosen):

Overall tendency: most active. ADD is the primary operation.
- Do any Case 3 insights describe a feedback pattern and how to choose the next guess after it, not covered by any existing tip? Which insights? If yes, ADD -- this is the most important source of playbook content. If multiple insights describe the same pattern, treat them as one and ADD once.
- Do any Case 3 insights relate to an existing tip that covers a related pattern but was not sufficient to prevent this failure? Which insights, which tip? If yes, UPDATE to make it more specific.
- The Generator failed -- REMOVE is almost never justified.

**Case 4 insights** (unsolved, turns lost to rejected or repeated guesses):

Overall tendency: active but constrained. Keep at most one tip about reply format and one about repeated guesses.
- Do any Case 4 insights name the concrete reply pattern that was rejected or repeated, and what to do instead, not yet covered by any existing tip? Which insights? If yes, ADD.
- Do any Case 4 insights relate to an existing tip about the same pattern whose guidance is too vague? Which insights, which tip? If yes, UPDATE to make it more concrete.
- Is the pattern already covered by a concrete existing tip? If yes, no operation.
- The Generator failed -- REMOVE is almost never justified.

### Step 5: Verify your output:

1. If the playbook is empty, did you choose UPDATE or REMOVE? That's invalid -- the only valid operations on an empty playbook are ADD or no operation.
2. If your operation is ADD, does it include only `type` and `content` with NO `bullet_id`?
3. If your operation is UPDATE, does it include `type`, `bullet_id`, and `content`?
4. If your operation is REMOVE, does it include only `type` and `bullet_id`?
5. Is your content a plain JSON string containing the tip itself, free of tip IDs like `[1]`, labels such as "Tip:" or "Tip content:", and insight/case classifications such as "Insight 1: Case 3"? The system assigns IDs automatically. Classifications and operation explanations belong only in `reasoning`, never in tip `content`.
6a. Is every tip you ADD or UPDATE written from the insights, not copied from this prompt's placeholders, and does it say something concrete (never "No tips", "Improved content", or a placeholder in angle brackets)? If not, drop the operation.
6b. Does your ADD repeat a tip already in the playbook? If yes, drop it.
6. Does any tip you ADD or UPDATE name the secret code, a digit the code contains, or the position of a digit? If yes, rewrite it as a rule for reasoning, or drop the operation.
7. Does every REMOVE cite concrete evidence that the tip is wrong or harmful, or identify an exact duplicate retained elsewhere? Similar subject matter alone is insufficient. Preserve complementary facts and correct tips used in a successful attempt.
8. Do any of your operations target the same tip or the same reasoning step? If yes, merge them into one operation.

## Current Playbook

[PLAYBOOK BEGINS HERE]

{{current_playbook}}

[PLAYBOOK ENDS HERE]

## Reflector's Latest Insights

{{guidebook}}

## Output Format

Complete the checks in Steps 1 through 5, then respond with exactly one top-level JSON object with two keys: `reasoning` and `operations`. Do not output the steps separately, add prose, or use a markdown code block.

- `reasoning` is a brief JSON string explaining the chosen operations and the supplied insights that support them. Put insight numbers and case classifications here only.
- `operations` is one flat JSON array. Each element is directly an operation object, not another object containing `reasoning` or `operations`.
- An ADD object has exactly `type` (the string `ADD`) and `content`.
- An UPDATE object has exactly `type` (the string `UPDATE`), `bullet_id`, and `content`.
- A REMOVE object has exactly `type` (the string `REMOVE`) and `bullet_id`.
- `bullet_id` is a JSON string identifying an existing tip in the Current Playbook.
- `content` is a plain JSON string with one or two sentences describing the feedback situation from the insight and how to reason about the next guess. Never put an object, array, encoded JSON, case label, outcome restatement, or copied prompt placeholder in `content`.

The single response shape is below. Fill in `reasoning` and put all chosen operation objects directly in the array. An empty array is valid only when the checks found no supported change; it is not the default when Step 4 requires an ADD.

{
  "reasoning": "",
  "operations": []
}

## Important Notes
- Playbook is located between [PLAYBOOK BEGINS HERE] and [PLAYBOOK ENDS HERE].
- Playbook is empty if and only if there's no content between [PLAYBOOK BEGINS HERE] and [PLAYBOOK ENDS HERE], or the only content is "(No strategies learned yet)".
- Count tips in the playbook carefully. NEVER assume or guess the number of tips.
- Tips should describe how to reason about the feedback, and must never name the secret code, its digits, or the position of a digit.
- A high partial reward does not mean the game was solved. Only a fully solved code counts as SUCCESS; use the outcome reported with each insight.
- The Generator is a small model that reads the playbook every turn. Keep each tip to one or two sentences.
\end{lstlisting}

\subsection{MBPP}
\label{app:prompts-mbpp}

\paragraph{Actor.}
\begin{lstlisting}[style=prompt]
You are Qwen, created by Alibaba Cloud. You are an expert Python programmer. Your task is to write Python code that solves the given problem.

## Problem Rules
1. Read the problem description carefully
2. Write clean, correct, simple Python code that solves the problem
3. Your code will be tested against test cases
4. If tests fail, analyze the error messages and revise your code
5. Read **Current Playbook** to help you answer the question.

## Current Playbook
You are provided with a playbook containing strategies learned from solving the same problem.

[PLAYBOOK BEGINS HERE]
{{ playbook }}
[PLAYBOOK ENDS HERE]

## Important Programming Reminders
- Make sure your function names match what the tests expect
- Handle edge cases properly
- Keep your code simple and readable

## Key Instructions
- First review the playbook before solving the problem
- Playbook constains question-specific tips. Apply a tip if it clearly matches your current problem.
- Show your thought process before writing code
- Provide your final code in a ```python``` code block
\end{lstlisting}

\paragraph{Reflector.}
\begin{lstlisting}[style=prompt]
You are an expert code analyst. Your job is to follow a 3-step process below to read a Generator's coding trajectory, understand why it succeeded or failed, and propose a concrete and specific key insight about the key characteristics of this problem (e.g., input edge cases, output requirements, what core algorithm or approach is needed, how that algorithm works and its constraints -- when it does or does not apply, common mistakes).

You are given the following information about a Generator's attempt to solve a coding task:

1. **Trajectory Summary**: The actual result, whether the Generator succeeded or failed, how many turns it used, and why it stopped. The test cases are NEVER wrong.
2. **Trajectory Record**: The full record of the Generator's attempt. This contains two kinds of information:
   - The Generator's own reasoning and code (which may contain wrong assumptions)
   - The test results and error messages from the environment (which are actual outcomes)

You must STRICTLY follow these 3 steps in **Your 3-Step Tasks**.

## Your 3-Step Tasks

### Step 1 - Extract Actual Facts:

Read the Trajectory Summary. Determine:
- Is the outcome SUCCESS or FAILED?
- How many test cases are passed?
- How many turns were used?
- What is the termination reason?

Then read the Trajectory Record. Find and quote:
- The error message from the test output (if any). Copy the exact error text.
- The test assertion that failed (if any). Copy the exact assert line.

### Step 2 - Understand the Generator:

Based on the actual facts from Step 1, analyze the Generator's behavior in depth.

If SUCCESS:
- What strategy did the Generator use? Why was it effective for this problem?
- What key characteristics of this problem did the Generator correctly handle?
- If multi-turn: what was incomplete or wrong in the first attempt? What feedback did the Generator receive? What did it change to succeed?

If FAILED:
- What did the Generator believe about its code? What does the evidence actually show? Where is the gap?
- Did the Generator make several changes across turns, or did it submit similar code, make only surface-level edits, or blame the test cases?
- What did the Generator never check or re-examine despite the error?
- What should the correct approach be for this problem?

### Step 3 - Determine the Case:

Based on your analysis in Step 2, which case does this belong to?

- Case 1: The Generator succeeded on the first turn.
- Case 2: The Generator was partially right or wrong initially, but corrected itself in a later turn.
- Case 3: The Generator failed AND made several attempts to fix the code but still could not solve it, suggesting a knowledge gap.
- Case 4: The Generator failed AND did not effectively revise its code, submitting similar code, making only superficial changes, or blaming the test cases.

## Episode Info

### Trajectory Summary

{{trajectory_summary}}

### Trajectory Record

The following shows the Generator's attempts to solve the task. Each turn contains the Generator's reasoning, its submitted code, and the test results.

[TRAJECTORY RECORD STARTS HERE]
{{conversation_history}}
[TRAJECTORY RECORD ENDS HERE]

## Output Format

Complete Step 1 through Step 3 above before writing your JSON output.

After completing the 3 steps, respond with a JSON object. Do NOT wrap in markdown code blocks. Choose the format matching your case from Step 3.

Your key_insight should be specific to this problem's key characteristics (e.g., input edge cases, output requirements, what core algorithm or approach is needed, how that algorithm works and its constraints -- when it does or does not apply, common mistakes). Do NOT include exact code in the key_insight -- describe the approach as a hint, not as a solution.

If the Generator succeeded on the first turn (Case 1):

{
  "outcome": "SUCCESS",
  "reasoning": "What actual outcomes did you extract in Step 1? What strategy did you identify in Step 2? Why did you classify this as Case 1 in Step 3?",
  "key_insight": "What key characteristics of this problem did the Generator correctly handle? What did the Generator do that may not be reliably reproduced in future attempts at this same problem?"
}

If the Generator was partially right or wrong initially, but corrected itself in a later turn (Case 2):

{
  "outcome": "SUCCESS",
  "reasoning": "What actual outcomes did you extract in Step 1? What did the Generator miss and how did it correct itself, as you analyzed in Step 2? Why did you classify this as Case 2 in Step 3?",
  "key_insight": "What key characteristics of this problem did the Generator overlook on its first attempt? How did it correct itself? What is easy to miss in future attempts at this same problem?"
}

If the Generator failed and made substantive attempts to fix the code but still could not solve it, suggesting a knowledge gap (Case 3):

{
  "outcome": "FAILED",
  "reasoning": "What actual outcomes did you extract in Step 1? What knowledge gap did you identify in Step 2 -- did the Generator fail to recognize what algorithm or approach is needed, not know the strategy, or misunderstand how a mechanism works? What did it attempt in later turns and why did it still fail? Why did you classify this as Case 3 in Step 3?",
  "key_insight": "What key characteristics of this problem did the Generator fail to understand? What is the correct approach -- specific enough to change behavior but without exact code? If a misunderstanding was involved, what does not work and why?"
}

If the Generator failed and did not effectively revise its code, i.e., submitting similar code, making only superficial changes, or blaming the test cases (Case 4):

{
  "outcome": "FAILED",
  "reasoning": "What actual outcomes did you extract in Step 1? What behavioral pattern did you identify in Step 2 -- did the Generator fail to understand the error signal, persist with a failing approach, or fail to locate the error? What did it attempt in later turns and why was it ineffective? Why did you classify this as Case 4 in Step 3?",
  "key_insight": "What failure situation did the Generator face in this problem? How did it respond to the error signal? What signal should it have recognized? What specific action should it have taken instead?"
}

## Important Notes
- The test cases are NEVER wrong. Think carefully when Generator keeps saying the code is correct but actually wrong.
- Your insight should be specific to this problem's key characteristics, not a general strategy applicable to all tasks.
- Success does not update the Generator weights between attempts and does not show that the next attempt can recover the fact without advice. Never describe a characteristic as mastered or no longer needed because the Generator handled it in this attempt.
- Do NOT include exact code in the key_insight. Describe the approach as a hint that provides marginal, additional, complementary help -- not a complete solution.
\end{lstlisting}

\paragraph{Curator.}
\begin{lstlisting}[style=prompt]
You are an editor for a per-problem Python coding playbook. The playbook is a cheatsheet for solving this specific problem, it should be concise, but it should have essential hints about the key characteristics of this problem (e.g., input edge cases, output requirements, what core algorithm or approach is needed, how that algorithm works and its constraints -- when it does or does not apply, common mistakes) that provide marginal, additional, complementary help for the Generator to solve this task on the next attempt. Tips in the playbook should NOT contain exact code.

You are given:
- **Current Playbook**: The accumulated playbook from prior epochs for this problem (at **Current Playbook** section below).
- **Reflector's Latest Insights**: Multiple insights from the most recent epoch's trajectory analyses. Each insight comes from a different rollout of the same problem, and includes the outcome (SUCCESS or FAILED), reasoning, and key insight.

Follow the **Your 5-Step Task** section below **step by step** to decide how to manage the playbook BEFORE producing your output.

## Your 5-Step Task

### Step 1: Describe the current status of the Playbook as following:

1. Can you identify tips in the **Current Playbook** section? If the playbook is blank, note that it is empty.
2. How many tips are in the Current Playbook? Aim for at most 10 tips. When the book is longer, merge exact duplicates while retaining their useful facts. Do not delete a correct tip merely to meet this target; future attempts still need the same evidence.

### Step 2: Evaluate each insight independently as following:

For each insight (Insight 1, Insight 2, ...):

First, evaluate the quality of this insight:
- Is this insight empty, missing, or not provided? If yes, skip this insight.
- Is this insight specific and actionable? Does it describe a key characteristic of this problem, not just restate the outcome? If not, skip this insight.

Then, determine which case this insight belongs to by reading its content:
- Does this insight describe a successful strategy the Generator used on the first attempt? If yes, this is **Case 1**.
- Does this insight describe something the Generator overlooked initially but corrected itself later? If yes, this is **Case 2**.
- Does this insight describe a knowledge gap, i.e., the Generator failed and could not self-correct despite substantive attempts? If yes, this is **Case 3**.
- Does this insight describe a behavioral problem, i.e., the Generator failed and did not effectively revise its code? If yes, this is **Case 4**.

After evaluating all insights, summarize: Which insights passed quality filtering? Which case does each belong to? (e.g., "Insight 1: Case 3, Insight 3: Case 1, Insight 4: Case 3. Insights 2 and 5 were skipped due to low quality.")

### Step 3: Compare the insights against existing tips as following:

If the playbook is empty, skip this step -- there are no existing tips to compare against.

If the playbook is non-empty, read each existing tip and ask:
- Does any insight ask to remove this tip? Only plan REMOVE if the insight gives concrete trajectory evidence that this tip is wrong or harmful, or if the same fact remains in an explicitly identified retained tip. A request to remove is not evidence by itself.
- Do any of the remaining insights (those that passed quality filtering in Step 2) relate to the same characteristic of this problem as this tip? Which insights?
- If yes, do those insights collectively provide more helpful information for solving this problem than this tip? Do they add concrete detail that this tip lacks?
- If yes, do any of those insights contradict this tip with corrected information?
- If no insights relate to this tip, this tip is unaffected.

Then ask: Are there any remaining insights that do not relate to any existing tip? Which ones? These point to potential new areas for the playbook.

### Step 4: Choose operations based on case and comparison as following:

Process each case group separately. Within each case, use the corresponding guidance below.

**Case 1 insights** (Generator succeeded on first attempt):

Overall tendency: conservative. No operation is the default.
- Is the current playbook adequate given the Generator's success? If yes, no operation.
- Preserve tips that helped the successful attempt. A success does not establish that a fresh Generator attempt will remember the fact without the playbook. REMOVE requires concrete evidence the tip is wrong or harmful, or an exact duplicate whose fact remains in an identified retained tip.
- Do any Case 1 insights reveal a key characteristic of this problem that is not yet covered in the playbook and that the Generator may not reliably attend to in future attempts? Which insights? If yes, consider ADD. If multiple insights point to the same characteristic, treat them as one and ADD once.
- Do any Case 1 insights provide more precise information about an existing tip? Even so, the Generator already succeeded with the current tip -- UPDATE is rarely justified.

**Case 2 insights** (Generator was initially incomplete but corrected itself):

Overall tendency: more active. ADD and UPDATE are the primary operations.
- Do any Case 2 insights describe a characteristic of this problem that was missed, and is not yet covered by any existing tip? Which insights? If yes, ADD a reminder for this point. If multiple insights describe the same characteristic, treat them as one and ADD once.
- Do any Case 2 insights relate to an existing tip that covers a related characteristic but misses this particular point? Which insights, which tip? If yes, UPDATE to expand coverage.
- Are what the insights describe minor or already covered by an existing tip? If yes, no operation.
- Did the Generator need two turns to get it right? This signals instability, not mastery -- REMOVE is almost never justified.

**Case 3 insights** (Generator failed due to knowledge gap, could not self-correct):

Overall tendency: most active. ADD is the primary operation.
- Do any Case 3 insights describe a knowledge gap about this problem that is not covered by any existing tip? Which insights? If yes, ADD -- this is the most important source of playbook content. If multiple insights describe the same gap, treat them as one and ADD once.
- Do any Case 3 insights relate to an existing tip that covers a related characteristic but was not sufficient to prevent this failure? Which insights, which tip? If yes, UPDATE to make it more specific or accurate.
- Is the knowledge gap already well-covered by an existing tip? If yes, no operation.
- The Generator failed -- REMOVE is almost never justified.

**Case 4 insights** (Generator failed, did not effectively revise):

Overall tendency: active but constrained. ADD and UPDATE are primary, but keep tip count small.
- Do any Case 4 insights describe a behavioral pattern not yet covered by any existing tip, with a clear description of what the Generator should have done instead? Which insights? If yes, ADD. If multiple insights describe the same pattern, treat them as one and ADD once.
- Do any Case 4 insights relate to an existing tip that covers a related behavioral pattern but its guidance is too vague? Which insights, which tip? If yes, UPDATE to make it more concrete.
- Is this behavioral pattern already covered by an existing tip? If yes, no operation -- the issue may be that the Generator is not following existing guidance, which cannot be solved by adding more tips.
- The Generator failed -- REMOVE is almost never justified.

### Step 5: Verify your output:

1. If the playbook is empty, did you choose UPDATE or REMOVE? That's invalid -- the only valid operations on an empty playbook are ADD or no operation.
2. If your operation is ADD, does it include only `type` and `content` with NO `tip_id`?
3. If your operation is UPDATE, does it include `type`, `tip_id`, and `content`?
4. If your operation is REMOVE, does it include only `type` and `tip_id`?
5. Is your content free of tip IDs like `[1]`? The system assigns IDs automatically.
6. Does every REMOVE cite concrete evidence that the tip is wrong or harmful, or identify an exact duplicate retained elsewhere? Similar subject matter alone is insufficient. Preserve complementary facts and correct tips used in a successful attempt.
7. Do any of your operations target the same tip or the same characteristic? If yes, merge them into one operation.

## Current Playbook

[PLAYBOOK BEGINS HERE]

{{current_playbook}}

[PLAYBOOK ENDS HERE]

## Reflector's Latest Insights

{{guidebook}}

## Output Format

Complete Steps 1 through 5 above before writing your JSON output. Make sure you have chosen your operations based on the case-specific guidance in Step 4 and verified your output in Step 5.

Respond with JSON only (no markdown code block). Your output may contain multiple operations from processing different insights. Use the format matching the operations you chose in Step 4:

ADD a new tip (no tip_id):

{
  "reasoning": "Why adding: what characteristic of this problem does the insight contribute. Which insights support this.",
  "operations": [
    {"type": "ADD", "content": "Tip content."}
  ]
}

UPDATE an existing tip (must reference tip_id):

{
  "reasoning": "Why updating: what is more helpful or corrected. Which insights support this.",
  "operations": [
    {"type": "UPDATE", "tip_id": "1", "content": "Improved content."}
  ]
}

REMOVE a tip (must reference tip_id):

{
  "reasoning": "Why removing: which tip and why. Which insights support this.",
  "operations": [
    {"type": "REMOVE", "tip_id": "1"}
  ]
}

No operation needed:

{
  "reasoning": "Why no change. Briefly explain which insights were considered and why none led to an operation.",
  "operations": []
}

Mixed operations (from processing different insights):

{
  "reasoning": "For each operation: why, and which insights support it.",
  "operations": [
    {"type": "ADD", "content": "Tip content from Case 3 insights."},
    {"type": "UPDATE", "tip_id": "2", "content": "Improved content from Case 2 insights."}
  ]
}

## Important Notes
- Playbook is located between [PLAYBOOK BEGINS HERE] and [PLAYBOOK ENDS HERE].
- Playbook is empty if and only if there's no content between [PLAYBOOK BEGINS HERE] and [PLAYBOOK ENDS HERE].
- Count tips in the playbook carefully. NEVER assume or guess the number of tips.
- Tips should be specific to this problem's key characteristics, NOT general tips applicable to all tasks.
- Do NOT include exact code in tip content. Tips are hints that provide marginal, additional, complementary help, NOT complete solutions.
- You are given multiple insights from different rollouts of the same problem. Always reference specific insights (Insight 1, Insight 2, ...) when explaining your reasoning. Do not treat all insights as a single undifferentiated block.
\end{lstlisting}

\subsection{AppWorld}
\label{app:prompts-appworld}

\paragraph{Actor.}
\begin{lstlisting}[style=prompt]
USER:

I am your supervisor and you are a super intelligent AI Assistant whose job is to achieve my day-to-day tasks completely autonomously with the help of a playbook.

## Environment Rules
1. To do this, you will need to interact with apps (e.g., spotify, venmo etc) using their associated APIs on my behalf. For this you will undertake a *multi-step conversation* using a python REPL environment. That is, you will write the python code and the environment will execute it and show you the result, based on which, you will write python code for the next step and so on, until you've achieved the goal. This environment will let you interact with appss using their associated APIs on my behalf.

2. Here are three key APIs that you need to know to get more information:

- To get a list of apps that are available to you
    print(apis.api_docs.show_app_descriptions())

- To get the list of apis under any app listed above, e.g. spotify
    print(apis.api_docs.show_api_descriptions(app_name='spotify'))

- To get the specification of a particular api, e.g. spotify app's login api
    print(apis.api_docs.show_api_doc(app_name='spotify', api_name='login'))

Each code execution will produce an output that you can use in subsequent calls.

3. Read **Current Playbook** to help you complete the task.

4. When you have completed the task, call `apis.supervisor.complete_task()` to finish:
  - If the task asks for information (e.g., "how many", "give me a list of"), pass the raw value as the answer: 
     `apis.supervisor.complete_task(answer=79)` or 
     `apis.supervisor.complete_task(answer="Song A, Song B")`
  - If the task asks you to perform an action (e.g., "send money", "delete a file"), 
     call `apis.supervisor.complete_task()` with NO argument after completing the action.
  - Never wrap answers in sentences. Pass the value directly.

## Examples

Here is an example of how to solve a task step-by-step. Notice how each step writes only ONE small code block, waits for the output, then proceeds to the next step.

[EXAMPLE BEGINS HERE]
AGENT: Okay. Lets first find which APIs are available to use in Spotify.
Code:
```
print(apis.api_docs.show_api_descriptions(app_name='spotify'))
```

ENVIRONMENT: ["login : Login to your account.", "logout : Logout from your account.", ...]

AGENT: Okay. Looks like I can use the `login` api. Lets find its specifications.
Code:
```
print(apis.api_docs.show_api_doc(app_name='spotify', api_name='login'))
```

ENVIRONMENT: {"app_name": "spotify", "api_name": "login", "description": "Login to your account. This is required before you can access any other API.", ...}

AGENT: Now I'll log in to spotify using my email and I need to find out my password. I can use the supervisor app to get my account password.
Code:
```
password = [acc for acc in apis.supervisor.show_account_passwords() if acc["account_name"] == "spotify"][0]["password"]
result = apis.spotify.login(username="user@example.com", password=password)
print(result)
```

ENVIRONMENT: {"access_token": "abc123...", "expires_in": 3600, "token_type": "Bearer"}

AGENT: Great, I am now logged in. Now I need to find the API to show my playlists.
Code:
```
print(apis.api_docs.show_api_doc(app_name='spotify', api_name='show_playlist_library'))
```

ENVIRONMENT: {"app_name": "spotify", "api_name": "show_playlist_library", "description": "Show user's playlist library.", ...}

AGENT: Now I'll get my playlists and count them.
Code:
```
playlists = apis.spotify.show_playlist_library(access_token=result['access_token'])
print(f"Number of playlists: {len(playlists)}")
apis.supervisor.complete_task(answer=len(playlists))
```
[EXAMPLE ENDS HERE]

## Current Playbook
You are provided with a playbook containing strategies learned from solving the same problem.

[PLAYBOOK BEGINS HERE]
{{ playbook }}
[PLAYBOOK ENDS HERE]
## Key Instructions

1. Write ONE small code block per step. Wait for the output before writing the next step.
2. Always check API specs using apis.api_docs.show_api_doc() before calling any API. Do not guess API names or parameters.
3. Use the "supervisor" app for your account info and credentials. Use the "phone" app for contacts.
4. If an API call fails, use apis.api_docs.show_api_descriptions() to discover what APIs exist. Never repeat failing code.
5. Many APIs return items in pages. Loop over `page_index` to get all results.
6. When calling complete_task(), follow Environment Rule 4: pass the raw value for information tasks (e.g., answer=79, answer="Song A, Song B"),  or call with no argument for action tasks. Never wrap answers in sentences.
7. Before calling complete_task(), always print and verify your result. If the result is empty or seems wrong (e.g., 0 items found), investigate before submitting.
8. Do NOT call complete_task() until you have successfully fetched and printed the data you need.
9. Playbook contains task-specific tips. Apply a tip if it clearly matches your current situation.
10. Playbook tips may be inaccurate or incomplete. Always verify against the actual API documentation before following a tip.

## Task

Using these APIs, generate code to solve the actual task. Start by discovering the available APIs. You can refer to playbook when necessary.

My name is: {{ main_user.first_name }} {{ main_user.last_name }}. My personal email is {{ main_user.email }} and phone number is {{ main_user.phone_number }}.
Task: {{ input_str }}
\end{lstlisting}

\paragraph{Reflector.}
\begin{lstlisting}[style=prompt]
You analyze one AppWorld attempt to help the next independent attempt solve the same task. Extract a precise, evidence-supported correction or a necessary invariant. The next attempt starts fresh; preserve useful facts even after a successful attempt.

You receive the original task, the current playbook, a trajectory summary, and a record of actions and environment replies. The original task defines the requested result. The agent's interpretation and existing tips are claims, not task requirements. Use visible action code, API documentation, and environment replies to check them. An unsuccessful outcome alone does not identify the faulty action.

## Analysis

1. Read the original task before the playbook or trajectory. Identify its entities, source of information, required actions, selection conditions, and action order. Keep every required part of the result. Then read the summary: outcome, submission status, turns used, and termination reason.
2. Find the earliest unresolved mismatch between the requested result and the actual attempt. Check whether an existing tip introduced that mismatch. Quote the task requirement and the exact tip, code, documented field, or short environment reply that supports your diagnosis. Distinguish an error that was corrected from a mistake that remained. Check the data source and its scope, the selection predicate, and how intermediate values were initialized, accumulated, or reused. A valid API response or a call to complete_task does not prove that the requested state was achieved.
3. State what the next attempt must do differently and the condition that must hold before submission. Correct a contradicted tip rather than adding checks around its wrong premise. Give the corrected operation, predicate, or code fragment when it follows from the supplied evidence. If an attempt succeeded, preserve the concrete operation or invariant needed to reproduce that success. If evidence is incomplete, name the unresolved fact and a specific check supported by the visible documentation instead of inventing a correction.

## Requirements for the handoff

- Put the complete actionable lesson in `key_insight`; the curator may not receive your `reasoning` or the trajectory. For each lesson, include: the relevant action or mistake; a short quote or exact fact supporting it; the correction or invariant; and the concrete check that would catch recurrence.
- Be precise about the operation that changes. "Reset the relevant accumulator before recomputing" is useful only when you identify the accumulator and the recomputation from the recorded code. "Use the right data" or "check the answer" is insufficient: name the observed data scope, predicate, or expected condition when available.
- Preserve distinctions in the task and documented APIs. A source for personal history is not evidence about a social feed. A returned result or an empty result establishes only what the documented request tests; do not turn it into an unsupported account-existence rule.
- Do not replace a task entity with an assumed entity type, omit a required action, or introduce a new prerequisite. A failed search, authentication attempt, or unmet prerequisite does not prove that the task is impossible or that submitting an explanation satisfies it. Separate a failed route from the goal, and propose a documented recovery check only when the supplied evidence supports one.
- Separate attempted values from values that worked. Do not promote a failed credential, parameter, or operation into a proven recipe. When diagnosing an execution error, quote the actual submitted code and reply; do not infer that a different operation caused it.
- Use only module names, API names, parameters, fields, identifiers, and literal values visible in the supplied record. Do not invent a missing API, corrected value, or hidden test requirement. A correction inferred from visible code must be labeled as inferred, not as a successful execution.
- Observations may be truncated or missing. Do not infer the omitted content. If the missing portion prevents diagnosis, say so and do not present a guessed fix as a fact.
- Prefer one to three distinct actionable lessons. Do not produce generic reminders, narrate every turn, repeat equivalent advice, or fabricate lessons to meet a count. If no specific lesson is supported, set `key_insight` to an empty string and explain the limit in `reasoning`.

## Episode Info

### Original Task

[ORIGINAL TASK STARTS HERE]
{{task_instr}}
[ORIGINAL TASK ENDS HERE]

### Current Playbook

These tips may contain mistakes. Check them against the original task and recorded evidence.

[CURRENT PLAYBOOK STARTS HERE]
{{playbook}}
[CURRENT PLAYBOOK ENDS HERE]

### Trajectory Summary

{{trajectory_summary}}

### Trajectory Record

[TRAJECTORY RECORD STARTS HERE]
{{conversation_history}}
[TRAJECTORY RECORD ENDS HERE]

## Output Format

Return one JSON object only, without markdown code blocks or additional fields. Copy the outcome as `SUCCESS` or `FAILED` from the trajectory summary. Keep `reasoning` and `key_insight` as strings.

{
  "outcome": "FAILED",
  "reasoning": "Observed outcome, the relevant action and evidence, and the supported diagnosis or its limits.",
  "key_insight": "Action or mistake: ... Evidence: ... Correction or invariant: ... Check before submission: ..."
}

If no specific lesson is supported:

{
  "outcome": "FAILED",
  "reasoning": "Explain why the supplied evidence supports no concrete new lesson. Use SUCCESS instead if that is the recorded outcome.",
  "key_insight": ""
}
\end{lstlisting}

\paragraph{Curator.}
\begin{lstlisting}[style=prompt,literate={–}{{-}}1]
You edit a per-task AppWorld playbook for the next independent attempt. Preserve concrete task facts, API signatures and parameters, required preconditions, safe workflows, and verification checks supported by the recorded evidence. The next attempt starts fresh; one successful attempt does not mean that later attempts have mastered the task.

You receive the original task, the current playbook, and the reflector's latest insights. The original task defines the requested result; existing tips and reflector interpretations do not add task requirements. You do not receive the full trajectory. Check each proposed rule against the original task and the specific evidence cited in the insights. Do not invent facts or treat the outcome alone as evidence for a new rule.

Before choosing an operation:
1. Identify the current tip IDs and the number of tips. An empty playbook has no valid UPDATE or REMOVE targets.
2. Compare the task's entities, information source, required actions, conditions, and action order with the playbook and each insight. Correct a tip that demonstrably contradicts the task instead of strengthening its wrong premise. For each insight, identify the observed action or mistake, the quoted or specific evidence, and the exact correction or invariant for the next attempt. A tip must change a concrete operation or check; skip generic reminders, unsupported diagnoses, and redundant claims. Preserve an uncertainty as an uncertainty, not a proven correction.
3. Compare supported insights with existing tips. Preserve helpful facts and prioritize the correction for an observed failure. Aim for 3-5 concise, nonduplicative actionable tips; keep fewer when fewer lessons are supported. Prefer one concise UPDATE that merges related details to several overlapping ADD operations. Preserve essential details when merging; never pad the playbook or drop a unique useful fact merely to meet the count.
4. Use ADD for a supported fact that is missing. Use UPDATE for a concrete correction or useful operational detail. Use REMOVE only when the original task or cited evidence shows the tip is incorrect, fully duplicated, or irrelevant. Merge useful details before removing a duplicate.
5. A success alone never justifies REMOVE. A useful tip followed during a successful attempt should be retained. A failure alone does not show an existing tip is incorrect; the agent may not have followed it.
6. Keep the exact supported operation, data scope, predicate, or initialization/reset rule in the tip instead of weakening it to "use the right API" or "verify the answer." Include the concrete condition to check before submission when the evidence supplies one. Use only module names, API names, parameters, fields, identifiers, and literal values supplied in the original task, current playbook, or supported insights; never fill a gap by guessing.
7. Do not add a prerequisite, account rule, or stopping rule without support from the original task or cited API evidence. A failed route does not prove the goal is impossible, and a failure explanation does not replace the requested action. Failed attempted values are not verified recipes. Keep every required task outcome; preserve the task's operation order.
8. If the insight is empty or generic, do not add or rewrite tips from speculation. You may correct or remove an existing tip when its text demonstrably conflicts with the original task; state the exact conflict in reasoning. Otherwise, if no supported new fact or correction changes the next attempt, return an empty operations list.

Use the canonical key `bullet_id` for UPDATE and REMOVE. Copy its value from an existing tip, without brackets. Do not use `tip_id`. Operations in one response refer to the IDs displayed below; do not predict renumbered IDs. Do not target the same tip more than once. ADD content must not contain an assigned tip ID.

## Original Task

[ORIGINAL TASK STARTS HERE]
{{task_instr}}
[ORIGINAL TASK ENDS HERE]

## Current Playbook

[PLAYBOOK BEGINS HERE]

{{current_playbook}}

[PLAYBOOK ENDS HERE]

## Reflector's Latest Insights

{{guidebook}}

## Output Format

Return JSON only. In `reasoning`, identify the evidence and explain why each change helps the next attempt. For every REMOVE, state the evidence that the tip is incorrect, fully duplicated, or irrelevant. The examples below show the allowed operation shapes; choose only the operations justified by this task.

{
  "reasoning": "Specific evidence and the change it supports.",
  "operations": [
    {"type": "ADD", "content": "Supported task-specific instruction."},
    {"type": "UPDATE", "bullet_id": "1", "content": "Corrected or more specific instruction."},
    {"type": "REMOVE", "bullet_id": "2"}
  ]
}

When no change is justified:

{
  "reasoning": "The supported lesson is already captured, or no actionable lesson is supported.",
  "operations": []
}
\end{lstlisting}

\subsection{MiniSWEBench}
\label{app:prompts-miniswe}

\paragraph{Actor.}
\begin{lstlisting}[style=prompt]
<CURRENT PLAYBOOK>
You are provided with a playbook of tips learned from earlier attempts at this same issue. Read it before acting. A tip that names a file, function, or behavior tells you where to look first: check it against the code, and follow it if the code agrees.

[PLAYBOOK BEGINS HERE]
{{ playbook }}
[PLAYBOOK ENDS HERE]
</CURRENT PLAYBOOK>

<BUDGET AND TOOLS>
These rules take priority over any longer procedure in the task description.
- You have a fixed number of turns (each observation shows "Steps Remaining"); one function call is one turn. Only your changes to non-test files under /testbed are graded. If the repository is unchanged when turns run out, you score 0.
- Plan the turns: find the relevant code within about 8 turns, make your first source edit by about turn 15, check it with a short script, and submit once the check passes. Do not write extra edge-case files or run the whole test suite.
- Long output is cut off. Never `view` a whole large source file and never `grep -r` all of /testbed. Search narrowly, for example `grep -rn "def function_name" /testbed/package --include=*.py | head -20`, then `view` with a `view_range` of at most 80 lines around the match.
- Never repeat a call you have already made; it returns the same output. If an output did not show what you needed, change the path, pattern, or view_range.
- Before `str_replace`, `view` the exact lines with a view_range and copy `old_str` from that output character for character, including indentation; keep it to the few lines you change plus one line of context so it is unique. If it fails with "old_str not found", view those lines again instead of retrying the same text.
- Submit only after a source edit returned "has been edited". Submitting an unchanged repository scores 0.
</BUDGET AND TOOLS>

<PLAYBOOK REMINDER>
- First review the playbook before solving the problem.
- Playbook contains question-specific tips. Apply a tip if it clearly matches your current problem.
</PLAYBOOK REMINDER>
\end{lstlisting}

\paragraph{Reflector.}
\begin{lstlisting}[style=prompt]
You are an expert code analyst. Your job is to follow a 3-step process below to read a Generator's coding trajectory, understand why it succeeded or failed, and propose a concrete and specific key insight that will help the Generator to keep succeeding or avoid failure in the next attempt at this same issue (e.g., which file and function the bug lives in, what behavior the fix must change, which search located the code quickly, which action wasted turns and what to do instead).

You are given four things, shown in <EPISODE INFO>:
1. The GitHub issue the Generator was asked to resolve.
2. A Trajectory Summary computed from the environment and the grader. It is always correct. It contains the Outcome, the grader's Verdict, whether the Generator submitted, the turns used out of the turn limit, tool-call counts, edit outcomes, the turn of the first successful edit, exact repeats of the previous action, whether a patch was produced, and the FAIL_TO_PASS / PASS_TO_PASS test counts when the run was graded.
3. The playbook the Generator was given for this attempt.
4. The Trajectory Record: each turn's action followed by the environment's reply. Replies are accurate but cut after about 1500 characters. The Generator's own reasoning is not shown.

An ideal attempt follows a 5-phase workflow within the turn limit: Locate - Reproduce - Apply fix - Verify - Submit. The Generator is already told to search narrowly, view at most 80 lines at a time, never repeat a call, edit by about turn 15, and submit once its check passes. An insight that only repeats these rules adds nothing; say instead which concrete action in this trajectory broke them and what the Generator should have run instead.

You must STRICTLY follow these 3 steps in <TASK>. Keep each step brief.

<TASK>

Step 1 - Extract Actual Facts from the Trajectory Summary. Copy them; do not recount:
1. What is the GitHub issue asking for, in one sentence?
2. Outcome and Verdict.
3. Did the Generator submit? How many turns were used?
4. Turn of the first successful edit, and the Repeated Actions line.
5. Patch Produced, and the FAIL_TO_PASS / PASS_TO_PASS counts if present.

Step 2 - Trace the trajectory and answer the following questions:
1. At which turns did the Generator locate, reproduce, apply, verify, and submit? Which phases never happened?
2. Where does the bug live? Name the file and function the Generator found (or should have found, judging from the issue and the replies).
3. If SUCCESS: what located the code quickly, and what edit made the tests pass? What may not be reliably reproduced next time?
4. If FAILED:
   - At which turn, and in which phase, did the Generator first go wrong?
   - Did it repeat an action, view a whole large file or grep the whole repository and receive a cut-off reply, or keep exploring without editing? Quote the action.
   - Did its edits fail (e.g. "old_str not found", "appears 2 times")? Which file and lines was it trying to change, and what did the actual file content look like in the replies?
   - Did it submit before any source edit succeeded?
   - If it patched: which behavior does the patch still get wrong, judging from the issue and the test counts?
5. Which playbook tip, if any, did the Generator follow? Did it help, or did it send the Generator the wrong way? Give the tip's number in brackets, e.g. [2].

Step 3 - Determine the Case from the Verdict:

- Case 1: Fixed cleanly. Verdict RESOLVED, and the Generator located, fixed, and verified without wasted turns.
- Case 2: Fixed with retries. Verdict RESOLVED, but the Generator wasted turns on wrong locations, failed edits, repeats, or unneeded work.
- Case 3: Patched but not fixed. Verdict WRONG_FIX or REGRESSION_ONLY: a real patch was graded and is wrong.
- Case 4: Failed to patch. Verdict NO_PATCH, PATCH_NOT_APPLIED, or UNGRADED: the Generator never produced a gradeable, on-target patch, typically because it ran out of turns exploring or looping.
- Case 5: Not graded. Verdict OTHER: the grading harness failed, and the outcome says nothing about the Generator.

</TASK>

<EPISODE INFO>

The GitHub issue the Generator was tasked to resolve.

[GITHUB ISSUE BEGINS HERE]
{{problem_statement}}
[GITHUB ISSUE ENDS HERE]

The trajectory summary.

[TRAJECTORY SUMMARY BEGINS HERE]
{{trajectory_summary}}
[TRAJECTORY SUMMARY ENDS HERE]

The playbook the Generator was given.

[PLAYBOOK BEGINS HERE]
{{playbook}}
[PLAYBOOK ENDS HERE]

The Generator's actions and the environment's replies.

[TRAJECTORY RECORD BEGINS HERE]
{{conversation_history}}
[TRAJECTORY RECORD ENDS HERE]

</EPISODE INFO>

<OUTPUT FORMAT>

Complete Step 1 through Step 3 above before writing your JSON output.

After completing the 3 steps, respond with a JSON object. Do NOT wrap it in a markdown code block. Choose the format matching your case from Step 3.

Rules for key_insight:
- Start it with "Verdict: <VERDICT>." copied from the summary.
- Then give one or two sentences of advice for the next attempt at this same issue. It must name a concrete file, function, or behavior, or a concrete replacement action (e.g. "grep for the definition of <function> under /testbed/<package>/<module> and view 60 lines around it").
- If a playbook tip misled the Generator, end with "Remove tip [N]: <reason>" and cite the concrete action or observed behavior that shows it was wrong or harmful.
- Preserve correct or helpful tips, including tips used in a successful attempt. Success does not update the Generator weights between attempts and does not show that the next attempt can recover the fact without advice. Never request removal merely because a tip was followed, validated, or appears mastered. An exact duplicate may be removed only if you identify the retained tip that carries the same fact.
- Do NOT include exact code or a literal patch, and do NOT name exact FAIL_TO_PASS / PASS_TO_PASS tests.
- Do NOT give advice about the environment, the container, or shell errors that the Generator did not cause.

Case 1 - Fixed cleanly:

{
  "outcome": "SUCCESS",
  "reasoning": "What facts did you extract in Step 1? What workflow did you identify in Step 2? Why is this Case 1?",
  "key_insight": "Verdict: RESOLVED. Where does the bug live and what did the fix change? Which step may not be reliably reproduced next time?"
}

Case 2 - Fixed with retries:

{
  "outcome": "SUCCESS",
  "reasoning": "What facts did you extract in Step 1? Which turns were wasted and why, as analyzed in Step 2? Why is this Case 2?",
  "key_insight": "Verdict: RESOLVED. Where does the bug live, and which wasted action should be replaced by what?"
}

Case 3 - Patched but not fixed:

{
  "outcome": "FAILED",
  "reasoning": "What facts did you extract in Step 1? Where did the patch go wrong, as analyzed in Step 2? Why is this Case 3?",
  "key_insight": "Verdict: WRONG_FIX (or REGRESSION_ONLY). Which behavior must the fix change, in which file and function, that this patch got wrong?"
}

Case 4 - Failed to patch:

{
  "outcome": "FAILED",
  "reasoning": "What facts did you extract in Step 1? At which turn and action did the attempt break down (e.g., repeats, whole-file views, exploring without editing), as analyzed in Step 2? Why is this Case 4?",
  "key_insight": "Verdict: NO_PATCH (or PATCH_NOT_APPLIED / UNGRADED). Where should the Generator look first, and which action should replace the one that wasted turns?"
}

Case 5 - Not graded:

{
  "outcome": "FAILED",
  "reasoning": "The grading harness failed; the Verdict is OTHER.",
  "key_insight": ""
}

</OUTPUT FORMAT>

<IMPORTANT>
- The Trajectory Summary is computed from the environment and the grader; trust it over anything the Generator's actions suggest.
- A good tip names the location or behavior of this specific bug. Generic workflow advice helps only when tied to a concrete action from this trajectory.
- A good tip should not begin with "The Generator...".
</IMPORTANT>
\end{lstlisting}

\paragraph{Curator.}
\begin{lstlisting}[style=prompt]
You are an editor for a per-bug SWE-bench fix playbook. The playbook is a cheatsheet for fixing this specific bug. It should be concise, but it should carry essential hints about the key characteristics of this problem (e.g., where the bug lives, what code paths are affected, what minimal change is needed, how the failing tests exercise the bug, common pitfalls when fixing it) that give marginal, additional, complementary help for the Generator to fix this bug on the next attempt. Tips in the playbook must NOT contain exact code.

You are given two things, shown below:
- The Current Playbook -- the accumulated playbook from prior epochs for this problem (in <CURRENT PLAYBOOK>).
- The Reflector Insights -- one or more supplied insights for this problem (in <REFLECTOR INSIGHTS>). Each insight includes an outcome (SUCCESS or FAILED) and a key insight that starts with the grader's Verdict. Process only the insights actually present; do not invent additional insights.

Follow the 5 steps in <TASK> strictly, in order, before producing your output.

<TASK>

Step 1 - Describe the current playbook.
1. Can you identify tips in the <CURRENT PLAYBOOK> section? If the playbook is blank, note that it is empty.
2. How many tips are in the Current Playbook? Count them carefully -- never guess. Aim for at most 5 tips. When the book is longer, merge exact duplicates while retaining their useful facts. Do not delete a correct tip merely to meet this target; future attempts still need the same evidence.

Step 2 - Evaluate each insight.
For each insight (Insight 1, Insight 2, ...):

First, decide whether to keep or reject it:
- Is the insight empty, missing, or not provided? If yes, reject it.
- Is the insight specific and actionable -- does it describe a key characteristic of this bug, not just restate the outcome? If not, reject it.
- Is the insight about the environment, the container, shell or harness errors, or a "corrupted" testbed rather than about this bug or the Generator's own actions? If yes, reject it.
- Is the insight a blanket ban on a basic tool (e.g. "avoid grep", "do not view files")? If yes, reject it; only a concrete replacement action (a narrower search, a view_range) is useful.
- Firewall: does the insight name an exact FAIL_TO_PASS / PASS_TO_PASS test, quote an exact test identifier (for example `test_x (module.ClassName)` or `path/to/test.py::Class::test_x`), or contain exact code or a literal patch? If yes, reject it -- such an insight would leak gold-test or solution information into the playbook that the Generator is not given. Do NOT create or update a tip from it.

Then, for each insight you kept, determine its Case. Each Key Insight starts with "Verdict: <VERDICT>." copied from the grader. Find that Verdict and use this lookup -- the same one the reflector itself used:
- Verdict OTHER -> Case 5 (Not graded): the run hit a grading-harness error. Reject this insight -- a harness error teaches nothing about the bug.
- Verdict RESOLVED -> Case 1 or Case 2. If the insight describes the Generator fixing the bug cleanly with no wrong turns, it is Case 1 (Fixed cleanly); if the fix was reached only after retries, a wrong first move, or redundant work, it is Case 2 (Fixed with retries).
- Verdict WRONG_FIX or REGRESSION_ONLY -> Case 3 (Patched but not fixed): the Generator delivered a real, graded patch that is wrong.
- Verdict NO_PATCH, PATCH_NOT_APPLIED, or UNGRADED -> Case 4 (Failed to patch): the Generator never delivered a gradeable, on-target patch.
If the Verdict is not stated in the insight, fall back to the `outcome` field (SUCCESS -> Case 1 or 2, FAILED -> Case 3 or 4, INCONCLUSIVE -> Case 5) and judge Case 3 vs 4 from the insight content (a wrong graded patch vs no usable patch).

After evaluating all insights, summarize: which insights did you keep, and which Case does each belong to? Which insights were rejected, and why? (e.g., "Insight 1: Case 3. Insight 3: Case 1. Insight 4: Case 3. Insight 2 rejected (empty). Insight 5 rejected (firewall -- named an exact test).")

Step 3 - Compare the kept insights against existing tips.
If the playbook is empty, there is nothing to compare against: every kept insight is a candidate for a new tip. Go to Step 4.
If the playbook is non-empty, read each existing tip and ask:
- Does any insight end with "Remove tip [N]: ..." for this tip? Only plan REMOVE if the insight gives concrete trajectory evidence that this tip is wrong or harmful, or if the same fact remains in an explicitly identified retained tip. A request to remove is not evidence by itself.
- Do any of the kept insights relate to the same characteristic of this bug as this tip? Which insights?
- If yes, do those insights collectively provide more helpful information than this tip -- do they add concrete detail it lacks?
- If yes, do any of those insights contradict this tip with corrected information?
- If no insights relate to this tip, this tip is unaffected.
Then ask: are there kept insights that do not relate to any existing tip? Which ones? These point to potential new areas for the playbook.

Step 4 - Choose operations based on Case and comparison.
First rule: if the playbook is empty and at least one insight was kept in Step 2, you MUST ADD one tip built from the most useful kept insight. An empty playbook gives the Generator no help on its next attempt.
Second rule: when a kept insight concerns the same file or function as an existing tip, prefer UPDATE of that tip over a new ADD.
Process each Case group separately. Case 5 insights were rejected in Step 2 and never reach this step.

Case 1 insights (Fixed cleanly):
- Is the current playbook adequate given the Generator's clean success? If yes, no operation.
- Preserve tips that helped the successful attempt. A success does not establish that a fresh Generator attempt will remember the fact without the playbook. REMOVE requires concrete evidence the tip is wrong or harmful, or an exact duplicate whose fact remains in an identified retained tip.
- Do any Case 1 insights reveal a key characteristic of this bug not yet in the playbook that a future attempt may not reliably attend to? If yes, consider ADD. If several insights point to the same characteristic, treat them as one and ADD once.
- UPDATE is rarely justified -- the Generator already succeeded with the current tips.

Case 2 insights (Fixed with retries):
- Do any Case 2 insights describe a characteristic of this bug that the Generator stumbled on, not yet covered by any tip? If yes, ADD a reminder. If several describe the same point, ADD once.
- Do any Case 2 insights relate to an existing tip that covers a related characteristic but misses this point? If yes, UPDATE to expand coverage.
- Are the insights minor or already covered? If yes, no operation.
- REMOVE is almost never justified.

Case 3 insights (Patched but not fixed):
- Do any Case 3 insights describe where the bug actually lives or what a correct fix must do, not covered by any tip? If yes, ADD -- this is the most important source of playbook content. If several describe the same gap, ADD once.
- Do any Case 3 insights relate to an existing tip that was not specific or accurate enough to prevent the wrong patch? If yes, UPDATE to sharpen it.
- Is the gap already well covered by an existing tip? If yes, no operation.
- REMOVE is almost never justified.

Case 4 insights (Failed to patch):
A Case 4 insight may be a system/tool workaround (a tool failed independent of whether the fix was correct) or a process correction (a budget or sequencing mistake) -- keep the tip as the insight frames it; do not convert it into a bug-knowledge claim.
- Do any Case 4 insights describe what this bug needs that the Generator did not deliver, with a concrete corrective action (e.g. "go straight to editing <area> instead of exploring", "reproduce the bug before editing"), not yet covered by any tip? If yes, ADD.
- Do any Case 4 insights relate to an existing tip whose guidance is too vague to have prevented the failure? If yes, UPDATE to make it concrete.
- Is the pattern already covered by an existing tip? If yes, no operation -- the Generator may simply not be following existing guidance, which adding more tips cannot fix.
- REMOVE is almost never justified.

Step 5 - Verify your output.
1. If the playbook is empty, did you choose UPDATE or REMOVE? That is invalid -- on an empty playbook the only valid operations are ADD or no operation.
2. If an operation is ADD, does it include only `type` and `content`, with NO `bullet_id`?
3. If an operation is UPDATE, does it include `type`, `bullet_id`, and `content`?
4. If an operation is REMOVE, does it include only `type` and `bullet_id`?
5. Is your content a plain JSON string containing the tip itself, free of tip IDs like `[1]`, the "Verdict: ..." prefix, labels such as "Tip:", and insight/case classifications such as "Insight 1: Case 3"? The system assigns IDs automatically. Classifications and operation explanations belong only in `reasoning`, never in tip `content`.
6. Does every REMOVE cite concrete evidence that the tip is wrong or harmful, or identify an exact duplicate retained elsewhere? Similar subject matter alone is insufficient. Preserve complementary facts and correct tips used in a successful attempt.
7. Do any of your operations target the same tip or the same characteristic? If yes, merge them into one operation.
8. Does any tip you ADD or UPDATE name an exact test or contain exact code? If yes, fix it -- tip content must never leak the gold tests or the solution.
9. Is each tip at most two sentences, and does it name where to look (file or function) or what behavior to change, or a concrete first action? If not, make it concrete or drop it.
10. Is your tip written from the insights rather than copied from this prompt's placeholders or generic rules the Generator already has (narrow grep, view_range, never repeat a call)? Spend the tip's words on this bug.

</TASK>

<CURRENT PLAYBOOK>

[PLAYBOOK BEGINS HERE]

{{current_playbook}}

[PLAYBOOK ENDS HERE]

</CURRENT PLAYBOOK>

<REFLECTOR INSIGHTS>

{{guidebook}}

</REFLECTOR INSIGHTS>

<OUTPUT FORMAT>

Complete the checks in Steps 1 through 5, then respond with exactly one top-level JSON object with two keys: `reasoning` and `operations`. Do not output the steps separately, add prose, or use a markdown code block.

- `reasoning` is a brief JSON string explaining the chosen operations and the supplied insights that support them. Put insight numbers and case classifications here only.
- `operations` is one flat JSON array. Each element is directly an operation object, not another object containing `reasoning` or `operations`.
- An ADD object has exactly `type` (the string `ADD`) and `content`.
- An UPDATE object has exactly `type` (the string `UPDATE`), `bullet_id`, and `content`.
- A REMOVE object has exactly `type` (the string `REMOVE`) and `bullet_id`.
- `bullet_id` is a JSON string identifying an existing tip in the Current Playbook.
- `content` is a plain JSON string with one or two sentences about this bug's location, required behavior, or a concrete first action supported by the insight. Never put an object, array, encoded JSON, case label, outcome restatement, or copied prompt placeholder in `content`. Do not include exact code or gold-test identifiers.

The single response shape is below. Fill in `reasoning` and put all chosen operation objects directly in the array. An empty array is valid only when the checks found no supported change; it is not the default when Step 4 requires an ADD.

{
  "reasoning": "",
  "operations": []
}

</OUTPUT FORMAT>

<IMPORTANT>
- The playbook is the content between [PLAYBOOK BEGINS HERE] and [PLAYBOOK ENDS HERE]. It is empty if and only if there is no content between those markers, or the only content is "(No strategies learned yet)".
- Count tips carefully. NEVER assume or guess the number of tips.
- Tips must be specific to THIS bug's key characteristics, NOT general SWE strategy that applies to all bugs.
- Do NOT include exact code in tip content. Tips are hints that give marginal, additional, complementary help -- not complete patches.
- Firewall: do NOT put exact FAIL_TO_PASS / PASS_TO_PASS test names or test identifiers into tip content. Tips are read by the Generator on its next attempt; naming the gold tests would hand it information it is not given. Describe the behavior a test requires, not the test's name.
- Reference the specific supplied insights (Insight 1, Insight 2, ...) when explaining your reasoning. Do not invent missing insights or treat the supplied insights as a single undifferentiated block.
</IMPORTANT>
\end{lstlisting}

\subsection{ERL Prompt Reviser}
\label{app:prompts-erl}

ERL uses the prompt-updater instruction of its release (copied verbatim from the pinned ERL release) on every benchmark. After a failed first attempt, the same model receives the system message below and a user message built from the current solver prompt, the attempt's reward, and its full conversation; the text inside \texttt{<prompt>} tags becomes the revised prompt $x'$ for the retry.

\paragraph{System message.}
\begin{lstlisting}[style=prompt]
You are an expert prompt updater. Analyze recent trajectories, rewards, and feedback to improve the solver system prompt. Return ONLY the revised prompt wrapped in <prompt>...</prompt> tags.
\end{lstlisting}

\paragraph{User message.}
\begin{lstlisting}[style=prompt]
## Current System Prompt
{current_prompt}

## Batch Metrics
- avg_first_reward: {reward}
- first_success_rate: 0.0000
- batch_size: 1.0000

## Recent Attempts
### Attempt 1
Trajectory:
- Turn 1 [{role}]: {content}
- Turn 2 [{role}]: {content}
...
Reward: {reward} | Correct: False
Feedback: INCORRECT | reward={reward} | termination={termination_reason}

Think step by step and provide an improved system prompt enclosed in <prompt>...</prompt> tags.
\end{lstlisting}

\clearpage
\section{Algorithmic Details}
\label{app:pipeline}

\subsection{From the Handoff Objective to Sparse Component Updates}
\label{app:design-rationale}

\label{app:credit-derivation}

\paragraph{From the future-only objective to delayed credit.}
We first derive the policy gradient of Eq.~\eqref{eq:ehrl-objective}, then identify the approximations used by \acerl{}. Write $s=(q,P)$, $X=R_q(\tau)$, and $Y=R_q(\tilde\tau)$ for the initial and subsequent actor rewards. The input distribution $s\sim\mu$ is held fixed for this local derivation. The objective concerns the value of one sampled candidate; selecting the best of several candidates defines a different outer execution law. Assume differentiable policies, parameter-independent environment dynamics and rewards, and conditions allowing differentiation under the expectation. All scores below are evaluated at the same policy, with role prompts and the deterministic formatter suppressed:
\begin{align}
  S_0&=\nabla_\theta\log\pi_\theta(\tau\mid q,P), &
  S_{\mathrm r}&=\nabla_\theta\log\pi_\theta(z\mid q,P,\tau),\nonumber\\
  S_{\mathrm c}&=\nabla_\theta\log\pi_\theta(u\mid P,z), &
  S_1&=\nabla_\theta\log\pi_\theta(\tilde\tau\mid q,P').
  \label{eq:app-scores}
\end{align}
An actor score sums over its generated action tokens; environment observations contribute no score. Factoring the execution probability into these four policy calls and applying the likelihood-ratio identity \citep{williams1992reinforce,sutton2018reinforcement} yields
\begin{equation}
  \nabla_\theta J_{\mathrm{ACE}}
  =\mathbb E\!\left[Y(S_0+S_{\mathrm r}+S_{\mathrm c}+S_1)\right].
  \label{eq:app-delayed-gradient}
\end{equation}
Parameter sharing does not merge these occurrences: the initial actor affects the evidence available to the handoff, whereas the subsequent actor must use the resulting playbook.

Define $Q_0=\mathbb E[Y\mid s,\tau]$, $Q_{\mathrm r}=\mathbb E[Y\mid s,\tau,z]$, and $Q_{\mathrm c}=\mathbb E[Y\mid s,\tau,z,u]$. For an admissible edit, $Q_{\mathrm c}=V_\theta(q,P')$, the expected reward of the actor using its candidate. By conditional expectation,
\begin{equation}
  \nabla_\theta J_{\mathrm{ACE}}
  =\mathbb E\!\left[Q_0S_0+Q_{\mathrm r}S_{\mathrm r}
                         +Q_{\mathrm c}S_{\mathrm c}+YS_1\right].
  \label{eq:app-conditional-gradient}
\end{equation}
Thus descendants provide delayed credit to each ancestor. For these identities, an inadmissible branch may be assigned terminal payoff zero and no evaluator score. We distinguish that convention from the implementation's admissible-only reflector average below.

\paragraph{A full comparison tree.}
Sample $N$ initial actor trajectories, $M$ reflections per trajectory, $C$ curator outputs per reflection, and $L$ evaluations per candidate. Let $Y_{ijk\ell}$ be a leaf reward. The nested estimates
\begin{equation}
  \widehat Q^{\mathrm c}_{ijk}=\frac1L\sum_{\ell=1}^L Y_{ijk\ell},\qquad
  \widehat Q^{\mathrm r}_{ij}=\frac1C\sum_{k=1}^C\widehat Q^{\mathrm c}_{ijk},\qquad
  \widehat Q^0_i=\frac1M\sum_{j=1}^M\widehat Q^{\mathrm r}_{ij}
  \label{eq:app-tree-values}
\end{equation}
are unbiased for their respective conditional values. Consequently,
\begin{align}
  \widehat g_{\mathrm{full}}={}&
  \frac1N\sum_i\widehat Q^0_iS_{0,i}
  +\frac1{NM}\sum_{ij}\widehat Q^{\mathrm r}_{ij}S_{\mathrm r,ij}\nonumber\\
  &+\frac1{NMC}\sum_{ijk}\widehat Q^{\mathrm c}_{ijk}S_{\mathrm c,ijk}
  +\frac1{NMCL}\sum_{ijk\ell}Y_{ijk\ell}S_{1,ijk\ell},
  \label{eq:app-full-tree-gradient}
\end{align}
with $\mathbb E[\widehat g_{\mathrm{full}}]=\nabla_\theta J_{\mathrm{ACE}}$. The result follows by applying the tower property to each term; leaves sharing ancestors need not be independent unconditionally. Siblings at every component share an input, providing the comparison groups for delayed, joint training. Subtracting an action-independent baseline preserves each expected score term. Choosing empirical group standardization and a clipped loss then gives a delayed joint \grpo{} \emph{surrogate} \citep{deepseekmath2024,ppo2017}; this choice is distinct from the exact identity in Eq.~\eqref{eq:app-full-tree-gradient}. Appendix~\ref{app:delayed-joint} shows that a chain-wise sampling pattern, one continuation per initial trajectory, is the special case $M=C=L=1$ of this estimator.

\paragraph{Conditional handoff learning and the current actor.}
The implemented actor trains on seed trajectories with their own reward $X$ and on candidate-evaluation trajectories with the candidate's reward $Y$. To isolate this credit choice, assume all outputs are admissible, omit filtering and normalization, and replace the implementation's token weighting by equal role-wise means. The resulting idealized direction is
\begin{align}
  g_{\mathrm{local}}&=\mathbb E[XS_0+YS_{\mathrm r}+YS_{\mathrm c}+YS_1],\nonumber\\
  \nabla J_{\mathrm{ACE}}-g_{\mathrm{local}}
  &=\mathbb E[(Y-X)S_0].
  \label{eq:app-local-credit-gap}
\end{align}
The candidate rollouts supply the direct reader term $YS_1$. In the implementation they share one normalized group with the seed rollouts, so their baseline is not specific to the candidate input. Current credit for the seed trajectory still requires an additional interpretation. Treat the initial trajectory as already collected evidence from a fixed checkpoint $\bar\theta$ and define
\begin{align}
  J_{\mathrm{cond}}(\theta;\bar\theta)
  &=\mathbb E_{\substack{s\sim\mu,\;\tau\sim\pi_{\bar\theta}\\
                       z,u,\tilde\tau\sim\pi_\theta}}[Y],\nonumber\\
  \left.\nabla_\theta J_{\mathrm{cond}}\right|_{\theta=\bar\theta}
  &=\mathbb E[Y(S_{\mathrm r}+S_{\mathrm c}+S_1)].
  \label{eq:app-conditional-objective}
\end{align}
The seed distribution is fixed only in this conditional objective; using a fixed checkpoint during rollout collection alone does not remove $S_0$ from Eq.~\eqref{eq:app-delayed-gradient}.

Playbook persistence explains how learning to use a \emph{retained} handoff can also be distributed across encounters. For a continuing sequence of the same task, let $P_i$ be the actual incumbent, $\tau_i$ its actor trajectory, and $R_i=R_q(\tau_i)$. The next input $P_{i+1}$ includes whatever selection and reset transition the harness performs. Pathwise,
\begin{equation}
  \frac1T\sum_{i=0}^{T-1}R_{i+1}
  =\frac1T\sum_{i=0}^{T-1}R_i+\frac{R_T-R_0}{T}.
  \label{eq:app-reward-reindex}
\end{equation}
Since rewards lie in $[0,1]$, the difference is at most $1/T$. When the long-run averages exist, they coincide without requiring identical marginal distributions of $P_i$ and $P_{i+1}$. At one fixed policy, write $S_{a,i}=\nabla_\theta\log\pi_\theta(\tau_i\mid q,P_i)$. The direct reader terms also satisfy
\begin{equation}
  \frac1T\sum_{i=0}^{T-1}R_{i+1}S_{a,i+1}
  =\frac1T\sum_{i=0}^{T-1}R_iS_{a,i}
    +\frac{R_TS_{a,T}-R_0S_{a,0}}{T}.
  \label{eq:app-score-reindex}
\end{equation}
The expected norm of the boundary vanishes if $\sup_i\mathbb E\|R_iS_{a,i}\|<\infty$. Hence a current encounter can supply the direct reader contribution of the preceding \emph{retained} handoff, with matching weighting. This is an identity for that gradient component, not a derivation of the full sequential gradient from convergence of reward averages.

In particular, persistence does not remove the actor's effect on future evidence. For the continuing finite-horizon objective $J_T=\mathbb E[\sum_{t=0}^T R_t]$, causality gives the contribution from incumbent actor trajectories
\begin{equation}
  g_{\mathrm{act},T}
  =\mathbb E\!\left[\sum_{i=0}^T S_{a,i}
                 \left(R_i+\sum_{t=i+1}^T R_t\right)\right].
  \label{eq:app-actor-return}
\end{equation}
Current-reward training omits the future-return term. A sufficient condition for its expectation to vanish is
$\mathbb E[\sum_{t=i+1}^T R_t\mid x_i,\tau_i]=b_i(x_i)$,
where $x_i$ includes the full relevant pre-action state; the weaker condition is orthogonality of future return to the actor score in expectation. Without such an assumption, treating the seed as fixed evidence is an explicit local approximation to the all-policy objective. Similarly, one-handoff rewards do not capture effects of reflection and curation beyond the next attempt.

There are three further limits to applying the temporal identity to training. First, the next encounter visits a retained winner or incumbent, whereas handoff credit uses all evaluated proposals; reindexing does not equate those distributions. Second, resets are valid transitions in Eq.~\eqref{eq:app-reward-reindex}, but a pre-reset proposal need not become the next input. Treating each reset interval as a separate proposal chain produces repeated boundary terms. Third, consecutive training encounters use changing policy checkpoints; identifying their score contributions requires an approximation that the policy changes slowly, or an explicit correction. Thus persistence motivates current actor training under the conditional interpretation, while it does not establish equality of the implemented update and the complete delayed gradient.

\paragraph{What sparse sampling preserves.}
For a fixed candidate with independent evaluation episodes,
\begin{equation}
  \widehat V_L=\frac1L\sum_{\ell=1}^L Y_\ell,\qquad
  \mathbb E[\widehat V_L\mid q,P']=V_\theta(q,P'),\qquad
  \operatorname{Var}(\widehat V_L\mid q,P')
       =\frac{\operatorname{Var}(Y\mid q,P')}{L}.
  \label{eq:app-one-evaluation}
\end{equation}
Setting $L=1$ preserves the conditional mean and weakly increases its variance. Replacing multiple curator descendants by one has the same property when inadmissible outcomes are counted consistently. Let $\mathcal T$ be a realized full tree, with reflector and curator group contributions $G_i^{\mathrm r}$ and $G_{ij}^{\mathrm c}$. Uniformly selecting $I\in[N]$ and $H\in[M]$ gives
\begin{equation}
  \mathbb E_I[G_I^{\mathrm r}\mid\mathcal T]
       =\frac1N\sum_iG_i^{\mathrm r},\qquad
  \mathbb E_{I,H}[G_{IH}^{\mathrm c}\mid\mathcal T]
       =\frac1{NM}\sum_{ij}G_{ij}^{\mathrm c}.
  \label{eq:app-group-sampling}
\end{equation}
These identities preserve raw score means with action-independent baselines, fixed scales, and matching role weights; sums require inverse inclusion-probability weights. Sampling an intact group also preserves the average of its normalized, clipped loss if that group's distribution is unchanged. Reducing the descendant samples used to compute its rewards does not satisfy that qualification.

\paragraph{Scope of equivalence for the implemented surrogate.}
Empirical \grpo{} normalization is nonlinear. Even mean centering with a sample's own group mean introduces a finite-group factor: for $G$ independent siblings with unbiased rewards and scores $S_i$,
\begin{equation}
  \mathbb E\!\left[\frac1G\sum_i(\widehat Q_i-\overline{\widehat Q})S_i\right]
  =\frac{G-1}{G}\,\mathbb E[QS].
  \label{eq:app-centering-factor}
\end{equation}
A leave-one-out mean removes this factor. Let $\sigma_Q$ be the sample standard deviation of the conditional values $Q_{1:G}$. Dividing by the sampled reward standard deviation further implies, in general,
\begin{equation}
  \mathbb E\!\left[\frac{\widehat Q_i-\overline{\widehat Q}}
                         {\widehat\sigma+\epsilon}\,\middle|\,Q_{1:G}\right]
  \ne \frac{Q_i-\bar Q}{\sigma_Q+\epsilon}.
  \label{eq:app-normalization-gap}
\end{equation}
Thus one evaluation retains the same surrogate form and an unbiased value estimate, but can change the expected normalized update as well as its variance. Clipping adds a further surrogate approximation; the raw score identities apply at the collection policy with clipping inactive and the stipulated credit and weighting.

The implementation also averages reflector reward only over admissible edits. If independent edits are admissible with probability $p$ and their mean conditional value is $v$, $n$ descendants have expected valid-only average $[1-(1-p)^n]v$, counting an empty valid set as zero. Its mean depends on $n$; counting invalid edits as zeros in the original denominator would instead give $pv$ for every $n$. Finally, a loss divided by the total retained token count changes role weights when branches are removed. Selecting among usable reflections and filtering groups can change inclusion probabilities as well. Exact sampling identities therefore do not imply equivalence of this random ratio to the full-tree estimator. The implemented method is a sparse component-wise \grpo{} surrogate with conditional handoff credit and current actor learning; its sampling and credit approximations are explicit rather than an asserted equality of expected gradients.

\paragraph{From the full tree to the sparse tree.}
The full tree above samples $M$ reflections of every actor rollout, $C$ edits of every reflection, and $L$ evaluations of every candidate. Four choices yield the implemented scheme.

\begin{itemize}[leftmargin=1.5em,itemsep=2pt]
\item \emph{One reflector group.} All $N$ seed rollouts are kept for actor training, but only one uniformly chosen seed is reflected on. This preserves a same-input reflector group and removes the factor $N$ from all downstream sampling.
\item \emph{One curator group.} One uniformly chosen usable reflection receives $C$ edits and every other usable reflection receives one. This preserves a same-input curator group and still supplies a downstream value for every reflection.
\item \emph{One evaluation per candidate.} Replacing an average of $L$ evaluations by one preserves the candidate's conditional expected verifier reward, at higher conditional variance. Empirical standardization and the implementation choices discussed above qualify equivalence of the resulting updates.
\item \emph{Candidate rollouts as actor rollouts.} The one rollout that scores each candidate is also added to the actor group, so the actor is trained on the $N$ seed rollouts and on up to $M-1+C$ candidate rollouts. This keeps the actor-rollout budget of \grpo{}, but the actor group then mixes inputs $(q,P)$ and $(q,P')$, so its normalization is no longer a same-input comparison.
\end{itemize}

Per task and training step, the full tree with $(N,M,C,L)=(4,2,3,1)$ generates $4$ seed rollouts, $8$ reflections, $24$ curator edits, and $24$ candidate evaluations, $60$ outputs in total. The reductions bring this to $4+2+4+4=14$ outputs for \acerl{}, of which $8$ are actor rollouts, the same number as a \grpo{} step with eight rollouts per task. The chain-wise design of Appendix~\ref{app:delayed-joint} uses $8+8+8+8=32$. Reflector and curator outputs add generation relative to \grpo{}, but they are short, and the actor-update compute per step is within $0.91$--$1.21\times$ that of \grpo{} (Appendix~\ref{app:compute}).

\subsection{Playbook Selection, Reset, and the Training Step}
\label{app:selection}
\label{sec:selection}

Training credit and memory selection are different decisions. The rewards above rank sampled outputs for learning; separately, the harness chooses which playbook task $q$ carries into its next encounter by Eq.~\eqref{eq:selection-rule} (Section~\ref{sec:sparse-tree}): among admissible candidates whose value is at least the seed's reward, the one with the highest value is kept (ties broken uniformly), and otherwise $P$ is kept. Rejecting an update does not erase its evaluated rewards; the corresponding outputs remain eligible for training subject to group filtering. Because edits accepted under early, weak parameters can persist and mislead later actors, we also reset each task's playbook to $P_0$ every $F$ training encounters of that task ($F=3$ in all main experiments). The reset bounds the age of retained experience and repeatedly exposes training to the empty-playbook condition that evaluation starts from. Appendix~\ref{app:design-rationale} derives the sampling tree of Algorithm~\ref{alg:acerl-main} as a reduction of the full tree: one evaluation per candidate preserves its expected verifier value while increasing estimation noise, and the finite-sample normalization need not preserve the full tree's expected gradient.

\subsection{Implementation Details}

\paragraph{Sampling and provenance.}
The main configuration samples $N=4$ seed rollouts per task with the current playbook, selects one seed uniformly, samples $M=2$ reflections of the seed, selects one usable reflection uniformly to receive $C=3$ curator samples, and gives the other usable reflection one curator sample. With both reflections usable this yields 4 seed, 2 reflector, and 4 curator outputs and up to 4 candidate playbooks, each scored by one fresh actor episode on a newly initialized environment for the same task. These candidate episodes join the seed rollouts in the task's actor group, so the actor trains on up to 8 rollouts per task, as in \grpo{}; the actor mini-batch of 64 rollouts also matches \grpo{}. Every candidate records its source task, seed trajectory, and reflection, so downstream values can be attributed to the outputs that produced them. Reflector and curator generations use the same sampling temperature ($0.7$) as the actor.

\paragraph{Reflector input.}
The benchmark adapter formats the seed trajectory for the reflector as a \emph{trajectory summary} (outcome, tests passed or verifier result, number of turns, termination reason) followed by the full \emph{trajectory record} (the actor's messages interleaved with environment observations), truncated to a history budget of 20k tokens. The final environment reply, such as the last test failure or the grader's verdict, is always kept when the history is clipped. On Mastermind, MBPP, and MiniSWEBench the reflector follows a three-step analysis (extract facts, explain the actor's behaviour, and classify the case); on AppWorld it checks the requested result against the recorded attempt, identifies the earliest unresolved mismatch, and supplies an evidence-supported correction and a check before submission. Reflectors return a JSON object with \texttt{outcome}, \texttt{reasoning}, and \texttt{key\_insight}. Prompts are listed in Appendix~\ref{app:prompts}.

\paragraph{Playbook and curator edits.}
A playbook is a list of lines \texttt{[id] tip}. The curator returns a JSON object with \texttt{reasoning} and a list of \texttt{operations}: \texttt{ADD} (new tip), \texttt{UPDATE} (rewrite the tip with a given \texttt{bullet\_id}), or \texttt{REMOVE} (by \texttt{bullet\_id}). Its prompt asks it to keep supported tips unless the evidence contradicts them. Operations are applied in order, duplicates are removed, and ids are renumbered. A curator output is \emph{admissible} when it parses, contains at least one operation that applies to the current playbook, changes the playbook, and the resulting playbook is at most 2,048 tokens; otherwise it receives reward $0$, is excluded from the set $\mathcal C(z)$ used to average its reflection's reward, and produces no candidate. The initial playbook is empty on every benchmark.

\paragraph{Comparison groups and filtering.}
The seed and candidate rollouts of a task share the actor group; the two reflections share a group keyed by the seed trajectory; the three curator outputs of the selected reflection share a group keyed by that reflection; the other curator output is a singleton. Before normalization, any group whose rewards are all $\le0$ or all $\ge1$ is dropped, and singletons are dropped with it under binary rewards. For MiniSWEBench, whose shaped rewards can be equal but strictly between $0$ and $1$, the filter instead requires a nonzero within-group range. With the threshold filter and fractional rewards, as in MBPP and Mastermind, a singleton can survive; the backend then uses mean $0$ and scale $1$, producing an outcome-weighted update rather than a group comparison. Constant multi-output groups that survive the threshold filter have zero centered advantage. Rollout batches are over-generated by a factor of two so that enough informative groups survive the filter \citep{dapo2025}. Reflector and curator groups are admitted to the update whether or not the selected playbook improves on the seed.

\paragraph{Objective.}
Each retained output $i$ receives the scalar advantage $A_i=(r_i-\mu_G)/(\sigma_G+\epsilon)$, where $\mu_G$ and $\sigma_G$ are the mean and standard deviation of rewards in its role's group $G$ (Section~\ref{sec:sparse-tree}), broadcast over its generated tokens. Writing $\rho_{i,t}(\theta)=\pi_\theta(y_{i,t}\mid x_{i,t})/\pi_{\theta_{\mathrm{old}}}(y_{i,t}\mid x_{i,t})$ and $M_{i,t}$ for the loss mask over generated tokens, the shared policy maximizes
\begin{equation}
  \mathcal L_{\mathrm{GRPO}}(\theta)=\frac{\sum_{i\in\mathcal B}\sum_t M_{i,t}\,\ell_{\mathrm{clip}}\bigl(\rho_{i,t}(\theta),A_i\bigr)}{\sum_{i\in\mathcal B}\sum_t M_{i,t}},\qquad
  \mathcal B=\mathcal B^{\mathrm{act}}\cup\mathcal B^{\mathrm{ref}}\cup\mathcal B^{\mathrm{cur}},
  \label{eq:shared-objective}
\end{equation}
with the asymmetric clipped surrogate $\ell_0(\rho,A)=\min\{\rho A,\operatorname{clip}(\rho,1-\delta_-,1+\delta_+)A\}$, $(\delta_-,\delta_+)=(0.2,0.28)$ \citep{dapo2025}, and a dual clip $\max\{\ell_0,cA\}$ with $c=3$ for $A<0$. Token-mean aggregation means roles are weighted by their token counts; we do not impose explicit role weights. There is no KL penalty, no entropy bonus, and no critic. Candidate values and advantages are constants with respect to $\theta$.

\paragraph{Selection and reset.}
Playbook selection follows Eq.~\eqref{eq:selection-rule} (Appendix~\ref{app:selection}). Each task's playbook is reset to empty before its 3rd, 6th, \ldots training encounter; the counter is per task, not per optimizer step, and the evaluation store is never reset by the training schedule. An AppWorld ablation that resets every 8 encounters reaches RSI@3 of $0.74$ at step 490 and $0.63$ at step 500, against $0.70$ and $0.67$ with $F=3$; with one run each, the two settings are not distinguished.

\paragraph{Hyperparameters.}
All methods use the same actor rollout count, token budgets, learning rate, and clip settings within a benchmark. The outputs of all trained roles are concatenated into one batch and updated jointly with Eq.~\eqref{eq:shared-objective}; the PPO-based methods replace the group-relative advantage with GAE from a critic. Methods differ in their advantage and sampling design. The algorithmic differences include the advantage estimator (group-relative or GAE), the group filter (off for critic-based methods and ERL), the handoff structure (per-role groups for \acerl{}; complete chains credited with their downstream reward for delayed-joint PPO; prompt rewriting and internalization for ERL), and the actor training source (seed and candidate rollouts for \acerl{}; all four chain segments for delayed-joint PPO). Delayed-joint PPO uses 6 tasks per step on MBPP and 8 elsewhere. ERL follows its released recipe, with a KL loss coefficient of $0.001$ and 32-task mini-batches. Table~\ref{tab:hyperparams} lists shared optimizer and rollout settings; Table~\ref{tab:benchmark-settings} lists per-benchmark budgets.

\begin{table}[t]
  \caption{Shared training hyperparameters. Baseline-specific settings: PPO and delayed-joint PPO add a critic initialized from the policy (learning rate $10^{-5}$, GAE with $\gamma=1$, $\lambda=0.95$, 10 warm-up steps) and disable group filtering so the critic sees constant-reward batches.}
  \label{tab:hyperparams}
  \centering
  \small
  \resizebox{\linewidth}{!}{%
\begin{tabular}{ll}
    \toprule
    Setting & Value \\
    \midrule
    Tasks per step / actor rollouts per task & 8 / 8 \\
    \acerl{} seed rollouts $N$; reflector, curator samples $(M,C)$ & 4; 2, 3 \\
    Candidate evaluations per candidate & 1 \\
    Playbook reset interval $F$ & 3 encounters \\
    Playbook length cap & 2,048 tokens \\
    Sampling temperature & 0.7 (MiniSWEBench eval actor: 0.6) \\
    Actor learning rate & $10^{-6}$ \\
    PPO epochs per batch / actor mini-batch & 1 / 64 rollouts \\
    Clip $(\delta_-,\delta_+)$, dual clip $c$ & $(0.2,0.28)$, 3 \\
    KL loss / entropy bonus & none / none \\
    Loss aggregation & token mean \\
    Group over-generation factor & 2 \\
    Training steps & 300 (\acerl{} and \grpo{} on AppWorld: 500) \\
    Turns per actor episode & 20 (MBPP: 6) \\
    Checkpoint and evaluation interval & every 10 steps \\
    Training runs per method & 1 \\
    Hardware per run & 2 GPUs (H100 or H200; H200 for AppWorld and MiniSWEBench) \\
    \bottomrule
  \end{tabular}}

\end{table}
\section{Benchmarks and Reward Design}
\label{app:benchmarks}

We use four benchmarks that span short feedback loops and long-horizon tool use (Table~\ref{tab:benchmark-settings}). All roles share one model per benchmark, the playbook is initially empty, and every reward lies in $[0,1]$. The information a playbook can carry differs across benchmarks: implied constraints on the hidden code, failing assertions and edge cases, API dependencies and entity bookkeeping, and relevant files and failure causes. The 0.6B model is small enough that recursive self-improvement is weak before training (Figure~\ref{fig:motivation}); the 4B instruct model tests whether an already post-trained model shows the same train--inference mismatch.

\begin{table*}[t]
  \caption{\textbf{Benchmark settings.} Turn limits are per actor episode; timeouts are per training trajectory.}
  \label{tab:benchmark-settings}
  \centering
  \small
  \setlength{\tabcolsep}{4pt}
  \resizebox{\textwidth}{!}{%
\begin{tabular}{L{0.15\linewidth}L{0.15\linewidth}L{0.27\linewidth}cccL{0.19\linewidth}}
    \toprule
    Benchmark & Model & Reward per episode & Turns & Timeout & Train / held-out & Playbook content \\
    \midrule
    Mastermind & Qwen3-0.6B (non-thinking) & best guess score $(e+v/2)/L$; $1$ when solved & 20 & 30 min & 200 / 100 & constraints implied by prior guesses \\
    MBPP & Qwen3-0.6B (non-thinking) & best fraction of the 3 unit tests passed & 6 & 30 min & 374 / 100 & failing assertions, edge cases \\
    AppWorld & Qwen3-4B-Instruct-2507 & $1$ iff the task's state checks pass & 20 & 20 min & 90 / 57 (dev) & API dependencies, entity bookkeeping \\
    MiniSWEBench & Qwen3-4B-Instruct-2507 & $0$ / $0.05$ / $0.2$ / $1$ (see text) & 20 & 2 h & 100 / 100 (Verified) & relevant files, failure causes \\
    \bottomrule
  \end{tabular}}
\end{table*}

\paragraph{Mastermind.}
Each game draws a secret code of length $L\in\{3,4,5\}$ over digits $\{0,\ldots,D-1\}$ with $D\in\{4,5,6\}$, from a fixed generator seed; the 200 training and 100 held-out games share no seeds. The agent submits $L$ space-separated digits per turn and observes the number of exact matches $e$ (right digit, right position) and value matches $v$ (right digit, wrong position, with multiset accounting). A guess scores $(e+v/2)/L$, so a solved game scores $1$; a malformed guess scores $0$ and returns an error observation. The episode reward is the best guess score, and a game ends on an exact match or after 20 turns.

\paragraph{MBPP.}
We use the full MBPP release \citep{mbpp2021}, training on its 374-problem train split and evaluating on the first 100 problems of its test split. Each problem has three unit tests. The actor submits a program in a code block each turn; each assert runs in a subprocess with a 10-second timeout, and a turn's reward is the fraction of asserts that pass. After a failed turn the environment shows up to three failing asserts with their error messages, and the episode ends when all tests pass or after six turns. The episode reward is the best turn reward, so held-out rewards are multiples of $1/3$ per task. Because the displayed asserts are the ones that are scored, a harness can carry them forward; every method sees the same feedback. A first delayed-joint PPO run learned to end the test process with a zero exit code before a failing assertion could fail; the grader now counts a test only if a random marker printed after its assertion appears, and that run is excluded. All reported MBPP evaluations except those of delayed-joint \acerl{} and the delayed-joint PPO rerun predate this fix; \acerl{} and PPO evaluation trajectories at steps 170, 250, and 300 contain no exit calls.

\paragraph{AppWorld.}
The actor writes Python that calls app APIs through the AppWorld server and finishes by calling the task's completion API \citep{appworld2024}. The episode reward is $1$ if AppWorld's state-based evaluator reports success for the task and $0$ otherwise, including when the actor runs out of its 20 turns. We train on the 90 tasks of the train split and evaluate on the 57 tasks of the dev split.

\paragraph{MiniSWEBench.}
Tasks are GitHub issues solved in the mini-SWE-agent scaffold\footnote{\url{https://github.com/SWE-agent/mini-swe-agent}} \citep{sweagent2024}: the actor inspects and edits a repository through shell commands in a container and submits within 20 actor turns. Training uses 100 instances of a SWE-Gym subset and evaluation the first 100 instances of SWE-bench Verified \citep{swebench2023}, both graded with the official evaluation harness when the episode ends. The shaped reward is $0$ when the final diff is empty, $0.05$ when a diff exists but grading fails to run, $0.2$ when the diff is graded but the issue is not resolved (including diffs that do not apply), and $1$ when all fail-to-pass and pass-to-pass tests pass. This shaping applies to training. Held-out evaluations report binary task success: $1$ only when the native test report marks the issue resolved, and $0$ otherwise.

\paragraph{Rollout accounting at evaluation.}
Vanilla evaluation draws 8 independent actor episodes per task. Harnessed evaluation runs 8 sequential visits per task, each with one actor episode, one reflector call, and one curator call; no additional rollout scores the curator's edit. RSI@$K$ and pass@$K$ therefore use the same $K$ actor episodes, and RSI@$K$ additionally uses $K$ reflector and $K$ curator calls.
\section{Additional Results and Diagnostics}
\label{app:additional-results}

\subsection{Training Compute and Checkpoint Selection}
\label{app:compute}

Checkpoints are evaluated every 10 steps (5 for MiniSWEBench). Every comparison in the paper uses cumulative update FLOPs as its training axis, as recorded by the shared VeRL estimator in each trainer's log, for the actor and, for PPO-based methods, the critic; generation is not included. Up to the compute budget, \acerl{} uses on average $1.14\times$, $1.01\times$, $1.17\times$, and $1.21\times$ the update FLOPs per step of \grpo{} on Mastermind, MBPP, AppWorld, and MiniSWEBench.

\subsection{Training Reward}
\label{app:training-reward}

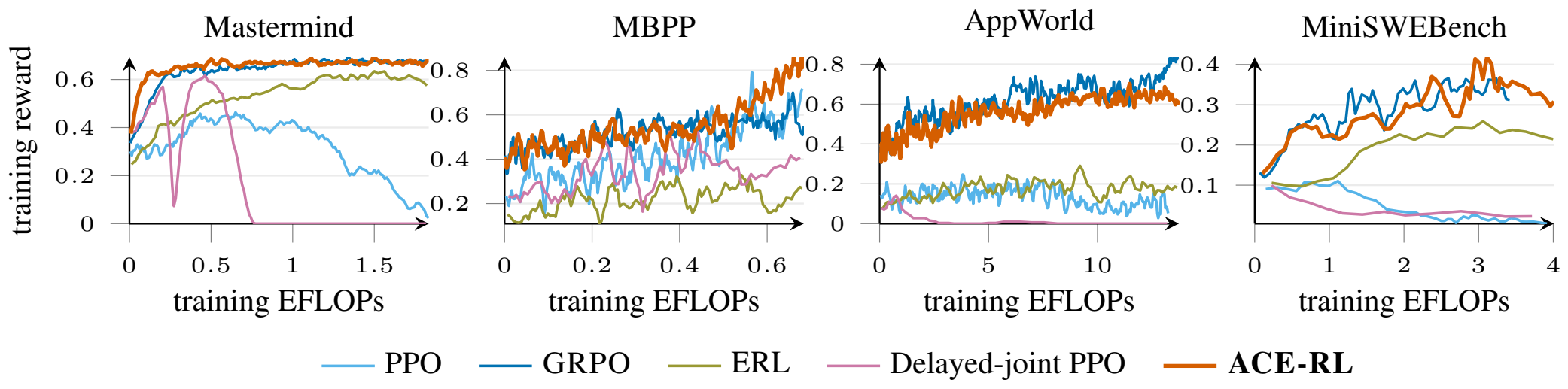
\begin{figure}[H]
  \centering
\resultsfigstyle
\begin{scaledfigure}[1pt]
\begin{tikzpicture}
\begin{groupplot}[group style={group size=4 by 1, horizontal sep=0.6cm}, panel, width=3.95cm, height=2.9cm]
\nextgroupplot[title={Mastermind}, xmin=0, xmax=1.8335, restrict x to domain=0:1.8335, xlabel={training EFLOPs}, ylabel={training reward}, legend to name=trainleg]
  \addlegendimage{m ppo}\addlegendentry{PPO}
  \addlegendimage{m grpo}\addlegendentry{\textsc{GRPO}}
  \addlegendimage{m erl}\addlegendentry{ERL}
  \addlegendimage{m jppo}\addlegendentry{Delayed-joint PPO}
  \addlegendimage{m ace}\addlegendentry{\textbf{\textsc{ACE-RL}}}
  \addplot[forget plot, m ppo] table[x=ppo_ef, y=ppo, col sep=comma] {data/campaign_train_mastermind.csv};
  \addplot[forget plot, m grpo] table[x=grpo_ef, y=grpo, col sep=comma] {data/campaign_train_mastermind.csv};
  \addplot[forget plot, m erl] table[x=erl_ef, y=erl, col sep=comma] {data/campaign_train_mastermind.csv};
  \addplot[forget plot, m jppo] table[x=joint_ppo_ef, y=joint_ppo, col sep=comma] {data/campaign_train_mastermind.csv};
  \addplot[forget plot, m ace] table[x=ace_rl_ef, y=ace_rl, col sep=comma] {data/campaign_train_mastermind.csv};
\nextgroupplot[title={MBPP}, xmin=0, xmax=0.6838, restrict x to domain=0:0.6838, xlabel={training EFLOPs}]
  \addplot[forget plot, m ppo] table[x=ppo_ef, y=ppo, col sep=comma] {data/campaign_train_mbpp.csv};
  \addplot[forget plot, m grpo] table[x=grpo_ef, y=grpo, col sep=comma] {data/campaign_train_mbpp.csv};
  \addplot[forget plot, m erl] table[x=erl_ef, y=erl, col sep=comma] {data/campaign_train_mbpp.csv};
  \addplot[forget plot, m jppo] table[x=joint_ppo_ef, y=joint_ppo, col sep=comma] {data/campaign_train_mbpp.csv};
  \addplot[forget plot, m ace] table[x=ace_rl_ef, y=ace_rl, col sep=comma] {data/campaign_train_mbpp.csv};
\nextgroupplot[title={AppWorld}, xmin=0, xmax=13.7373, restrict x to domain=0:13.7373, xlabel={training EFLOPs}]
  \addplot[forget plot, m ppo] table[x=ppo_ef, y=ppo, col sep=comma] {data/campaign_train_appworld.csv};
  \addplot[forget plot, m grpo] table[x=grpo_ef, y=grpo, col sep=comma] {data/campaign_train_appworld.csv};
  \addplot[forget plot, m erl] table[x=erl_ef, y=erl, col sep=comma] {data/campaign_train_appworld.csv};
  \addplot[forget plot, m jppo] table[x=joint_ppo_ef, y=joint_ppo, col sep=comma] {data/campaign_train_appworld.csv};
  \addplot[forget plot, m ace] table[x=ace_rl_ef, y=ace_rl, col sep=comma] {data/campaign_train_appworld.csv};
\nextgroupplot[title={MiniSWEBench}, xmin=0, xmax=4.0096, restrict x to domain=0:4.0096, xlabel={training EFLOPs}]
  \addplot[forget plot, m ppo] table[x=ppo_ef, y=ppo, col sep=comma] {data/campaign_train_miniswe.csv};
  \addplot[forget plot, m grpo] table[x=grpo_ef, y=grpo, col sep=comma] {data/campaign_train_miniswe.csv};
  \addplot[forget plot, m erl] table[x=erl_ef, y=erl, col sep=comma] {data/campaign_train_miniswe.csv};
  \addplot[forget plot, m jppo] table[x=joint_ppo_ef, y=joint_ppo, col sep=comma] {data/campaign_train_miniswe.csv};
  \addplot[forget plot, m ace] table[x=ace_rl_ef, y=ace_rl, col sep=comma] {data/campaign_train_miniswe.csv};
\end{groupplot}
\node[anchor=north] at ([yshift=-0.75cm]$(group c2r1.south)!0.5!(group c3r1.south)$) {\pgfplotslegendfromname{trainleg}};
\end{tikzpicture}
\end{scaledfigure}
  \caption{\textbf{Training reward}: mean reward of the rows in each trainer's update batch (centered mean over five steps). Actor-only methods act with an empty playbook. For \acerl{} and delayed-joint PPO the batch also contains reflector and curator outputs, scored by candidate values; ERL's batch contains its guided retries. Values are therefore not comparable across methods. The x-axis is training EFLOPs, up to the compute budget; stopped runs end at their last logged step.}
  \label{fig:training-reward}
\end{figure}

\subsection{Critic Diagnostics}
\label{app:critic}

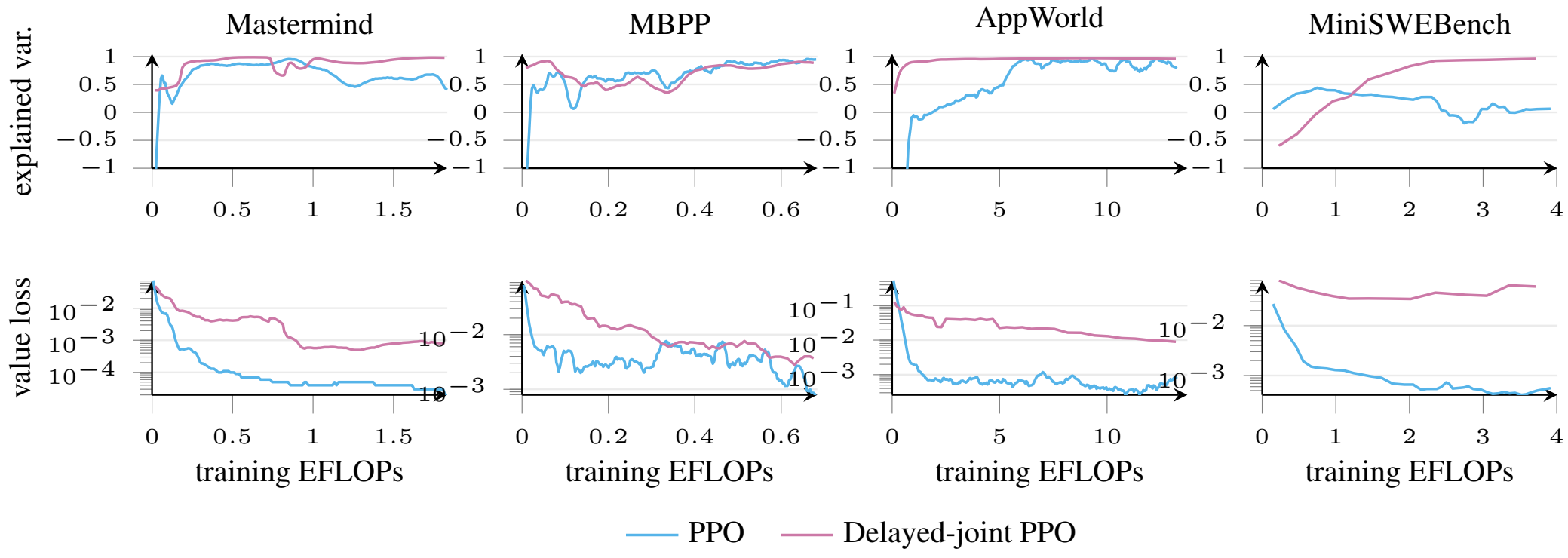
\begin{figure}[H]
  \centering
\resultsfigstyle
\begin{scaledfigure}[1pt]
\begin{tikzpicture}
\begin{groupplot}[group style={group size=4 by 2, horizontal sep=0.6cm, vertical sep=0.9cm}, panel, width=3.95cm, height=2.5cm, yticklabel style={/pgf/number format/fixed,/pgf/number format/precision=2}]
\nextgroupplot[xmin=0, xmax=1.8335, restrict x to domain=0:1.8335, ymin=-1, ymax=1.05, title={Mastermind}, ylabel={explained var.}, legend to name=critleg]
  \addlegendimage{m ppo}\addlegendentry{PPO}
  \addlegendimage{m jppo}\addlegendentry{Delayed-joint PPO}
  \addplot[forget plot, m ppo] table[x=ppo_ef, y=ppo_ev, col sep=comma] {data/campaign_diag_critic_mastermind.csv};
  \addplot[forget plot, m jppo] table[x=joint_ppo_ef, y=joint_ppo_ev, col sep=comma] {data/campaign_diag_critic_mastermind.csv};
\nextgroupplot[xmin=0, xmax=0.6838, restrict x to domain=0:0.6838, ymin=-1, ymax=1.05, title={MBPP}]
  \addplot[forget plot, m ppo] table[x=ppo_ef, y=ppo_ev, col sep=comma] {data/campaign_diag_critic_mbpp.csv};
  \addplot[forget plot, m jppo] table[x=joint_ppo_ef, y=joint_ppo_ev, col sep=comma] {data/campaign_diag_critic_mbpp.csv};
\nextgroupplot[xmin=0, xmax=13.7373, restrict x to domain=0:13.7373, ymin=-1, ymax=1.05, title={AppWorld}]
  \addplot[forget plot, m ppo] table[x=ppo_ef, y=ppo_ev, col sep=comma] {data/campaign_diag_critic_appworld.csv};
  \addplot[forget plot, m jppo] table[x=joint_ppo_ef, y=joint_ppo_ev, col sep=comma] {data/campaign_diag_critic_appworld.csv};
\nextgroupplot[xmin=0, xmax=4.0096, restrict x to domain=0:4.0096, ymin=-1, ymax=1.05, title={MiniSWEBench}]
  \addplot[forget plot, m ppo] table[x=ppo_ef, y=ppo_ev, col sep=comma] {data/campaign_diag_critic_miniswe.csv};
  \addplot[forget plot, m jppo] table[x=joint_ppo_ef, y=joint_ppo_ev, col sep=comma] {data/campaign_diag_critic_miniswe.csv};
\nextgroupplot[xmin=0, xmax=1.8335, restrict x to domain=0:1.8335, ymode=log, yticklabel style={/pgf/number format/sci}, xlabel={training EFLOPs}, ylabel={value loss}]
  \addplot[forget plot, m ppo] table[x=ppo_ef, y=ppo_loss, col sep=comma] {data/campaign_diag_critic_mastermind.csv};
  \addplot[forget plot, m jppo] table[x=joint_ppo_ef, y=joint_ppo_loss, col sep=comma] {data/campaign_diag_critic_mastermind.csv};
\nextgroupplot[xmin=0, xmax=0.6838, restrict x to domain=0:0.6838, ymode=log, yticklabel style={/pgf/number format/sci}, xlabel={training EFLOPs}]
  \addplot[forget plot, m ppo] table[x=ppo_ef, y=ppo_loss, col sep=comma] {data/campaign_diag_critic_mbpp.csv};
  \addplot[forget plot, m jppo] table[x=joint_ppo_ef, y=joint_ppo_loss, col sep=comma] {data/campaign_diag_critic_mbpp.csv};
\nextgroupplot[xmin=0, xmax=13.7373, restrict x to domain=0:13.7373, ymode=log, yticklabel style={/pgf/number format/sci}, xlabel={training EFLOPs}]
  \addplot[forget plot, m ppo] table[x=ppo_ef, y=ppo_loss, col sep=comma] {data/campaign_diag_critic_appworld.csv};
  \addplot[forget plot, m jppo] table[x=joint_ppo_ef, y=joint_ppo_loss, col sep=comma] {data/campaign_diag_critic_appworld.csv};
\nextgroupplot[xmin=0, xmax=4.0096, restrict x to domain=0:4.0096, ymode=log, yticklabel style={/pgf/number format/sci}, xlabel={training EFLOPs}]
  \addplot[forget plot, m ppo] table[x=ppo_ef, y=ppo_loss, col sep=comma] {data/campaign_diag_critic_miniswe.csv};
  \addplot[forget plot, m jppo] table[x=joint_ppo_ef, y=joint_ppo_loss, col sep=comma] {data/campaign_diag_critic_miniswe.csv};
\end{groupplot}
\node[anchor=north] at ([yshift=-0.75cm]$(group c2r2.south)!0.5!(group c3r2.south)$) {\pgfplotslegendfromname{critleg}};
\end{tikzpicture}
\end{scaledfigure}
  \caption{\textbf{Critic diagnostics of PPO and delayed-joint PPO} from the training logs (ten-step centered mean), against training EFLOPs up to the compute budget. Top: explained variance of the value function on the training batch, clipped below at $-1$. Bottom: value loss (log scale). Delayed-joint PPO's explained variance is above $0.95$ late in each run, after its reward has collapsed (Section~\ref{sec:results-compute}), so a well-fit critic does not prevent the collapse.}
  \label{fig:critic}
\end{figure}
\clearpage
\section{Qualitative Material}
\label{app:qualitative}

\subsection{What the Trained Playbooks Contain}
\label{app:playbook-examples}

We inspected the playbooks that the fourth harnessed visit reads, after three handoffs, at the final \acerl{} checkpoints evaluated when this analysis was done: Mastermind and MBPP at step 300 and AppWorld at step 490. Their tips are mostly distinct and task-specific. The 100 Mastermind playbooks hold 300 tips with 203 distinct texts, no text appearing more than three times; the tips name digits and positions implied by earlier guesses, although many still begin with the reflector's insight label (``Insight 1 suggests \ldots''), which the curator prompt asks it to omit. The 100 MBPP playbooks hold 301 tips with 237 distinct texts, but 94 of the 100 tasks are already solved at the first visit, so these tips rarely affect the reward. The 57 AppWorld playbooks hold 159 tips in 56 non-empty playbooks, with 154 distinct texts; they name the APIs, entities, and checks the task requires. MBPP tips are not used as examples because some learned MBPP solutions hardcode the displayed assertions (Appendix~\ref{app:benchmarks}).

\subsection{Examples of a Helpful Handoff}
\label{app:examples}

Each illustrative example is a held-out task at the checkpoint stated below: the first visit fails, the curator writes the displayed playbook, and visits 2--4 succeed. The sequence shows the playbook that preceded those visits; it does not isolate its causal effect.

\paragraph{AppWorld (step 490; task df61dc5\_1; rewards over visits 1--4: 0, 1, 1, 1).}
After the failed first visit, the playbook reads:
\begin{quote}\small
[1] The code must identify and like all transactions to and from roommates for the ongoing year, not just one.\\
{}[2] Check before submission: The total number of transactions to like must be greater than zero and all transactions identified must be liked before submission.\\
{}[3] The code must process all transactions to and from roommates by checking both directions (sent to and received from) for each roommate and ensure all are included in the like operation.\\
{}[4] Check before submission: The response from like\_transaction for each transaction must be successful, and the total count of liked transactions must match the total count of transactions to and from roommates.
\end{quote}
Visits 2--4 complete the task.

\paragraph{Mastermind (step 300; task mastermind\_33156\_3\_4; rewards over visits 1--4: 0.67, 1, 1, 1).}
After the first visit ends with partial credit, the playbook reads:
\begin{quote}\small
[1] Insight 1 suggests the next best guess could be a 1 in the second position to ensure all constraints are met. This provides a clear direction for the Generator to narrow potential code candidates.
\end{quote}
Visits 2--4 solve the game.
\section{Limitations}
\label{sec:limitations}

\acerl{} requires a harness whose handoff artifacts can be resampled and evaluated in isolation; agents that entangle acting, summarizing, and remembering in one generation fall back to chain-level credit, which our delayed-joint variant approximates. \acerl{} matches \grpo{}'s actor rollouts per task, but its reflector and curator calls still add generation to each step; our compute comparison (Appendix~\ref{app:compute}) counts update FLOPs only. The playbook is task-local, and we do not study transfer of experience across tasks. Our evidence concerns models up to 4B parameters, one training run per method, and four benchmarks; the edits-discarded control on AppWorld is evaluated only beyond its compute budget, and the MBPP results use an environment whose feedback shows the scored assertions (Appendix~\ref{app:benchmarks}); whether the same recipe is needed, or sufficient, at frontier scale is open. Finally, training models to pass experience to their future selves in autonomous software and interface agents raises safety questions about scoping, monitoring, and release that our task-local, verifier-bounded setting does not resolve.

\end{document}